\documentclass{article} % For LaTeX2e
\usepackage{collas2026_conference,times}
\usepackage{easyReview}

\usepackage{amsmath,amssymb,amsfonts}
\usepackage{algorithmic}
\usepackage{graphicx}
\usepackage{url}
\usepackage{multirow}
\usepackage{makecell}
\usepackage{bm}
\usepackage{hyperref}
\usepackage{booktabs}
\usepackage{xcolor}
\usepackage{wrapfig}
\usepackage{enumitem}
\usepackage{subcaption}

\definecolor{greenx}{rgb}{0.0, 0.5, 0.0}

\definecolor{redx}{rgb}{0.84, 0.04, 0.33}

\usepackage{amsmath,amsfonts,bm}

\def\eqref#1{equation~\ref{#1}}
\def\1{\bm{1}}

\DeclareMathAlphabet{\mathsfit}{\encodingdefault}{\sfdefault}{m}{sl}
\SetMathAlphabet{\mathsfit}{bold}{\encodingdefault}{\sfdefault}{bx}{n}

\usepackage{hyperref}
\hypersetup{
    colorlinks=true,
    linkcolor=red,
    filecolor=magenta,
    urlcolor=blue,
    citecolor=purple,
    pdftitle={Overleaf Example},
    pdfpagemode=FullScreen,
    }

\title{Rethinking the Teacher-Student Framework\\for Test-Time Adaptation}

\author{
\parbox{\textwidth}{
\centering
\vspace{2em}
Damian Sójka$^{1,2}$\thanks{Corresponding author: damian.sojka@doctorate.put.poznan.pl}\quad
Marc Masana$^{3}$\quad
Bartłomiej Twardowski$^{4,5}$\quad
Sebastian Cygert$^{6,7}$\\[1em]
\normalfont
$^1$Poznan University of Technology \qquad 
$^2$Akces NCBR \qquad
$^3$Graz University of Technology \\
$^4$IDEAS Research Institute \qquad
$^5$Computer Vision Center, Universitat Autonoma de Barcelona \\
$^6$NASK - National Research Institute \qquad
$^7$Gdańsk University of Technology \\ %[0.75em]
}
}

\collasfinalcopy % Uncomment for camera-ready version, but NOT for submission.

\begin{document}

\maketitle

\begin{abstract}
Test-Time Adaptation (TTA) has recently emerged as a promising strategy that allows the adaptation of pre-trained models to changing data distributions at deployment time, without access to any labels. To mitigate error accumulation, researchers have widely adopted the teacher-student framework, though its long-term stability is often taken for granted. In this work, we challenge the common strategy of setting the teacher weights to an exponential moving average of the student by showing that error accumulation still occurs, although it is mostly apparent on longer sequences compared to those commonly utilized. We analyze the stability-plasticity trade-off within the teacher-student framework and propose to use an intransigent teacher that does not update its weights. Surprisingly, we show that this simple change allows TTA methods to significantly improve their performance on multiple datasets with longer scenarios and result in increased robustness to changes in hyperparameters. Finally, we show that those changes can be seamlessly and effectively applied to various architectures and experimental setups, including semantic segmentation. The code is available at \url{https://github.com/dmn-sjk/intransigent_teacher}.
\end{abstract}

\section{Introduction}
\sloppy
Machine learning models typically assume that training and testing data originate from a similar distribution. However, in real-world applications, distribution shifts between training (source) and testing (target) data domains are common and can lead to performance issues throughout inference~\citep{texturebias,hendrycks2019robustness,wilds}. Test-Time Adaptation (TTA)~\citep{TENT} allows for an online adaptation of a pre-trained model to the changing data distributions during testing, in an unsupervised way. While many TTA methods have been developed in recent years~\citep{NOTE,goyal2022test,MarsdenD024roid,eata,SAR,rusakSPEGBBB22,sun2020TTT,COTTA,rotta}, important challenges remain, such as adaptation over very long scenarios~\citep{rdumb}, robustness to noisy data~\citep{gong2023sotta}, and sensitivity to hyperparameter change~\citep{boudiafMAB22_parameterfree, zhao_2023_pitfallsTTA,cygert2024realistic}.

The teacher-student paradigm is a popular framework in TTA~\citep{AdaContrast,dobler2023CVPRmeanteacher,Sojka_2023_ICCV,COTTA,rotta,zhou2024resilient}, where the teacher weights are updated as the Exponential Moving Average (EMA) of the student weights (i.e., EMA teacher). This strategy follows pioneering works in semi-supervised learning~\citep{laine2017temporal,tarvainen2017mean}, representation learning~\citep{BYOL,he2020momentum,oquab2024dinov}, and learning under label noise~\citep{liu2020early,Nguyen2020SELF}. The averaged model provides more accurate and consistent predictions, which the student uses for training. While recent TTA survey~\citep{maharana2026continual} suggests this framework inherently prevents model collapse and enables state-of-the-art performance, our findings challenge this consensus.

\begin{wrapfigure}{R}{0.48\textwidth}
\vspace{-1.7em}
\centering
\includegraphics[width=0.45\textwidth]{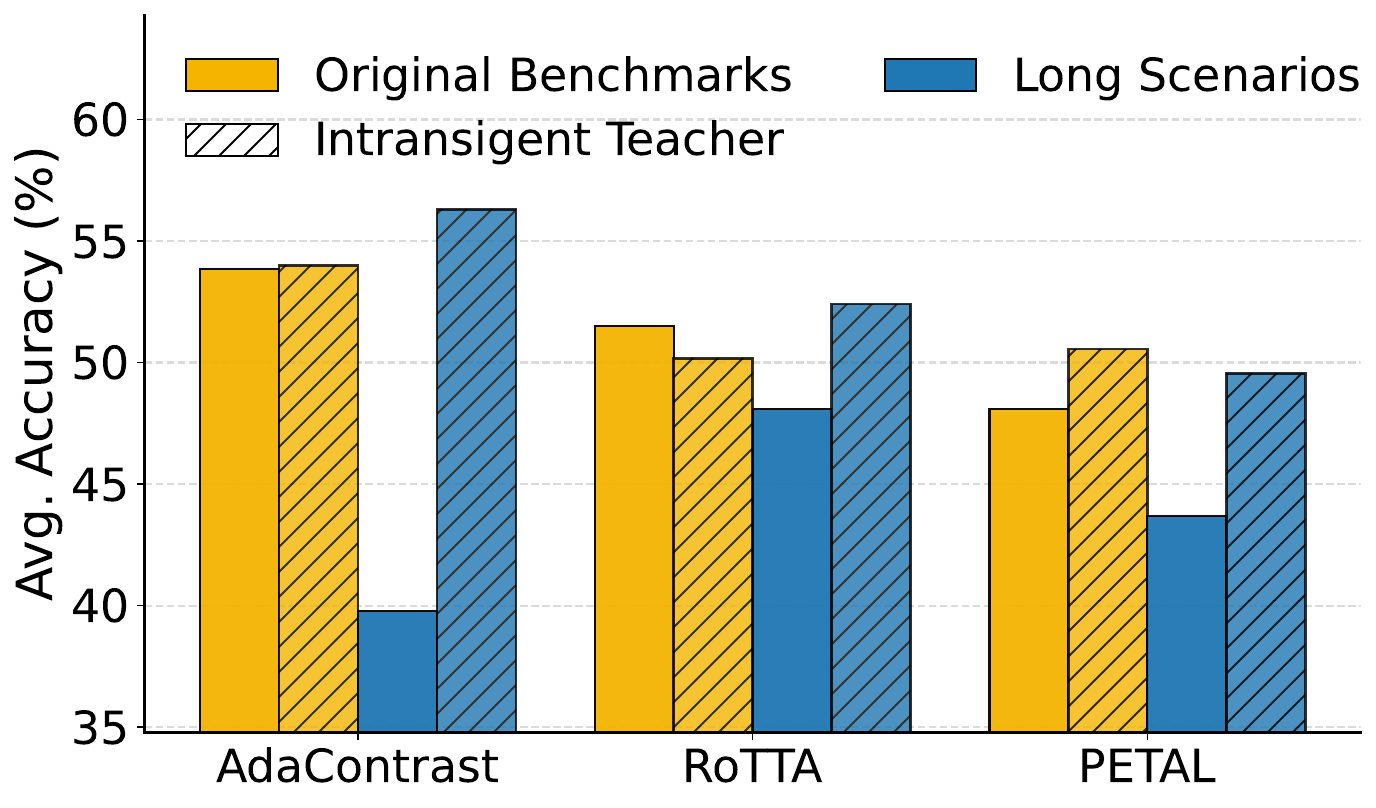}
\caption{
Average classification accuracy [\%] for the baseline methods, with and without our Intransigent Teacher technique, evaluated across the original benchmarks and their long variants (20$\times$ repetitions). % using batch sizes of 10 and 64. 
While baseline methods exhibit performance degradation as test sequence length increases, our Intransigent Teacher modification to the teacher-student framework effectively mitigates this decay and maintains robust accuracy.
}
\label{fig:teaser}
\vspace{-3em}
\end{wrapfigure}

In this work, we present experimental evidence indicating that the teacher-student framework does not strictly prevent error accumulation, which can result in model collapse (i.e., falling below the accuracy of the source model). Moreover, we demonstrate that state-of-the-art TTA methods relying on this stabilization~\citep{AdaContrast,COTTA,rotta,petal} are prone to significant accuracy degradation on extended test sequences. The EMA teacher merely postpones the effects of error accumulation. This reveals a critical flaw in common evaluation protocols, which have previously failed to detect this issue. In search for a solution, we observe that employing a simple technique of making the teacher more intransigent (not updating the teacher's weights) prevents model collapse over very long testing sequences (see Fig.~\ref{fig:teaser}). Interestingly, such a teacher can also enable students to fulfill the age-old cliche of surpassing their teacher (see Fig.~\ref{fig:old_teaser}). Based on that finding, we take a closer look at the stability-plasticity trade-off~\citep{mermillod2013stability} within teacher-student frameworks and how it affects the final model performance. We show that while increased teacher plasticity can lead to better performance on the current data in the short run, using a more stable teacher leads to a better adaptation over longer scenarios.
This is due to the error accumulation of the plastic model, which adapts well to the current data but fails to maintain generalization over time, as explored in Sec.~\ref{sec:tradeoff}.

\noindent Our contributions are as follows:
\begin{itemize}[rightmargin=0.5em]
    \item We analyze the teacher-student framework commonly used in TTA, demonstrating that an EMA teacher fails to prevent model collapse on longer test sequences.
    \item We observe that maintaining the teacher's weights fixed during adaptation (Intransigent Teacher -- IT), effectively prevents model collapse while allowing the student to surpass the teacher. We find that IT works well because it avoids negative flips and tends to keep the representation strength intact. 
    \item This simple solution (i.e., IT) can be combined with state-of-the-art methods, mitigating error accumulation by easily replacing any EMA teacher and obtaining reliable performance. IT can serve as a simple yet effective comparative baseline for future TTA works.
\end{itemize}

% --------------------------------------------------------------
\section{Related work}

\noindent \textbf{Test-time adaptation.} 
TTA~\citep{TENT} aims to adapt a pre-trained model to shifting data distributions during testing, without any labels. The adaptation is typically performed using unsupervised optimization objectives, often relying on noisy pseudo-labels, which can lead to error accumulation over multiple iterations. Therefore, numerous strategies have been developed to circumvent that issue, including various reset strategies~\citep{COTTA, ZhangLF22Memo,rdumb}, filtering unreliable samples~\citep{eata,resitta}, and the teacher-student framework~\citep{AdaContrast,COTTA,rotta,Sojka_2023_ICCV,petal,shi2025controllable,resitta}, among others. The teacher-student framework was introduced to TTA by the CoTTA method~\citep{COTTA}, where the use of exponential moving average (EMA) of weights was proposed. However, as EMA of the student’s weights forms the teacher, error accumulation is merely delayed rather than eliminated. In this work, we analyze the teacher-student framework to address these limitations.
Other works have explored strategies that retain the fixed source model. ROID~\citep{MarsdenD024roid} continually ensembles source and gradient-updated model weights to stabilize the adaptation. GRoTTA~\citep{grotta} and TRIBE~\citep{tribe} expand on the teacher-student framework by incorporating the source model as a third one for additional regularization. While these methods employ complex designs with extra loss terms and models, we explore how a simple teacher-student framework modification can address the performance degradation problem in TTA, especially for longer sequences.
Parallel works~\citep{zhou2024resilient,hoang2024persistent} also propose to evaluate existing TTA methods over very long scenarios.
Our work is complementary, since they propose adaptation methods 
(which require parameter tuning and access to source data~\citep{hoang2024persistent}), while we focus on evaluating existing approaches on a more extensive experimental setup, and observe that a simpler modification improves the reliability in such conditions.

\noindent \textbf{Plasticity-stability trade-off.} 
EMA ensemble of weights is parametrized by the $\beta$ trade-off parameter, which determines the balance between retaining old averaged weights of the teacher and incorporating newly updated student ones, commonly set in TTA to 0.999~\citep{COTTA} (following temporal ensembling works~\citep{laine2017temporal}), where \mbox{$\beta\!=\!1$} means full stability (frozen model), and $\beta\!=\!0$ means maximal plasticity. The trade-off has been widely analyzed in continual learning~\citep{mermillod2013stability,Chaudhry_2018_ECCV,masana2022class}, where the learner needs to balance learning of new tasks with the risk of forgetting the previously acquired knowledge. While a variety of strategies have been developed in continual learning, the most successful ones are those that promote stability. Many recent works freeze the feature extractor and learn only the classification head~\citep{ma2023progressive,panos2023first,goswami2024fecam,tscheschner2025incremental}.
Our work is inspired by those studies and aims to analyze the plasticity-stability trade-off within TTA.

\noindent \textbf{Teacher-student framework.} In knowledge distillation~\citep{Hinton2015DistillingTK}, usually a larger model (teacher) guides the optimization of the target model (student) by providing informative training signals (teachers' outputs). Such a strategy is also commonly used in continual learning to prevent forgetting of previous tasks~\citep{der,kirkpatrick2017overcoming}. Self-distillation is a special case in which the teacher and student have the same architecture. It has been shown that in such a scenario, the student can outperform its teacher~\citep{BornAgain}. In their work, the teacher is updated at the end of every training epoch by copying students' weights. They show improvement gains until such a procedure is repeated three times. Since TTA lacks access to labeled data, updating the teacher might result in even more significant error accumulation.

% --------------------------------------------------------------
\section{Preliminaries}
This study addresses continual TTA. The goal is to adapt a pre-trained (source) model $f_{\bm{\theta}_0}(\cdot)$, initially trained on a labeled dataset $\mathcal{D}_s\!\!=\!\!\{(\bm{x}_i, \bm{y}_i)\}_{i=1}^{N_s}$ with $\bm{x}_i\!\sim\!\mathcal{P}_s(\bm{x})$, to a sequence of unlabeled test data batches arriving sequentially during inference:
\begin{equation*}
\mathcal{D}_t^{(1)},\;
  \cdots,\;
  \mathcal{D}_t^{(T)}
\;\;
\text{where}\;
\mathcal{D}_t^{(k)}\!=\!\{\bm{x}_j^{(k)}\}_{j=1}^{N_k},\;\;
\bm{x}_j^{(k)}\!\sim\!\mathcal{Q}^{(k)}(\bm{x}),
\end{equation*}
and test distributions evolve over time: \mbox{$\mathcal{Q}^{(k)}(\bm{x}) \neq \mathcal{Q}^{(k+1)}(\bm{x})$} for some $k$, with $\mathcal{Q}^{(k)}(\bm{x}) \neq \mathcal{P}_s(\bm{x})$ in general (i.e., test data is out-of-distribution and non-stationary). Adaptation must be performed on-the-fly using only the current and/or past unlabeled test batches, without access to source data or labels.

Both the student and the teacher start the adaptation with the same set of weights $\bm{\theta}_0$ pre-trained on the source data. During TTA, the student's weights $\bm{\theta}^{s}_n$ at step $n$ are updated via backpropagation. The teacher's weights are updated as an EMA of the student's weights:
\begin{equation}
    \bm{\theta}_n^{t} = \beta \cdot \bm{\theta}_{n-1}^{t} + (1 - \beta) \cdot \bm{\theta}_n^{s}
\label{eq:ema}
\end{equation}
where $\beta$ (e.g., 0.999) is the EMA factor controlling the momentum of teacher’s update. The teacher is usually used for the final predictions.

% --------------------------------------------------------------
\section{Preliminary experiments}
\label{sec:prel_exp}
In this section, we aim to gain a deeper understanding of the adaptation process within the teacher-student framework, with particular emphasis on the plasticity-stability trade-off.
To this end, we conduct two experiments. 
In the first experiment~(Sec.~\ref{sec:intrans_exp}), adaptation relies exclusively on a simple self-supervised loss combined with the teacher-student framework, deliberately excluding any additional regularization or stabilization techniques. This minimal setup allows us to isolate the intrinsic behavior of the teacher-student mechanism.
In the second experiment~(Sec.~\ref{sec:tradeoff}), we systematically vary the level of plasticity across the teacher-student frameworks used in existing TTA methods and analyze the effects of low-to-none plasticity.

\subsection{Effect of teacher-student framework on the simple self-supervised adaptation}
\label{sec:intrans_exp}
The teacher-student framework is widely studied in TTA, with great results being achieved by popular methods such as AdaContrast~\citep{AdaContrast} and CoTTA~\citep{COTTA}. Although those techniques rely on other components (e.g., memory queue or weight restoration), they share a common trend of incorporating a self-supervision loss. We dissect the objectives directly related to the learning (their self-supervised losses), which allows for analyzing the teacher-student framework impacted only by the component the most related to the stability-plasticity trade-off.
\begin{wrapfigure}{R}{0.5\textwidth}
    \vspace{-1em}
    \centering
    
    % Table 1
    \captionof{table}{Mean accuracy [\%] of student models on test-time adaptation benchmarks. ET stands for exponential moving average teacher, and IT indicates the intransigent teacher. IN-C (L) stands for the ImageNet-C adaptation sequence being repeated 20 times.}
    \label{tab:motivation}
    \resizebox{0.85\linewidth}{!}{
    % \begin{table}
% \caption{Mean accuracy [\%] of student models on test-time adaptation benchmarks. ET stands for exponential moving average teacher, and IT indicates the intransigent teacher. IN-C (L) stands for the ImageNet-C adaptation sequence being repeated 20 times.}
% \centering
% \resizebox{0.99\linewidth}{!}{

\begin{tabular}{l@{\hskip 0.1in}c@{\hskip 0.1in}ccc}
\toprule
\textbf{Loss} & \textbf{Teacher} & \textbf{IN-C} & \textbf{IN-C (L)} & \textbf{CCC} \\
\midrule
Source
& none & 18.0 & 18.0 & 16.8 \\
\midrule
\multirow{ 2}{*}{Consistency} 
& ET & 27.3 & 7.9 & 1.6 \\
& IT & \textbf{28.8} & \textbf{31.3} & \textbf{27.2} \\
\midrule
\multirow{ 2}{*}{Contrastive}
& ET & \textbf{35.5} & 22.1 & 5.8 \\
& IT & 35.4 & \textbf{36.9} & \textbf{31.8} \\
\bottomrule
\end{tabular}
% }
% \label{tab:motivation}
% \end{table}

    }%
        
    \vspace{1.5em}
    
    \includegraphics[width=0.99\linewidth]{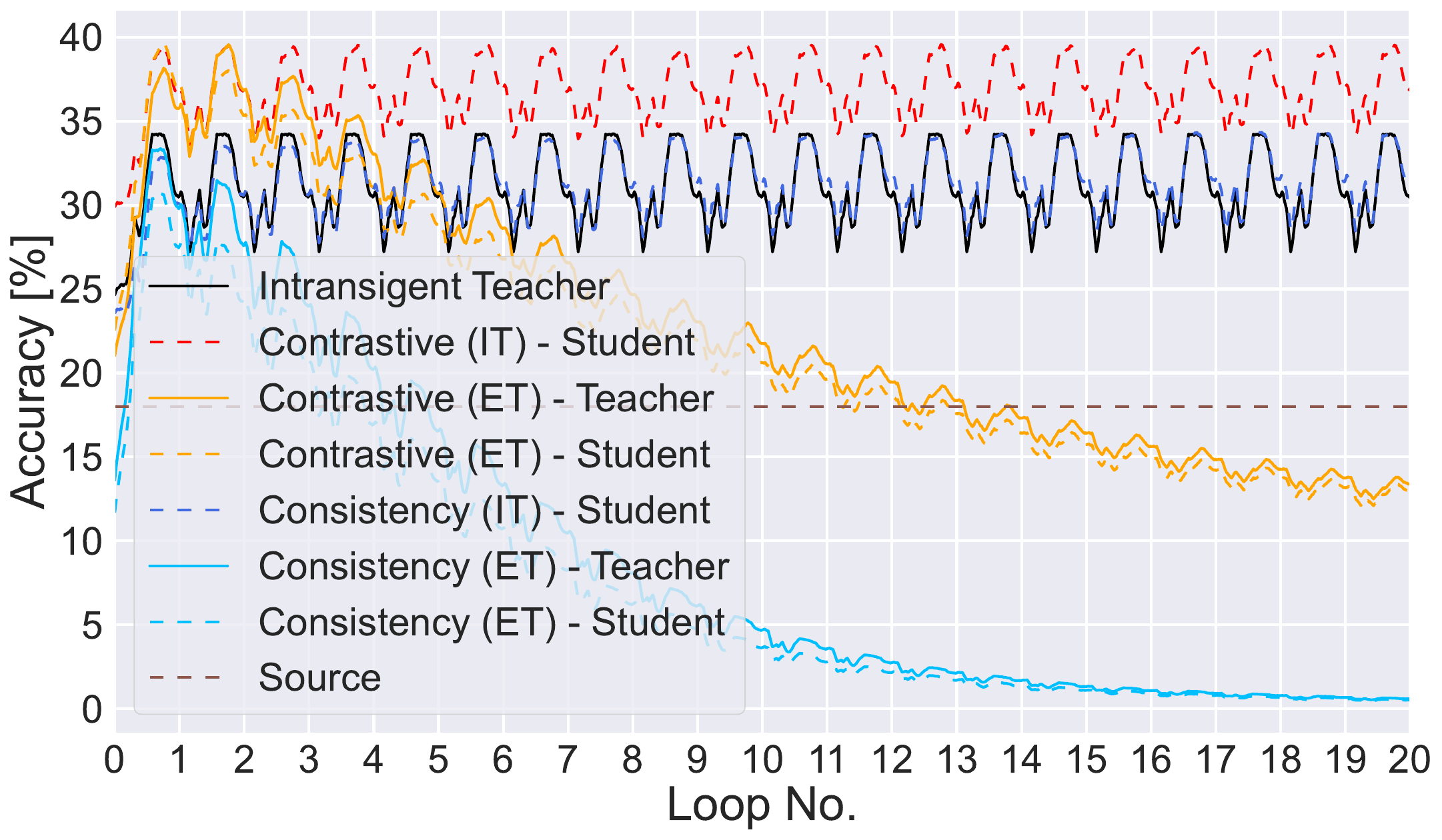}
    \captionof{figure}{Per-batch accuracy on repeated ImageNet-C (20 loops) with EMA teacher (ET) and intransigent teacher (IT), both for teacher (solid) and student (dashed). Only the self-supervised losses, without any additional components (or restart mechanisms), allow for successful adaptation over long sequences when the teacher is intransigent.}
    \label{fig:motivation}
    \vspace{-5em}
\end{wrapfigure}
We probe standard EMA teachers (ET) as proposed in their original works (both using $\beta\!=\!0.999$) and compare them with the presented intransigent teacher (IT) technique ($\beta\!=\!1$). Intransigent teachers have all trainable parameters frozen. Their batch normalization statistics are calculated on a per-batch basis~\citep{bn_stats_adapt}, as commonly done in TTA~\citep{eata, SAR, TENT, COTTA}. 
The final predictions are taken from the student model. As a motivation example, we evaluate on the popular \mbox{ImageNet-C} corruption benchmark~\citep{hendrycks2019robustness} on a common level~5 corruption sequence from~\citet{COTTA}. Furthermore, we evaluate on the CCC benchmark~\citep{rdumb} and introduce the setting with ImageNet-C repeated 20 times (ImageNet-C (L)), to better assess the teacher-student framework's performance on longer scenarios. The source model is trained on the clean ImageNet dataset~\citep{imagenet}. Figure~\ref{fig:motivation} shows accuracy over time and Table~\ref{tab:motivation} shows numerical results, where the evaluated objectives are described as:
\begin{itemize}
    \item Consistency:$\ $ The loss from CoTTA~\citep{COTTA}, which minimizes the cross-entropy consistency between teacher and student predictions. The input of the teacher is transformed via augmentations.
    \item Contrastive:$\ $ The adaptation objective from AdaContrast~\citep{AdaContrast}, which uses a MoCo-inspired~\citep{he2020momentum} contrastive task in which features from different views of the same image (positive pairs) are pulled closer, while features from different images (negative pairs) are pushed away. Input for both teacher and student is augmented by randomly drawing two strong augmentations.
\end{itemize}

\noindent The results from the above experiment lead to the following observations:

\noindent \textbf{Observation 1:}
Self-supervised objective combined with an intransigent teacher provides great reliability on its own. This is a very positive result, as recent work~\citep{rdumb} observed that on CCC, most of the current adaptation methods result in model collapse. This is especially interesting in the case of Contrastive, where the contrastive loss changes only the backbone of the model without changing the classification layer, mostly relying on feature alignment.

\noindent \textbf{Observation 2:}
An EMA on its own does not prevent error accumulation. As clearly shown in Figure~\ref{fig:motivation}, the problems when using EMA take some time to become apparent and are only visible over long sequences. For both losses, performance degradation starts rather early, after the 2nd and 3rd loops, respectively.

\noindent \textbf{Observation 3:}
When using EMA, there is a small gap between the teacher and student accuracies. When an intransigent teacher with consistency loss is used, the student performance is reliable and comparable to the teacher, while in the case of the contrastive loss, the student is able to outperform their teacher significantly.

\subsection{Effects of intransigence}
\label{sec:tradeoff}
\begin{figure}[t]
  \centering
  \begin{minipage}{0.48\columnwidth}
    \centering
    \includegraphics[width=0.76\columnwidth]{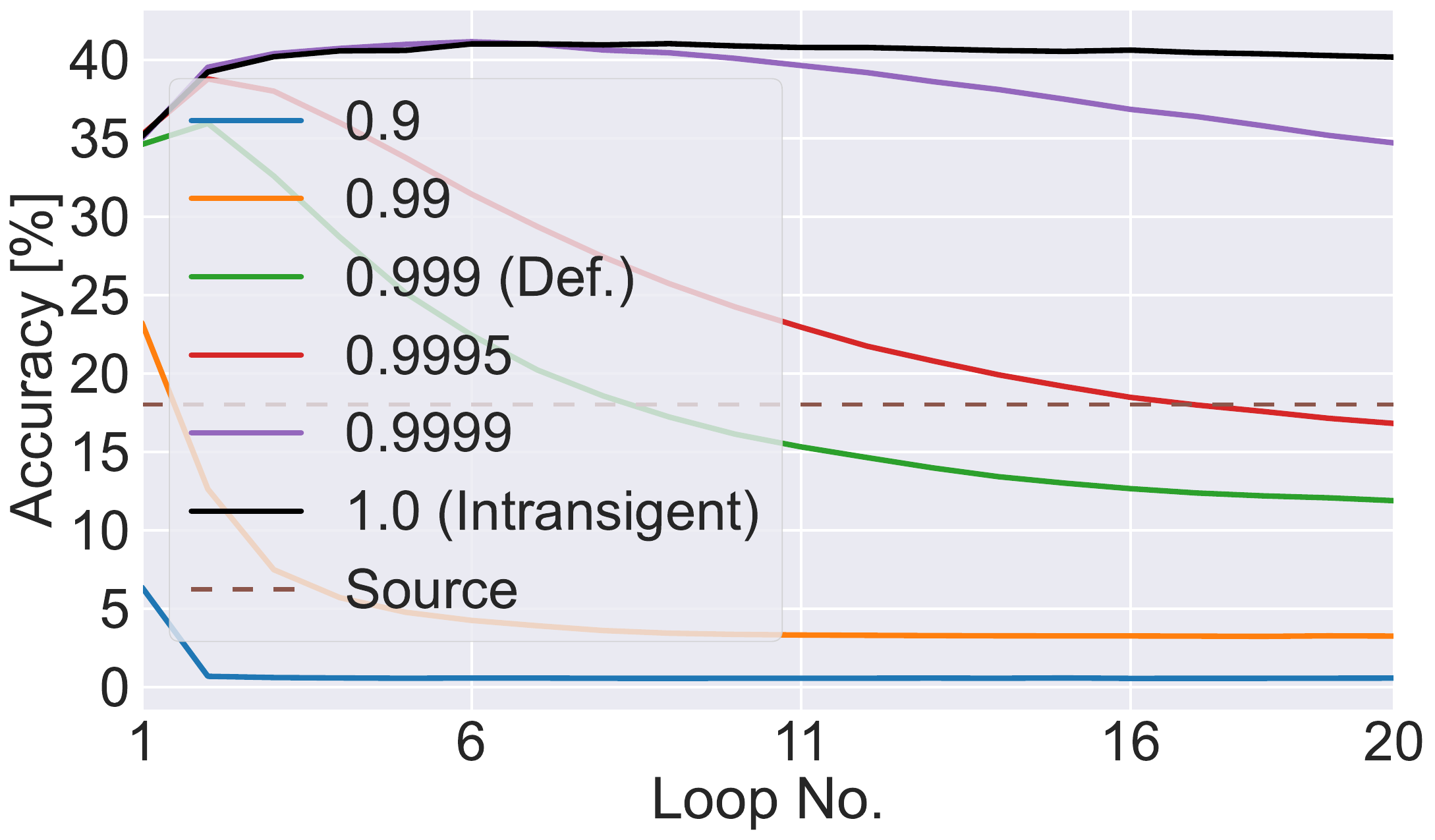}
  \label{fig:ema_ada_inc}
  \end{minipage}
  \begin{minipage}{0.48\columnwidth}
    \centering
    \includegraphics[width=0.76\columnwidth]{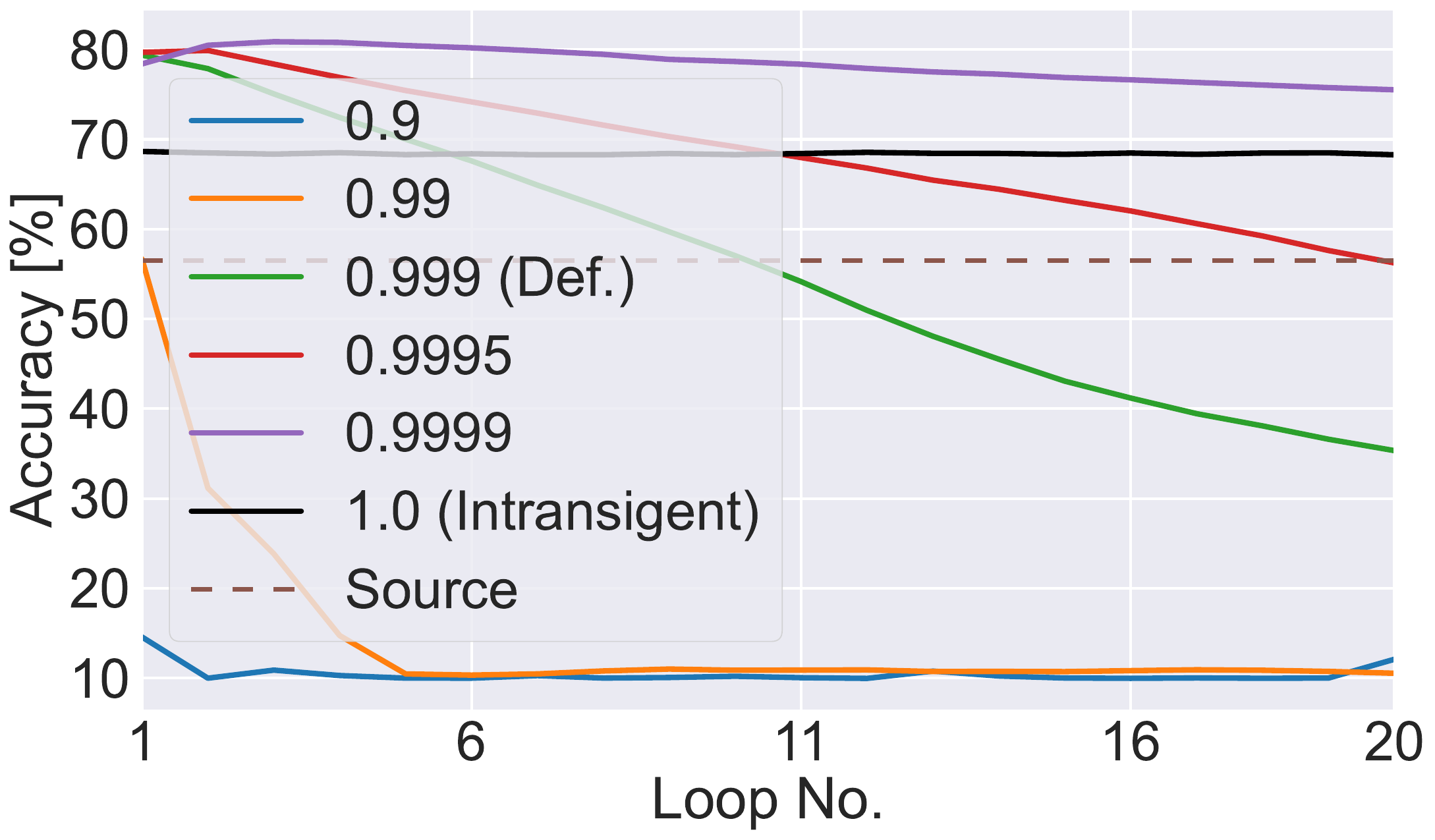}
  \label{fig:ema_cotta_cifar}
  \end{minipage}
    \caption{Mean accuracy [\%] for each loop of common testing sequence on ImageNet-C (L) with AdaContrast \textbf{(left)} and CIFAR10-C (L) with CoTTA \textbf{(right)}. The brown dashed line indicates the source model accuracy as a reference.}
  \label{fig:ema}
\end{figure}

To better understand the plasticity-stability trade-off, we evaluate performance when varying \mbox{$\beta\!\in\![0.9, 1.0]$}, where $1.0$ means using an intransigent teacher, and $0.999$ is the default value for the two original methods we compare: AdaContrast and CoTTA.
We evaluate on \mbox{CIFAR10-C} (C10-C) and ImageNet-C \mbox{(IN-C)}~\citep{hendrycks2019robustness} benchmarks, with each common adaptation sequence from~\citet{COTTA} repeated 20 times (L). The source model is either trained on clean CIFAR10~\citep{cifar} or ImageNet~\citep{imagenet}, depending on the benchmark. The average final performance is reported in Table~\ref{tab:ema_param} (Appendix), while the accuracy evolution throughout the sequence is shown in Figure~\ref{fig:ema}, with the accuracy on the common-length benchmarks indicated at the first tick of the x-axis. 
Results with different method--dataset combinations are presented in the Appendix~(Fig.~\ref{fig:ema_cotta_ada} and Fig.~\ref{fig:ema_petal}).

Intransigent teachers guide students more consistently than lenient alternatives. While increasing plasticity (decreasing $\beta$) improves short-term student accuracy, performance tend to collapse over longer sequences.
For example, using an EMA with $\beta=0.9999$ initially outperforms IT, but accuracy degrades significantly toward the end of the test sequence. Consequently, fixed teacher weights provide the most reliable long-term performance.
We further demonstrate this vulnerability in Figure~\ref{fig:100loops} in the Appendix using CoTTA on ImageNet-C. Although a slowly-updating teacher ($\beta=0.9999$) achieves superior results over 20 repetitions, its performance falls significantly below the IT baseline when the sequence extends to 100 repetitions. 
This highlights a critical failure mode in TTA. Methods often depend on adaptation sequence lengths. While carefully tuned EMA parameters may mitigate this, hyperparameter selection for TTA remains an open challenge \citep{boudiafMAB22_parameterfree,zhao_2023_pitfallsTTA} and may necessitate specific assumptions about the deployment environment, such as the length of the adaptation sequence.

% --------------------------------------------------------------
\section{Methodology}
As presented in previous sections, using an intransigent teacher provides a great model consistency throughout longer scenarios without encountering catastrophic error accumulation.  Here, we describe how we extend existing TTA methods with an intransigent teacher strategy for further experiments (Sec.~\ref{sec:method}) and provide the theoretical grounding for its effectiveness (Sec.~\ref{sec:theoretical_grounding}).

\begin{wrapfigure}{r}{0.4\textwidth}
    \centering
    \vspace{-3.5em}
    \includegraphics[width=0.95\linewidth]{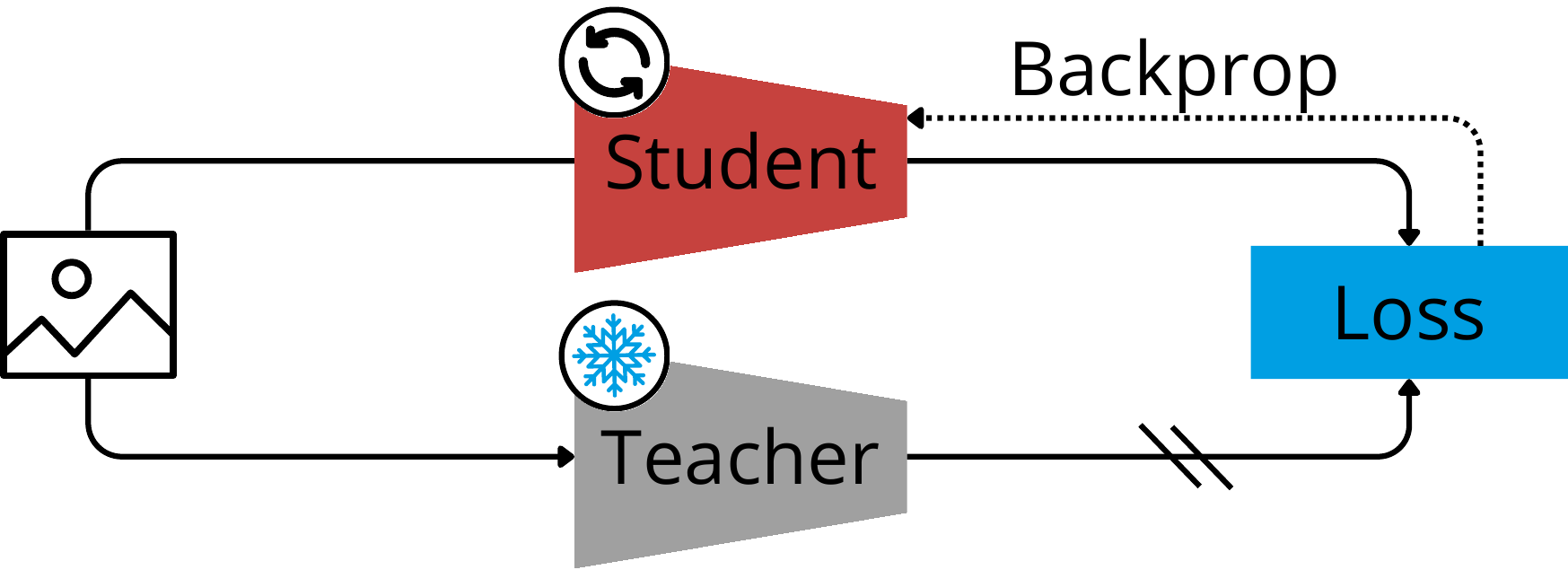}
    \caption{
    The Intransigent Teacher. The student's weights are updated using gradient-based optimization, while the teacher's weights remain fixed throughout adaptation.
    }
    \label{fig:it_diagram}
\end{wrapfigure}
\subsection{Appointing an intransigent teacher}
\label{sec:method}
We extend 4 popular TTA methods that originally utilize the teacher-student framework, AdaContrast~\citep{AdaContrast}, CoTTA~\citep{COTTA}, RoTTA~\citep{rotta}, and PETAL~\citep{petal}, with our proposed intransigent teacher. 
The main difference between the classic teacher-student framework with EMA teacher and the intransigent teacher is the value of the $\beta$ parameter. In the case of IT, the parameter $\beta$ is equal to 1, so the weights are fixed at their initial pre-trained values and Eq.~\ref{eq:ema} simplifies to:
\begin{equation}
\bm{\theta}_n^t = \bm{\theta}_{n-1}^t \quad\Longrightarrow\quad \bm{\theta}_n^t = \bm{\theta}_0 
\end{equation}

Additionally, the student model is used to generate the final predictions, regardless of which predictions are used in the original method.
We maintain the same usage and adaptation of batch normalization statistics as the underlying TTA method for both the teacher and student networks.
Figure~\ref{fig:it_diagram} illustrates our adaptation process. 
Descriptions of strategies extended with~IT are provided in the Appendix~(Sec.~\ref{sec:baseline_descr}).

\subsection{Analytical framework for intransigent teachers}
\label{sec:theoretical_grounding}
To understand why standard EMA teachers fail on extended sequences, we can formalize the feedback loop of noise inherent in TTA.
The student's weights $\bm{\theta}_{n}^{s}$ at step $n$ are updated via backpropagation on an adaptation loss $L$, relying on the teacher's pseudo-labels:
\begin{equation}
    \bm{\theta}_n^{s} = \bm{\theta}_{n-1}^{s} - \eta \cdot \nabla_{\bm{\theta}^s} L(\bm{\theta}_{n-1}^{s}, \bm{\theta}_{n-1}^{t}, \bm{x}_n)
\end{equation}
where $\eta$ is the learning rate, $\bm{\theta}^{t}_{n-1}$ represents the teacher weights at time $n-1$, and $\bm{x}_n$ is the test batch at time $n$.
Due to the absence of ground-truth supervision, this update consist of a constructive adaptation direction and a harmful gradient noise. We can explicitly decompose the student's weight change into these two directions:
\begin{equation}
    \bm{\theta}_n^s = \bm{\theta}_{n-1}^{s} + \bm{d}_n + \bm{\epsilon}_n
\end{equation}
Here, $\bm{d}_n = -\eta \nabla_{\bm{\theta}^s} L_{GT}(\bm{\theta}_{n-1}^{s}, \bm{y}_n^*, \bm{x}_n)$ represents the ideal gradient step dictated by an oracle providing ground-truth labels $\bm{y}_n^*$, and $\bm{\epsilon}_n$ captures the residual gradient noise injected by incorrect pseudo-labels generated by the teacher.
By unrolling the recurrence relation over time $n$ in the EMA teacher update (Eq.~\ref{eq:ema}), we can express the teacher's current weights as a geometrically weighted sum of the source model weights ($\bm{\theta}_0$) and all past student states:
\begin{equation}
    \bm{\theta}_n^t = \beta^n \bm{\theta}_0 + (1-\beta) \sum_{i=1}^n \beta^{n-i} \bm{\theta}_i^s
\end{equation}
This unrolled equation reveals the critical flaw of using $\beta < 1$ for continuous adaptation. Over infinite time horizons (as $n \to \infty$), the $\beta^n \bm{\theta}_0$ term decays to zero, meaning the teacher completely loses its anchor to the robust, pre-trained source distribution. Furthermore, the teacher permanently absorbs the accumulated student noise ($\sum \bm{\epsilon}_i$). Because the student's loss function uses the teacher as a reference, a degraded teacher generates lower-quality pseudo-labels for the next batch. This, in turn, increases the magnitude of the noise $\bm{\epsilon}_{n+1}$ in the student's next update. This creates a destructive positive feedback loop, illustrating why EMA merely delays, rather than prevents, catastrophic error accumulation on long sequences.
When we apply the IT strategy by setting $\beta = 1$, the EMA recurrence simplifies entirely:
$\bm{\theta}_n^t = \bm{\theta}_0$.
While it might seem that the student network would merely mimic the frozen teacher, it is instead optimized using consistency or contrastive losses across distinct, augmented views of the data. We attribute this effective learning signal to the forced alignment between the two networks, where one processes a heavily augmented view of the input while the other operates on a clean version.
By freezing the teacher, we decouple the pseudo-label generator from the student's noisy updates.
While the teacher's supervision is inherently noisy due to the domain shift, this noise distribution remains stationary and non-compounding. This prevents the destructive positive feedback loop of error accumulation.
We experimentally verify the error accumulation preventing behavior in Sec.~\ref{seq:negative_flips}.

% --------------------------------------------------------------
\section{Experiments}
Following the insights from preliminary experiments in Sec.~\ref{sec:prel_exp}, we evaluate the intransigent teacher over longer scenarios and with various state-of-the-art methods. The efficacy of the proposed baseline is validated across various model architectures and hyperparameter selections. We assess the error accumulation and the feature representation strength in student models for both original baselines and IT-enhanced approaches. Further details on the experimental setup are provided in Appendix Sec.~\ref{sec:app-exp-setup}.

\noindent \textbf{Benchmarks}.
We adopt the popular corruption benchmarks from~\citet{COTTA, eata, dobler2023CVPRmeanteacher}: \mbox{CIFAR10-C} (C10-C) and \mbox{ImageNet-C} (IN-C)~\citep{hendrycks2019robustness}, using the standard corruption sequence at the highest severity. The adaptation on natural domain shifts is tested on DomainNet-126 (DN-126)~\citep{peng2019moment} and ImageNet-R (IN-R)~\citep{hendrycks2021many}. For DN-126, we use the real domain as the source data and conduct experiments on the remaining domains. Our goal is to focus on very long~(L) adaptation sequences to evaluate the methods in terms of stability during long-time operation. For that reason, we repeat the standard test sequences 20 times, and additionally test on the  CCC~\citep{rdumb} long sequence without repetition.

\noindent \textbf{Baselines}. We apply our modification to 4 teacher-student state-of-the-art frameworks: AdaContrast~\citep{AdaContrast}, CoTTA~\citep{COTTA}, RoTTA~\citep{rotta}, PETAL~\citep{petal}, and report their performance with IT using (I-) prefix. We compare the results with the original methods and the following strategies that do not utilize classic teacher-student framework: 
TENT~\citep{TENT}, EATA~\citep{eata}, SAR~\citep{SAR}, RDUMB~\citep{rdumb}, MEMO~\citep{ZhangLF22Memo}, and PeTTA~\citep{hoang2024persistent}.
The non-adapted model is indicated as Source, and TestBN is a fixed model using batch normalization statistics from the current batch~\citep{li2016revisiting, bn_stats_adapt} as commonly done in TTA~\citep{TENT, COTTA, eata}.

\noindent \textbf{Architectures.} Consistent with previous works~\citep{COTTA, MarsdenD024roid}, we use \mbox{WideResNet-28}~\citep{zagoruyko2016wide} models with pre-trained weights from \textsc{RobustBench}~\citep{croce2021robustbench} for the main experiments on \mbox{CIFAR10-C}. Similarly, evaluation on ImageNet and DomainNet-126 employ the ResNet50 network with weights sourced from the same model zoo or those provided by~\citet{MarsdenD024roid} for DomainNet-126. Additional experiments in Sec.~\ref{sec:analysis} are carried out with ResNet-26GN (RN26GN)~\citep{WuH18GN}, ResNeXt-50~32x4d~(RNXt-50)~\citep{xie2017aggregated}, \mbox{ViT-B16}~\citep{DosovitskiyViT}, \mbox{SwinViT-T} (SViT-T)~\citep{liu2021swin} and ConvNeXt tiny (CNXt tiny)~\citep{convnext} architectures. Weights for RN26GN are taken from~\citet{ZhangLF22Memo}, as in~\citet{MarsdenD024roid}. \textsc{Torchvision} 
is used to initialize the rest of the mentioned models.

\subsection{On adaptation over long scenarios}
\label{sec:long}
\noindent \textbf{The vulnerability of EMA teachers is revealed on extended test sequences.}
In long scenarios (see Tab.~\ref{tab:long}), methods based on teacher-student framework often achieve lower performance compared to other baselines or even cause the model to collapse, which is especially apparent on \mbox{ImageNet-C} and CCC benchmarks. 

Those shortcomings are amplified in the lower batch size setting, potentially caused by a more significant error accumulation due to difficulties in estimating batch statistics and a larger number of adaptation steps. While the goal of using EMA is to provide a more stable adjustment and more accurate pseudo-labels for adaptation, we find this to not hold true for the longer settings. This is in contrast to the evaluation on common sequence length (see Appendix~Tab.~\ref{tab:short}), where the EMA teacher performs well and the original methods don't cause model collapses in most cases, except for the more challenging lower batch size setting. Results show that the performance of state-of-the-art TTA methods is clearly test sequence-length dependent.

\noindent \textbf{Intransigent teachers are very effective at collapse prevention.}
The evaluated original teacher-student adaptation methods exhibit at least 1\% decrease in model accuracy in 21 out of 40 cases for long sequences (Tab.~\ref{tab:long}, results of AdaContrast, CoTTA, RoTTA, and PETAL on five benchmarks and two batch sizes). When using an intransigent teacher, the collapse happens only three times for the small batch size. However, in these cases, performance is at least close to the source model and may still yield great improvements over the baseline - for instance, I-CoTTA achieves a 44.3\% increase in accuracy on \mbox{DomainNet-126 (L)}.

\begin{wraptable}{R}{0.5\textwidth}
    % \begin{table}[b!]
\caption{Classification accuracy [\%] for long (L) scenarios. Superscript indicates improvements over the baseline. \textbf{Bold} indicates best performing method. \textcolor{gray}{Gray} color indicates model collapse -- performance worse than the non-adapting model (Source). Results averaged from 3 random seeds. \textsuperscript{*} indicates approximated result, details in Appendix.
}
\centering
\resizebox{1.\linewidth}{!}{
\begin{tabular}{lcccccc}
\toprule
\textbf{Method} & \textbf{C10-C (L)} & \textbf{IN-C (L)} & \textbf{IN-R (L)} & \textbf{DN-126 (L)} & \textbf{CCC} & \textbf{Avg.}
\\ 
\midrule
Source & 56.5 & 18.0 & 36.2 & 54.7 & 16.8 & 36.4 \\
MEMO & 65.6 & 25.0 & 40.9 & \textcolor{gray}{53.2} & 19.3\textsuperscript{*} & 40.8 \\

\midrule
\multicolumn{7}{c}{\textsc{batch size 10}} \\
\midrule

TestBN & 75.1 & 26.9 & 36.2 & \textcolor{gray}{49.6} & 22.5 & 42.1 \\
TENT & \textcolor{gray}{39.0} & \textcolor{gray}{4.7} & \textcolor{gray}{17.4} & \textcolor{gray}{10.9} & \textcolor{gray}{0.7} & 14.5 \\
EATA & 73.6 & 36.4 & \textbf{44.0} & \textcolor{gray}{54.0} & \textbf{29.7} & 47.6 \\
SAR & 75.2 & 30.6 & 43.6 & \textcolor{gray}{50.8} & 20.3 & 44.1 \\
RDUMB & 76.8 & 34.3 & 40.1 & \textcolor{gray}{51.5} &  28.1 & 46.2 \\

PeTTA & 83.8 & 37.8 & 40.5 & 55.8 & \textcolor{gray}{13.7} & 46.3 \\

AdaCont. & 72.1 & \textcolor{gray}{2.3} & \textcolor{gray}{8.0} & \textcolor{gray}{47.0} & \textcolor{gray}{0.4} & 26.0  \\
I-AdaCont. 
& \textbf{84.1}\textsuperscript{\textcolor{greenx}{+12.0}}
& \textbf{39.5}\textsuperscript{\textcolor{greenx}{+37.2}}
& \textcolor{gray}{35.3}\textsuperscript{\textcolor{greenx}{+27.3}}
& \textbf{63.2}\textsuperscript{\textcolor{greenx}{+16.2}}
& 21.8\textsuperscript{\textcolor{greenx}{+21.4}}
& \textbf{48.8}\textsuperscript{\textcolor{greenx}{+22.8}} \\

CoTTA & \textcolor{gray}{23.8} & \textcolor{gray}{3.8} & \textcolor{gray}{33.7} & \textcolor{gray}{6.0} & 17.3 & 16.9 \\
I-CoTTA 
& 69.7\textsuperscript{\textcolor{greenx}{+45.9}}
& 27.6\textsuperscript{\textcolor{greenx}{+23.8}}
& 37.4\textsuperscript{\textcolor{greenx}{+3.7}}
& \textcolor{gray}{50.3}\textsuperscript{\textcolor{greenx}{+44.3}}
& 25.7\textsuperscript{\textcolor{greenx}{+8.4}}
& 42.1\textsuperscript{\textcolor{greenx}{+25.2}}
\\

RoTTA & 82.5 & 24.4 & 43.0 & \textcolor{gray}{45.6} & \textcolor{gray}{1.1} & 39.3 \\
I-RoTTA 
& 79.0\textsuperscript{\textcolor{redx}{-3.5}}
& 33.6\textsuperscript{\textcolor{greenx}{+9.2}} 
& 39.9\textsuperscript{\textcolor{redx}{-3.1}}
& 57.8\textsuperscript{\textcolor{greenx}{+12.2}}
& 23.4\textsuperscript{\textcolor{greenx}{+22.3}}
& 46.7\textsuperscript{\textcolor{greenx}{+7.4}}
\\
PETAL & 68.0 & \textcolor{gray}{3.9} & 37.9 & \textcolor{gray}{47.3} & \textcolor{gray}{0.5} & 31.5 \\
I-PETAL 
& 74.2\textsuperscript{\textcolor{greenx}{+6.2}}
& 28.7\textsuperscript{\textcolor{greenx}{+24.8}} 
& 36.5\textsuperscript{\textcolor{redx}{-1.4}}
& \textcolor{gray}{50.7}\textsuperscript{\textcolor{greenx}{+3.4}}
& \textcolor{gray}{13.9}\textsuperscript{\textcolor{greenx}{+13.4}}
& 40.8\textsuperscript{\textcolor{greenx}{+9.3}}
\\

\midrule
\multicolumn{7}{c}{\textsc{batch size 64}} \\
\midrule

TestBN & 79.1 & 31.4 & 39.6 & \textcolor{gray}{54.4} & 27.3 & 46.4 \\
TENT & \textcolor{gray}20.1 & \textcolor{gray}{11.1} & 36.4 & \textcolor{gray}{18.4} & \textcolor{gray}{1.2} & 17.4 \\

EATA & 61.6 & 43.3 & 49.2 & 61.9 & 36.3 & 50.5\\
SAR & 79.2 & 39.9 & 47.3 & 59.2 & 22.3 & 49.6 \\
RDUMB & 81.1 & 41.7 & 47.5 & 59.0 & \textbf{37.0} & \textbf{53.3} \\

PeTTA & 83.7 & 36.9 & 42.0 & 57.2 & \textcolor{gray}{13.4} & 46.6 \\

AdaCont. & 81.8 & 18.8 & \textcolor{gray}{26.5} & 61.7 & \textcolor{gray}{2.4} & 38.2 \\
I-AdaCont. 
& \textbf{85.4}\textsuperscript{\textcolor{greenx}{+3.6}}
& 40.4\textsuperscript{\textcolor{greenx}{+21.6}}
& 38.1\textsuperscript{\textcolor{greenx}{+11.6}} 
& \textbf{64.4}\textsuperscript{\textcolor{greenx}{+2.7}}
& 23.6\textsuperscript{\textcolor{greenx}{+21.2}} 
& 50.4\textsuperscript{\textcolor{greenx}{+12.2}}
\\

CoTTA & \textcolor{gray}{56.0} & \textbf{52.8} & \textbf{50.5} & \textcolor{gray}{45.6} & \textcolor{gray}{8.3} & 42.7 \\ %1 aug
I-CoTTA 
& 68.3\textsuperscript{\textcolor{greenx}{+12.3}}
& 35.4\textsuperscript{\textcolor{redx}{-17.4}}
& 39.5\textsuperscript{\textcolor{redx}{-11.0}}
& 56.0\textsuperscript{\textcolor{greenx}{+10.4}}
& 26.3\textsuperscript{\textcolor{greenx}{+18.0}}
& 45.1\textsuperscript{\textcolor{greenx}{+2.4}} % (1 aug)
\\

RoTTA & 82.3 & \textcolor{gray}{13.2} & 43.4 & \textcolor{gray}{50.3} & \textcolor{gray}{1.1} & 38.1 \\
I-RoTTA 
& 79.7\textsuperscript{\textcolor{redx}{-2.6}}
& 32.9\textsuperscript{\textcolor{greenx}{+19.7}}
& 39.7\textsuperscript{\textcolor{redx}{-3.7}}
& 56.6\textsuperscript{\textcolor{greenx}{+6.3}}
& 21.7\textsuperscript{\textcolor{greenx}{+20.6}}
& 46.1\textsuperscript{\textcolor{greenx}{+8.0}}
\\

PETAL & \textcolor{gray}{56.3} & 36.8 & 40.7 & 58.6 & \textcolor{gray}{13.1} & 41.1 \\
I-PETAL
& 78.8\textsuperscript{\textcolor{greenx}{+22.5}}
& 32.3\textsuperscript{\textcolor{redx}{-4.5}}
& 39.8\textsuperscript{\textcolor{redx}{-0.9}}
& 55.4\textsuperscript{\textcolor{redx}{-3.2}}
& \textcolor{gray}{16.3}\textsuperscript{\textcolor{greenx}{+3.2}}
& 44.5\textsuperscript{\textcolor{greenx}{+3.4}}
\\

\bottomrule
\end{tabular}
}
\label{tab:long}
% \end{table}

\end{wraptable}

\noindent \textbf{Intransigent teachers provide great reliability, out-of-the-box.}
Keeping the teacher model intransigent on a batch size 64 allows for accuracy improvements of 12.2 (\mbox{AdaContrast}), 2.4 (CoTTA), and 8.0 (RoTTA) percentage points on average. It is even more effective on a smaller batch size, where the respective improvements are 22.8, 25.2, and 7.4.
Note that applying this strategy did not require any hyperparameter tuning nor additional parameters. In fact, it removes the need to set $\beta$. 

\noindent \textbf{Teacher intransigence and the plasticity bottleneck.}
While the intransigent teacher provides a robust baseline, it prioritizes stability over plasticity. This trade-off limits adaptation to rapid distribution shifts and prevents the student from exploring parameter spaces distant from the source model.
This limitation is most evident when using CoTTA on ImageNet-C (L) and ImageNet-R (L) with a batch size of 64 (Tab.~\ref{tab:long}). In these scenarios, the use of IT reduces accuracy by 17.4 and 11.0 percentage points, respectively. We attribute this difference to the high baseline performance of CoTTA in these settings, as it outperforms all other tested methods. Here, IT over-regularizes the student and hinders its adaptation capabilities. While an EMA teacher may be preferable when hyperparameters can be precisely tuned, IT remains effective as it still outperforms the source model.
Crucially, IT prevents long-term model collapse. Negative flip rate analysis (Sec.~\ref{seq:negative_flips}) shows that for CoTTA, the number of samples correctly predicted by the source but incorrectly by the adapted model increases monotonically over repetitions. 
In contrast, the IT negative flip rate remains constant. 
This suggests that long enough sequences could eventually cause model collapse, similarly to the 100 repetitions experiment in Appendix Figure~\ref{fig:100loops}.

\noindent \textbf{Intransigent teacher provides competitive stability.}
PeTTA~\citep{hoang2024persistent}, a TTA method focused on long-term adaptation stability, employs a complex suite of techniques, including a class-balanced memory bank, robust batch normalization, weight- and prediction-based regularization from the source model, an adaptive EMA factor to control teacher updates, and the use of source feature statistics. Despite this complexity, our baseline methods, augmented with an intransigent teacher, achieve competitive or superior performance (Tab.~\ref{tab:long}). Specifically, I-AdaContrast outperforms PeTTA by 1.7 percentage points on CIFAR10-C (L) (batch size 64), while I-RoTTA leads by 2 percentage points on ImageNet-C (L) (batch size 10). On average, I-AdaContrast improves accuracy over PeTTA by 3.8 percentage points (50.4\% vs. 46.6\%). These results demonstrate that preventing error accumulation in the teacher model is a highly effective, yet simple, strategy for ensuring long-term stability.

\subsection{Detailed analysis}
\label{sec:analysis}

\begin{wrapfigure}{R}{0.5\textwidth}
    \centering
    \includegraphics[width=0.99\linewidth]{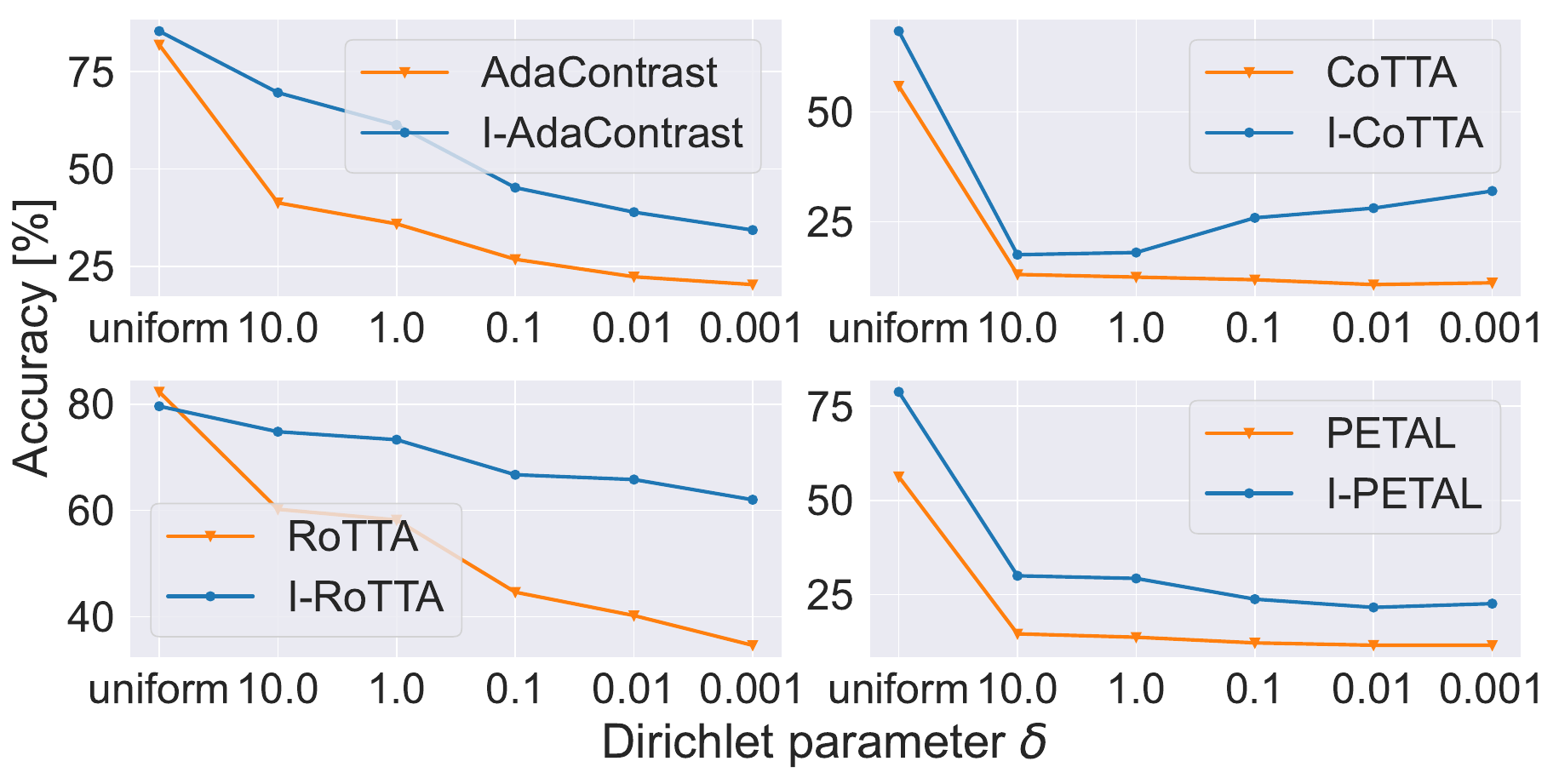}
    \captionof{figure}{Classification accuracy on CIFAR10-C (L). Samples are sorted by class for different levels of correlation, by varying the Dirichlet concentration parameter $\delta$.}
    \label{fig:correlation}

    \vspace{2em}

    \includegraphics[width=0.99\linewidth,trim={0 39 0 0},clip]{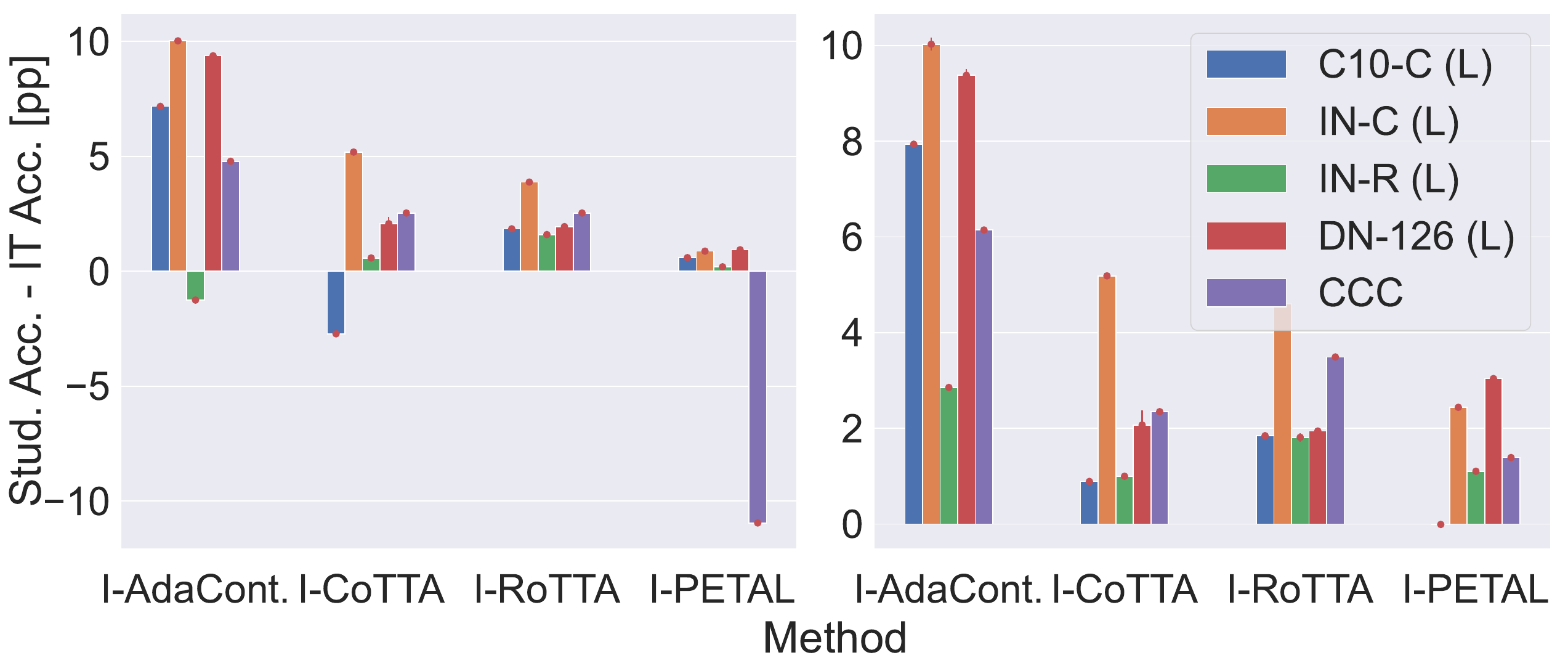}
    \caption{Accuracy difference between student and intransigent teacher in percentage points, with default learning rate values \textbf{(left)} and after parameter search \textbf{(right)}.
    }
    \label{fig:student_vs_teacher}
\end{wrapfigure}

\noindent \textbf{Intransigent teachers work across different architectures.}
We verify the architectural universality of IT on \mbox{CIFAR10-C (L)} and \mbox{ImageNet-C (L)} using various network families (e.g., ViT, ConvNeXt). 
It should be noted that the RN26GN, ViT-B16, SViT-T, and CNX-tiny architectures do not include batch normalization layers, so IT's performance is not impacted by statistics recalibration.
Because default learning rates are architecture-specific and difficult to set without target labels in TTA, we explore an upper-bound scenario where the learning rate is tuned via an Oracle for both the baseline and IT-augmented methods (Tab.~\ref{tab:arch}). 
This controlled comparison isolates architectural robustness by providing an optimal hyperparameter setting for less stability-oriented TTA methods. Under these conditions, IT consistently prevents catastrophic collapse and either improves upon or maintains competitive baseline accuracy across the tested architectures.
The advantages of IT become even more pronounced in a more realistic evaluation setup, where the learning rate is tuned exclusively for the original methods on standard-length benchmarks and directly reused for IT-augmented versions (see Tab.~\ref{tab:arch_defseq_tuned} in the Appendix). This demonstrates IT's robustness to suboptimal, baseline-specific hyperparameters. 
While IT-augmented CoTTA and PETAL underperform their base counterparts specifically on ResNeXt50, both still exceed the source model accuracy, resulting in successful adaptation. 
We attribute this occasional underperformance to the plasticity bottleneck, discussed in Sec.~\ref{sec:long}.
Further ablations in the Appendix confirm that IT maintains or improves robustness even when the source model's initial quality is artificially degraded with noise (Sec.~\ref{sec:weaker_source}).

\noindent \textbf{On robustness to learning rate selection.}
In Table~\ref{tab:lr_sens}, we verify the learning rate sensitivity. Default AdaContrast shows robustness when $10\times$ learning rate is used, but collapses with a $50\times$ learning rate on \mbox{ImageNet-C}. IT allows increased robustness in all settings and achieves a non-collapsed solution, even for the highest learning rate. CoTTA lacks learning rate robustness, which is mitigated by our IT extension. Our strategy improves RoTTA's performance on \mbox{CIFAR10-C}. However, it rendered our approach almost ineffective on \mbox{ImageNet-C}. Overall, IT does not fully prevent the collapse when significantly altering the learning rate, but seems to promote a better-performing parameter space.
Further analyses of hyperparameter tuning complexity (using Oracle and Transfer setups) are detailed in Appendix Sec.~\ref{sec:add_exp/lr_tune}, and we show that our observations extend to the semantic segmentation task in Appendix Sec.~\ref{sec:semantic_segmentation}.

\begin{table}[t]
    \begin{minipage}[t]{0.49\textwidth}
        \caption{Classification accuracy [\%] %for long scenarios on CIFAR10-C and ImageNet-C
with different neural network architectures. Superscript indicates improvements over the baseline. The learning rate is adjusted using an Oracle on long (L) benchmarks.} % for both original and I- ones.}
\centering
\resizebox{0.95\linewidth}{!}{
\begin{tabular}{l@{\hskip 0.3in}c@{\hskip 0.3in}cccc}
\toprule
& \textbf{C10-C (L)} &  \multicolumn{4}{c}{\textbf{IN-C (L)}} \\
& RN26GN & RNXt-50 & ViT-B16 & SViT-T & CNXt tiny \\ 
\midrule
Source & 67.3 & 21.1 & 39.8 & 28.3 & 29.1 \\

AdaCont. & 75.7 & 38.8 & 41.5 & 30.6 & 33.4 \\
I-AdaCont. 
& 79.6\textsuperscript{\textcolor{greenx}{+3.9}} 
& 42.7\textsuperscript{\textcolor{greenx}{+3.9}} 
& 43.5\textsuperscript{\textcolor{greenx}{+2.0}} 
& 30.9\textsuperscript{\textcolor{greenx}{+0.3}} 
& 32.5\textsuperscript{\textcolor{redx}{-0.9}} 
\\

CoTTA & 57.0 & 42.1 & 41.7 & 28.4 & 29.1 \\
I-CoTTA 
& 67.3\textsuperscript{\textcolor{greenx}{+10.3}}
& 38.3\textsuperscript{\textcolor{redx}{-3.8}} 
& 40.7\textsuperscript{\textcolor{redx}{-1.0}} 
& 28.9\textsuperscript{\textcolor{greenx}{+0.5}}
& 30.5\textsuperscript{\textcolor{greenx}{+1.4}}
\\

RoTTA & 70.2 & 35.6 & 40.6 & 28.8 & 29.0 \\
I-RoTTA 
& 72.5\textsuperscript{\textcolor{greenx}{+2.3}} 
& 36.2\textsuperscript{\textcolor{greenx}{+0.6}} 
& 42.9\textsuperscript{\textcolor{greenx}{+2.3}} 
& 28.9\textsuperscript{\textcolor{greenx}{+0.1}} 
& 29.7\textsuperscript{\textcolor{greenx}{+0.7}} 
\\

PETAL & 24.8 & 40.6 & 42.8 & 23.6 & 26.5 \\
I-PETAL
& 60.8\textsuperscript{\textcolor{greenx}{+36.0}} 
& 35.4\textsuperscript{\textcolor{redx}{-5.2}} 
& 41.1\textsuperscript{\textcolor{redx}{-1.7}} 
& 27.0\textsuperscript{\textcolor{greenx}{+3.4}} 
& 28.1\textsuperscript{\textcolor{greenx}{+1.6}} 
\\

\bottomrule
\end{tabular}
}
\label{tab:arch}

    \end{minipage}
    \hfill
    \begin{minipage}[t]{0.49\textwidth}
        \caption{Classification accuracy [\%] for long (L) scenarios on CIFAR10-C and ImageNet-C with $1\times$, $10\times$, $50\times$ learning rate (LR) scaling. Superscript indicates improvements over the baseline.
}
\centering
\resizebox{\linewidth}{!}{
\begin{tabular}{l@{\hskip 0.2in}ccc@{\hskip 0.2in}ccc}
\toprule
& \multicolumn{3}{@{\hskip -0.1in}c}{\textbf{C10-C (L)}} & \multicolumn{3}{c}{\textbf{IN-C (L)}}  \\
& $1\times$LR & $10\times$LR & $50\times$LR & $1\times$LR & $10\times$LR & $50\times$LR \\
\midrule
Source & \multicolumn{3}{@{\hskip -0.2in}c}{56.5} & \multicolumn{3}{c}{18.0} \\

AdaCont. & 81.8 & 83.6 & 80.4 & 18.8 & 19.6 & 12.4 \\
I-AdaCont.
& 85.4\textsuperscript{\textcolor{greenx}{+3.6}}
& 86.1\textsuperscript{\textcolor{greenx}{+2.5}} 
& 85.2\textsuperscript{\textcolor{greenx}{+4.8}} 
& 40.4\textsuperscript{\textcolor{greenx}{+21.6}} 
& 36.3\textsuperscript{\textcolor{greenx}{+16.7}} 
& 32.7\textsuperscript{\textcolor{greenx}{+20.3}} \\

CoTTA
& 56.0 & 10.3 & 10.1 & 52.8 & 4.1 & 0.1 \\
I-CoTTA 
& 68.3\textsuperscript{\textcolor{greenx}{+12.3}}  
& 51.9\textsuperscript{\textcolor{greenx}{+41.6}} 
& 41.4\textsuperscript{\textcolor{greenx}{+31.3}} 
& 35.4\textsuperscript{\textcolor{redx}{-17.4}} 
& 25.7\textsuperscript{\textcolor{greenx}{+21.6}} 
& 17.3\textsuperscript{\textcolor{greenx}{+17.2}} 
\\

RoTTA & 82.3 & 80.6 & 28.9 & 13.2 & 0.2 & 0.1 \\
I-RoTTA 
& 79.7\textsuperscript{\textcolor{redx}{-2.6}} 
& 77.9\textsuperscript{\textcolor{redx}{-2.7}} 
& 70.6\textsuperscript{\textcolor{greenx}{+41.7}} 
& 32.9\textsuperscript{\textcolor{greenx}{+19.7}} 
& 4.1\textsuperscript{\textcolor{greenx}{+3.9}} 
& 2.1\textsuperscript{\textcolor{greenx}{+2.0}}
\\

PETAL & 56.3 & 15.9 & 18.0 & 36.8 & 0.4 & 0.1 \\
I-PETAL 
& 78.8\textsuperscript{\textcolor{greenx}{+22.5}} 
& 59.7\textsuperscript{\textcolor{greenx}{+43.8}} 
& 35.8\textsuperscript{\textcolor{greenx}{+17.8}} 
& 32.3\textsuperscript{\textcolor{redx}{-4.5}} 
& 36.2\textsuperscript{\textcolor{greenx}{+35.8}} 
& 8.1\textsuperscript{\textcolor{greenx}{+8.0}}
\\

\bottomrule
\end{tabular}
}
\label{tab:lr_sens}

    \end{minipage}
    \vspace{-1em}
\end{table}

\noindent \textbf{Effects varying temporal correlation.}
Figure~\ref{fig:correlation} illustrates the impact of varying degrees of class temporal correlation~\citep{NOTE} on methods enhanced with IT. The analysis reveals that as temporal correlation increases, IT-enhanced methods consistently outperform their original versions, demonstrating improved robustness in these scenarios, and even when the original method achieved better results on uniform class distribution (RoTTA). However, it should be noted that while IT-enhanced methods show improved performance, they are also negatively impacted by the effects of correlation. The degree of improvement varies across different experimental settings.

\begin{figure*}[t!]
  \centering
  \begin{minipage}{0.49\textwidth}
    \centering
    \includegraphics[width=0.8\textwidth]{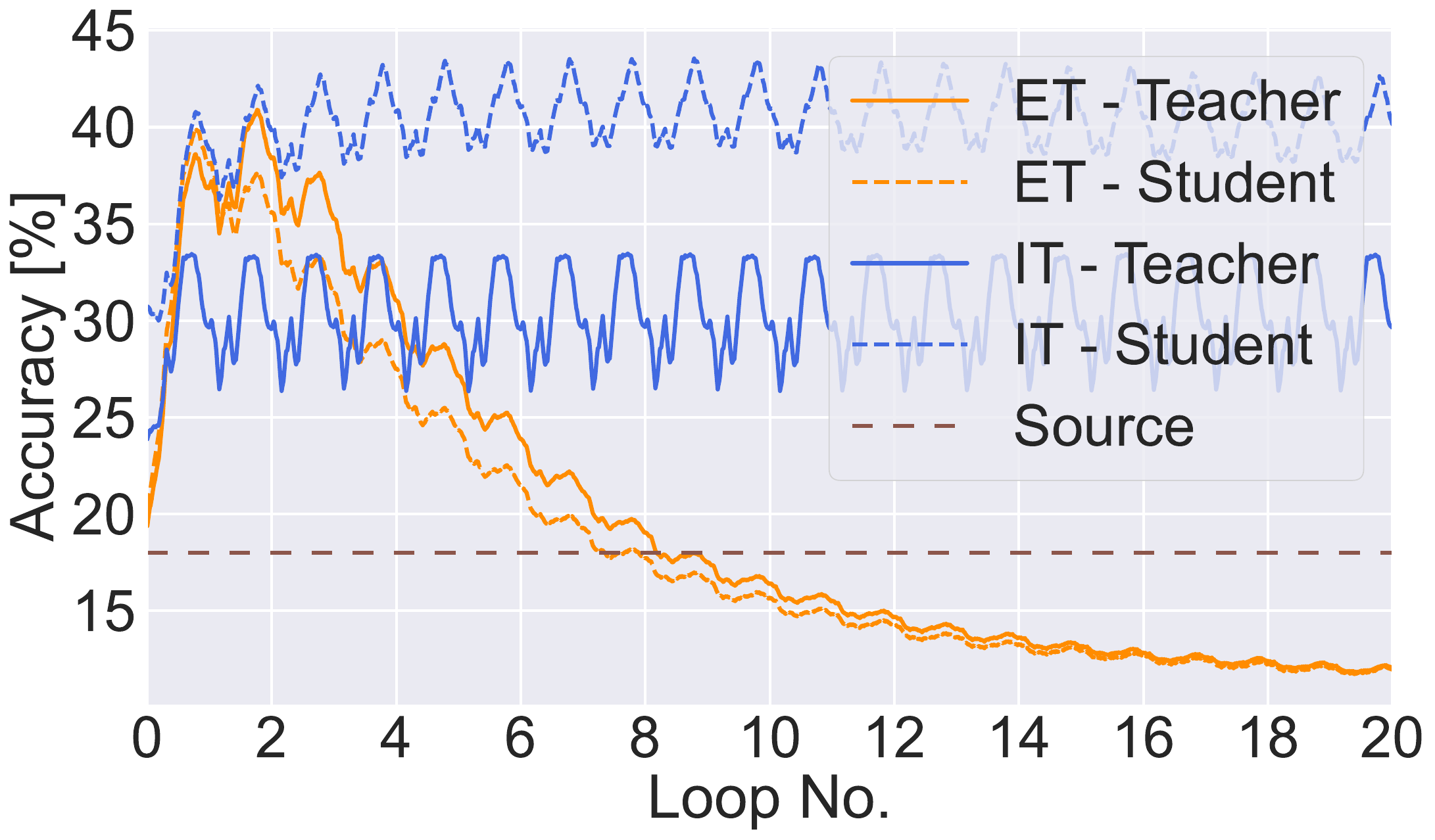}
  \end{minipage}
  \hfill
  \begin{minipage}{0.49\textwidth}
    \centering
    \includegraphics[width=0.8\textwidth]{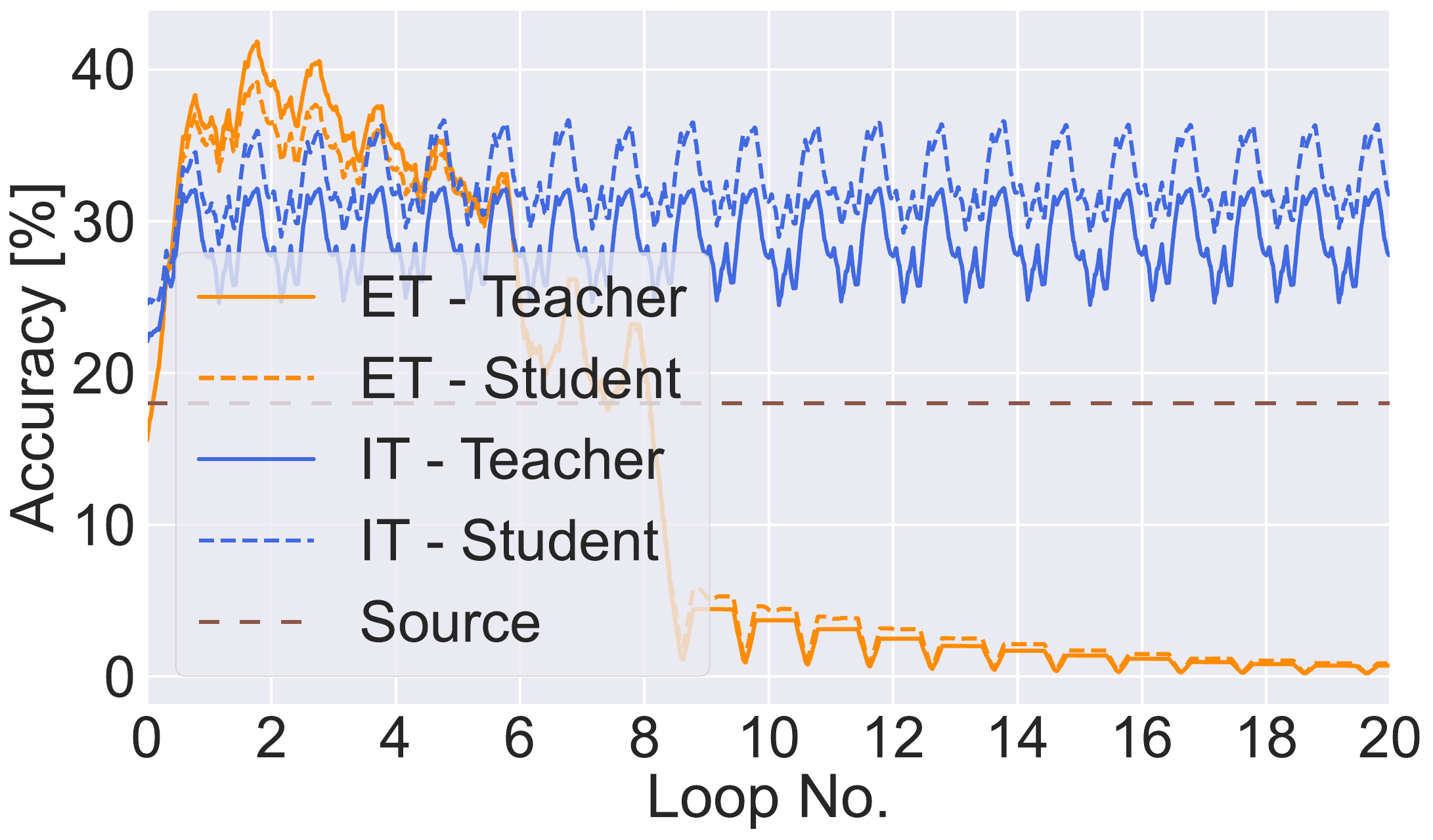}
  \end{minipage}
    \caption{Per-batch accuracy on repeated ImageNet-C (20 loops) using ResNet50. AdaContrast \textbf{(left)} and RoTTA \textbf{(right)} are compared with an EMA teacher (ET, orange) and our intransigent teacher (IT, blue), both for teacher (solid) and student (dashed). The EMA teacher visibly suffers from error accumulation as the sequence gets longer. After the 3rd loop, IT manages to avoid model collapse and, surprisingly, the student performs significantly better.}
    \label{fig:old_teaser}
\end{figure*}

\noindent \textbf{Are students better than intransigent teachers?}
When paired with intransigent teachers, student models often outperform their teachers, whereas EMA keeps teacher-student performance tightly coupled (Fig.~\ref{fig:old_teaser}). 
As shown in Figure~\ref{fig:student_vs_teacher}, IT enables the student to exceed teacher average accuracy in nearly all configurations, with only two exceptions: I-AdaContrast on ImageNet-R (L) and I-CoTTA on CIFAR10-C (L). In those cases, IT-induced stability limits performance decline but is insufficient to prevent degradation entirely. However, when the learning rate is optimized via an Oracle, students outperform teachers in all cases (Fig.~\ref{fig:student_vs_teacher}, right).
Per-batch accuracy plots for remaining method--dataset combinations are presented in the Appendix (Figs.~\ref{fig:imagenet_results} and~\ref{fig:cifar_results}).

\begin{figure*}[t!]
  \centering
  \includegraphics[width=0.99\linewidth]{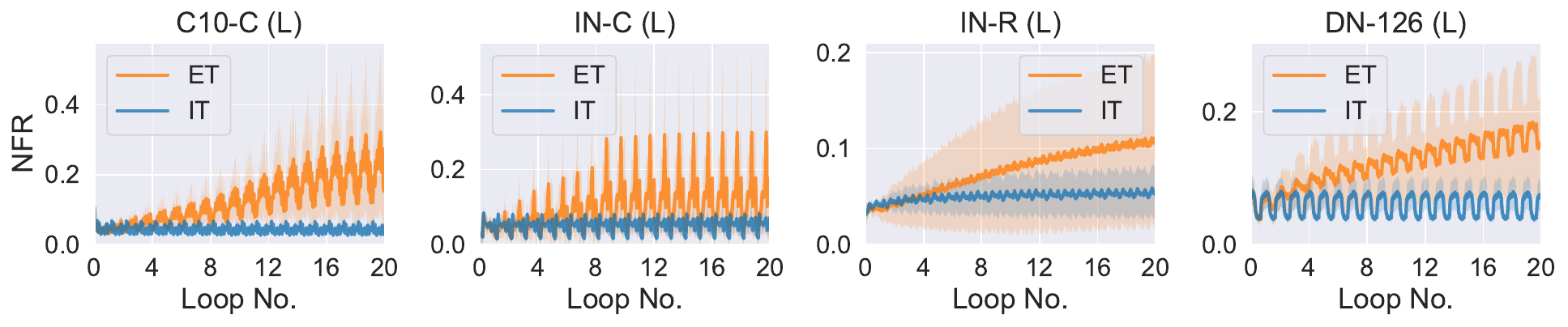}
  \caption{Per-batch Negative Flips Rate (NFR) of the \textbf{student} model averaged over AdaContrast, CoTTA, RoTTA and PETAL methods (ET) on each dataset, along with their IT-enhanced versions (IT). The shaded regions represents $\pm$ standard deviation illustrating the variability across methods.}
  \label{fig:negative_flips_averaged}
\end{figure*}

\noindent \textbf{Evaluating the error accumulation.}
\label{seq:negative_flips}
We hypothesize that the IT-enhanced methods' success in longer scenarios stems from their ability to mitigate error accumulation. We attempt to compare the degree of accumulated error between methods and employ the Negative Flip Rate (NFR) as defined in~\citet{yan2021positive}. The NFR quantifies the fraction of samples initially predicted correctly by the source model but incorrectly by the adapted model. This metric captures degradation of the initial model's knowledge, the direct effect of error accumulation.
In Figure~\ref{fig:negative_flips_averaged}, we show per-batch NFR of the student model averaged over AdaContrast, CoTTA, RoTTA, and PETAL methods for each benchmark. For specific examples, see the Appendix~(Fig.~\ref{fig:negative_flips}).
The NFR for IT-enhanced methods remains consistently stable, showing no increasing trend across any benchmark. It suggests that the fixed weights of the teacher prevent the student from accumulating errors, thereby preserving the initial knowledge. In contrast, the original methods exhibit increasing NFR in most experiments, indicating that the adaptation process degrades the model's knowledge, which suggests error accumulation. In some experiments, the original methods exhibit a stable, low NFR, indicating reliable adaptation that correctly improves the model, which also correlates with higher accuracy in Table~\ref{tab:long}.
Complementary to these findings, we observe that IT also prevents the internal degradation of the model's feature space. As detailed in Appendix~Sec.~\ref{sec:effective_rank}, tracking the effective rank of the student's representations reveals that baseline collapses correlate with a severe decrease in feature diversity. By anchoring the adaptation process, IT consistently maintains high effective rank and prevents dimensional collapse.
Overall, IT-enhanced methods offer more consistent behavior and robustness across benchmarks.

% --------------------------------------------------------------
\section{Conclusions}
In this work, we explore the well-known strategy of using an exponential moving average teacher for TTA. Although effective for common sequence lengths, this strategy fails to mitigate long-term error accumulation, which can lead to model collapse. To address the resulting plasticity-stability trade-off, we propose an intransigent teacher strategy compatible with current state-of-the-art methods. Our approach often prevents model collapse across diverse experimental setups and improves robustness to hyperparameter choice without requiring manual tuning. Empirical results demonstrate that the intransigent teacher halts the accumulation of errors and helps to maintain high feature diversity. These findings suggest that TTA evaluation must encompass varied sequence lengths to ensure reliability. Our method provides a robust, simple baseline for future long-term adaptation research.

\smallskip
\noindent\textbf{Discussion and limitations.} 
While the intransigent teacher strategy proved highly effective across various scenarios, it can result in slower adaptation in certain cases due to the teacher's fixed behavior. To address this issue, we also present results where the teacher model was allowed to adapt for a fixed number of initial steps (Appendix~Sec.~\ref{sec:add_exp/adaptive_beta}). Further exploration of this approach is left for future work.
Furthermore, we note that by adding the intransigent teacher to existing methods, we inherit their limitations.

\subsubsection*{Acknowledgments}
This research was funded in whole or in part by National Science Centre, Poland, grants no 2024/53/N/ST6/03156 and 2023/51/D/ST6/02846. 
We gratefully acknowledge Polish high-performance computing infrastructure PLGrid (HPC Center: ACK Cyfronet AGH) for providing computer facilities and support within computational grants no. PLG/2025/018634, PLG/2025/018644, and PLG/2026/019363.
This research was funded in whole or in part by the Austrian Science Fund (FWF) 10.55776/COE12.
Bart{\l}omiej Twardowski acknowledges the grant RYC2021-032765-I.

\bibliography{bib}
\bibliographystyle{collas2026_conference}

\renewcommand{\thefigure}{A.\arabic{figure}}
\setcounter{figure}{0}
\renewcommand{\thetable}{A.\arabic{table}}
\setcounter{table}{0}

\clearpage

\appendix
\section*{Appendix}
This appendix provides comprehensive evidence and analysis to support the findings presented in the main paper.
Sec.~\ref{sec:add_details} details the experimental framework, including the setup (Sec.~\ref{sec:app-exp-setup}), baseline descriptions and implementations (Sec.~\ref{sec:baseline_descr}, \ref{sec:add_details/base_impl_det}), and specific details for the intransigence experiments (Sec.~\ref{sec:add_details/sec3_it}) and the MEMO results on CCC benchmark (Sec.~\ref{sec:add_details/memo_ccc}). It also covers the negative flip rate (NFR) calculation (Sec.~\ref{sec:add_details/nfr}) and computational requirements (Sec.~\ref{sec:add_details/compute}).

Sec.~\ref{sec:add_exp} presents several additional experiments to better understand the described problem and the limitations of the proposed technique.
These include evaluation of feature representation strength (Sec.~\ref{sec:effective_rank}), semantic segmentation experiment (Sec.~\ref{sec:semantic_segmentation}), analysis of the dependency on source model performance (Sec.~\ref{sec:weaker_source}), experiment with tuning the learning rate value for long scenarios (Sec.~\ref{sec:add_exp/lr_tune}), wall-clock time comparison (Sec.~\ref{sec:walltime}), discussion on single image adaptation methods (Sec.~\ref{sec:single_image_methods}), results on common-length benchmarks (Sec.~\ref{sec:short}), an investigation of the potential of the adaptive $\beta$ value for teacher momentum (Sec.~\ref{sec:add_exp/adaptive_beta}), an analysis of student-only adaptation performance (Sec.~\ref{sec:add_exp/only_student}), 
an extended experiment showing the effects of intransigence for over 100 loops with the CoTTA method (Sec.~\ref{sec:add_exp/it_amount}), discussion on model reset mechanism (Sec.~\ref{sec:hyperparam/reset}), the novelty clarification with regard to RDumb (Sec.~\ref{sec:rdumb}),
augmentation count ablation for CoTTA and PETAL methods~\ref{sec:aug_abl}, 
decomposition of the IT student’s gain into BN statistics recalibration and teacher distillation~\ref{sec:decomp_stats_distill},
and experiments with fixed batch normalization statistics~\ref{sec:add_exp/fixed_bn}.
We also include the remaining results of the experiments mentioned in the main paper (Sec.~\ref{sec:add_plots}).

\section{Additional Details}
\label{sec:add_details}

\subsection{Experimental setup}
\label{sec:app-exp-setup}
\noindent \textbf{Benchmarks}. 
Our experiments include popular corruption benchmarks \mbox{CIFAR10-C} (C10-C) and \mbox{ImageNet-C} (IN-C)~\citep{hendrycks2019robustness}, training on clean CIFAR10~\citep{cifar} or ImageNet~\citep{imagenet} images and testing on corrupted images with 15 corruption types at 5 severity levels. We adopt the setup from~\citet{COTTA, eata, dobler2023CVPRmeanteacher}, using the standard corruption sequence at the highest severity. The adaptation on natural domain shifts is tested utilizing DomainNet-126 (DN-126)~\citep{peng2019moment} and ImageNet-R (IN-R)~\citep{hendrycks2021many} datasets. For DN-126, we use the real domain as the source data and conduct experiments on the remaining domains. The clean ImageNet dataset serves as the source data for experiments on ImageNet-R. Our goal is to focus on very long adaptation sequences to evaluate the methods in terms of stability during long-time operation. For that reason, we repeat the standard test sequences of benchmarks described above 20 times. We call this setup a Long~(L)~scenario. Additionally, we take advantage of the CCC~\citep{rdumb} benchmark, which has a very high sequence length without repetitions. We use a CCC-Medium sequence with a 1k transition speed, which consists of 7,500,000 images.
For experiments with semantic segmentation, we utilize the \mbox{CarlaTTA} dataset~\citep{carlatta}. The source data consists of images from the clear domain, i.e., images captured during the day in clear weather. For test data, we use the \textit{day2night} sequence, which goes from the clear domain, gradually transitioning to night. 
Consistent with image classification experiments, we used the same approach to create the Long~(L)~scenario.

\noindent \textbf{Architectures.} Consistent with previous works~\citep{COTTA, MarsdenD024roid}, we use \mbox{WideResNet-28}~\citep{zagoruyko2016wide} models with pre-trained weights from the \textit{RobustBench}~\citep{croce2021robustbench} model zoo for the main experiments on \mbox{CIFAR10-C}. Similarly, the tests on ImageNet-based benchmarks and DomainNet-126 employ the ResNet50 network with weights sourced from the same model zoo or those provided by~\citet{MarsdenD024roid} for DomainNet-126. Additional experiments in Sec.~\ref{sec:analysis} are carried out with ResNet-26GN (RN26GN)~\citep{WuH18GN}, ResNeXt-50~32x4d~(RNXt-50)~\citep{xie2017aggregated}, \mbox{ViT-B16}~\citep{DosovitskiyViT}, \mbox{SwinViT-T} (SViT-T)~\citep{liu2021swin} and ConvNeXt tiny (CNXt tiny)~\citep{convnext} architectures. Weights for RN26GN are taken from~\citet{ZhangLF22Memo}, as in~\citet{MarsdenD024roid}. The \textit{Torchvision}~\citep{torchvision2016} library is used to initialize the rest of the mentioned models.
For semantic segmentation, following~\citet{carlatta}, we utilize the DeepLab-V2~\citep{chen2017deeplab} architecture with ResNet-101 backbone. We use the pre-trained source checkpoint provided~by~\citet{carlatta}.

\noindent \textbf{Implementation details.} As a testbed for experiments, we adopt the framework from~\citet{MarsdenD024roid} and use the parameters reported in the respective original papers. We use parameters from common sequence length settings while testing on the long~(L)~scenarios, unless otherwise stated. We utilize two batch sizes for image classification: a standard value of 64, as in multiple previous works~\citep{MarsdenD024roid, eata, rdumb, TENT}, and a lower one (equal to 10), resulting in a more challenging setup with more model updates. When running experiments on a smaller batch size, the learning rate is decreased accordingly.
In the case of semantic segmentation, we follow the usual batch size of 1~\citep{carlatta}. In-depth implementation details regarding the baselines are provided in Sec.~\ref{sec:add_details/base_impl_det}).

\subsection{Descriptions of baselines extended with the Intransigent Teacher}
\label{sec:baseline_descr}

\noindent \textbf{AdaContrast} conducts weight updates using a three-part loss function: cross-entropy loss, diversity regularization, and contrastive loss. Pseudo-labels are refined by keeping a buffer of previous image features and their predictions. Refined predictions are based on the nearest neighbors of the current feature within the buffer. Statistics in batch normalization layers are updated with EMA. 

\noindent \textbf{CoTTA} updates the student model by minimizing the cross-entropy consistency between the teacher and the student predictions. Depending on the prediction confidence, the pseudo-labels are the result of averaging predictions on multiple, differently augmented images. At each iteration, there is a small probability for each of the student's weights to be reset to the source pre-trained value. It calculates batch normalization statistics on a per-batch basis. 

\noindent \textbf{RoTTA} keeps the class-balanced memory buffer of images and uses it to perform the optimization in constant intervals. The loss is weighted based on how long the sample has been stored. Cross-entropy-based consistency between the student and teacher models is utilized for a loss function. Batch normalization statistics are updated via EMA.

\noindent \textbf{PETAL} is similar to CoTTA, but it enhances the learning objective by the regularizer term based on a posterior distribution over a source model weights and a data-dependent prior. Moreover, it improves CoTTA's stochastic model reset scheme with the Fisher Information Matrix. 

\subsection{Baselines implementation details}
\label{sec:add_details/base_impl_det}
The experiments are conducted by adapting the code repository of previous test-time adaptation works~\citep{MarsdenD024roid, dobler2023CVPRmeanteacher}, which provide the implementation of all evaluated state-of-the-art methods, except PETAL and PeTTA. We integrated its original implementation into this unified codebase ourselves. In terms of hyperparameters, we followed the typical implementation for test of batch size of 64. 

TENT~\citep{TENT}, EATA~\citep{eata}, SAR~\citep{SAR}, and RDUMB~\citep{rdumb} use Adam optimizer with a learning rate of 0.001 for \mbox{CIFAR10-C} and SGD optimizer with a learning rate of 0.00025 for other benchmarks. AdaContrast~\citep{AdaContrast} utilizes an SGD optimizer with a learning rate set to 0.0002 for all of the benchmarks. CoTTA~\citep{COTTA} uses Adam optimizer with a learning rate of 0.001 for \mbox{CIFAR10-C} and SGD with a learning rate of 0.01 for the rest of the benchmarks. Adam optimizer with a learning rate set to 0.001 is used by RoTTA~\citep{rotta} for all of the tested datasets. MEMO~\citep{ZhangLF22Memo} uses an SGD optimizer with a learning rate of 0.005 for CIFAR10-C and 0.00025 for other datasets. PETAL~\citep{petal} in the original paper uses Adam optimizer with a learning rate of 0.001 for CIFAR10-C and SGD with a learning rate of 0.01 for other datasets. However, since we often experienced poor performance using these values on long scenarios, we utilized 10 times lower learning rates. Due to PETAL's regularizer term becoming unstable at higher learning rates, which leads to model corruption, we excluded PETAL from the learning rate scaling experiments shown in Tab.~\ref{tab:lr_sens}. This decision avoided presenting potentially compromised results.
PeTTA~\citep{hoang2024persistent} use Adam optimizer with learning rate of 0.001 for all of the benchmarks.
The learning rate used in experiments with batch size set to 10 was adjusted accordingly by scaling it linearly.

CoTTA~\citep{COTTA} and PETAL~\citep{petal} methods update the student network using a consistency loss between the student and teacher. If the prediction confidence of the source model is below a certain threshold, the teacher's predictions are averaged over 32 different augmentations of the image which adds 31 additional forward operations per batch. It creates a significant computation overhead and causes the methods to be significantly slower, compared to other state-of-the-art methods. It is especially problematic for long adaptation scenarios, which make the main part of our experiments. Our ablations indicate that using a single augmentation does not alter the results notably~(Sec.~\ref{sec:aug_abl}). Therefore, for the ease of experimentation, we reduce the number of augmentations to~1.

\subsection{Details on intransigent teacher experiment from Section~\ref{sec:intrans_exp}}
\label{sec:add_details/sec3_it}
Our preliminary intransigent teacher experiment is conducted on ImageNet-C, a long ImageNet-C scenario ($20\times$ loop of standard ImageNet-C testing sequence), and CCC. We utilize the same model as for our main experiments on ImageNet-based benchmarks - ResNet50 with pre-trained weights from the \textit{RobustBench}~\citep{croce2021robustbench} model zoo. We use a single loss within the teacher-student framework for the model adaptation during test time -- either Consistency from CoTTA or Contrastive from AdaContrast. Any other components of the mentioned state-of-the-art methods are not included. Batch normalization statistics are recalculated for each batch. For both of the tested approaches, we use the SGD optimizer with a learning rate of 0.00025.

\subsection{Details on MEMO results on CCC benchmark from Tab.~\ref{tab:long}}
\label{sec:add_details/memo_ccc}
The result is based on the first 623,000 images of the benchmark, providing an initial estimate of the method's accuracy. However, due to the benchmark's extensive size (7,500,000 images) and the method's requirement for a batch size of 1, we cannot complete the full experiment in a reasonable amount of time. We estimate that processing the entire dataset requires approximately 972 hours on a single NVIDIA GeForce RTX 4080 GPU. This substantial time requirement underscores the method's significant computational inefficiency.

\subsection{Details on Negative Flip Rate}
\label{sec:add_details/nfr}
In Sec.~\ref{sec:analysis}, we attempt to compare the degree of accumulated error between methods and employ the Negative Flip Rate (NFR) as defined in~\citet{yan2021positive}. The NFR quantifies the fraction of samples initially predicted correctly by the source model but incorrectly by the adapted model, calculated as:
\begin{equation} 
    \text{NFR} = \frac{1}{N} \sum_{i=1}^{N} \mathbf{1}(\hat{y}^0_i = y_i, \ \hat{y}_i \neq y_i)\ \,
\end{equation}
where $y_i$ is the ground truth label, $\hat{y}^0_i$ is the source model's prediction, $\hat{y}_i$ is the current model's prediction, and $\mathbf{1}(\cdot)$ is the indicator function.  This metric captures degradation of the initial model's knowledge, the direct effect of error accumulation.

\subsection{Compute details}
\label{sec:add_details/compute}
All experiments are conducted on a single GPU. We utilize either NVIDIA A100 with 40GB of memory or NVIDIA GeForce RTX 4080 with 16GB of memory. The execution time of the experiment greatly varied and was dependent on the dataset, scenario (standard or long), tested method, and batch size. The fastest experiments take about 30 minutes, whereas the longest last up to 36 hours.

\section{Additional Experiments}
\label{sec:add_exp}

\subsection{Evaluation of feature representation strength.}
\label{sec:effective_rank}
To gain additional insight into why the IT works effectively, we assess changes in the representation strength of features extracted from the model's encoder during the adaptation process. 

We focus on measuring the diversity of the features and compute the effective rank~\citep{roy2007effective} of the feature representations during the adaptation process. We collect a domain-balanced set of $n$ ($n\!=\!150$) randomly sampled images from the adaptation sequence of a given benchmark. During TTA, we extract features for this entire set of images using the encoder of the adapted model and construct a feature matrix $A$. We then compute its singular values $\sigma_1, \cdots, \sigma_n$. Given the distribution of singular values, defined as $p_{i}\!=\!\frac{\sigma_i}{|\sigma|_1}$, we calculate the effective rank of feature representation matrix $A$ as follows:
\begin{equation}
    E\text{-rank}(A) = \exp(-\sum_{i=1}^{n}p_i\log(p_i)).
\end{equation}

\begin{figure*}[ht!]
\centering
\begin{minipage}{\textwidth}
\centering
\includegraphics[width=0.99\linewidth]{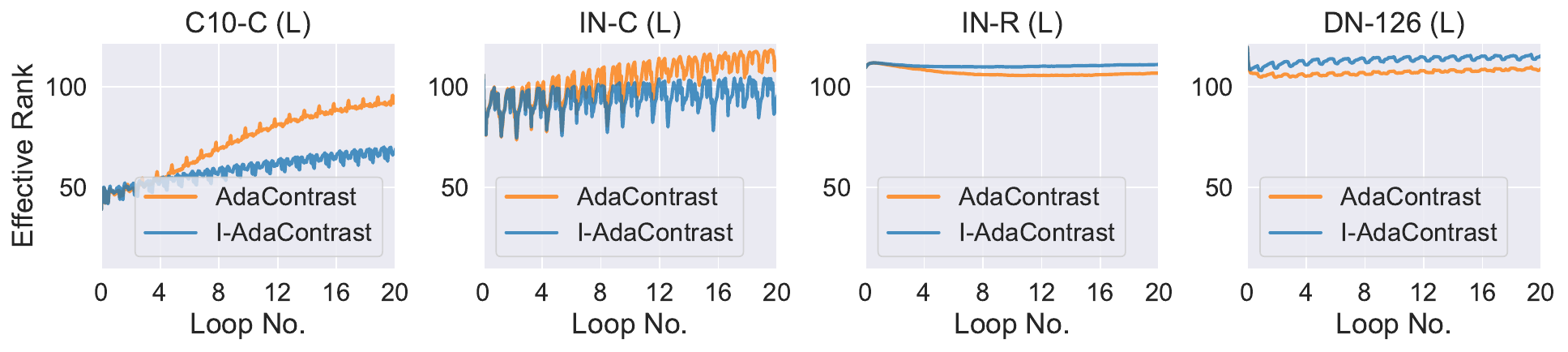}\\
\textbf{(a)} AdaContrast
\end{minipage}

\vspace{1em}  % space between images

\begin{minipage}{\textwidth}
\centering
\includegraphics[width=0.99\linewidth]{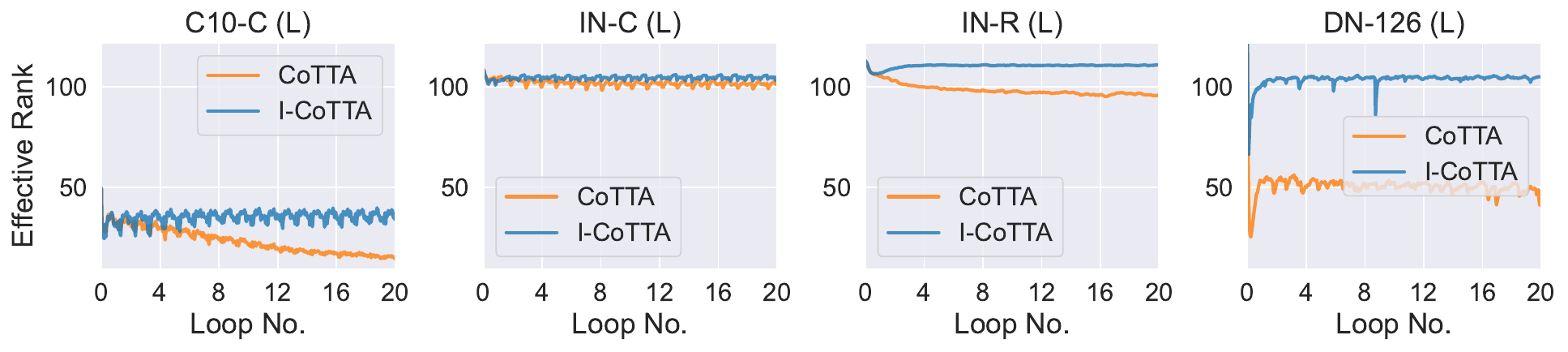}\\
\textbf{(b)} CoTTA
\end{minipage}

\vspace{1em}

\begin{minipage}{\textwidth}
\centering
\includegraphics[width=0.99\linewidth]{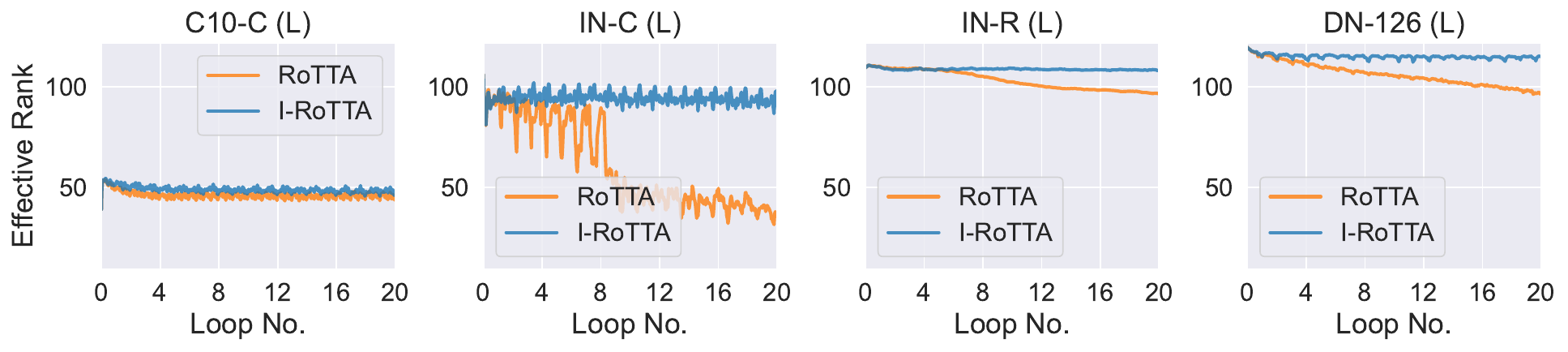}\\
\textbf{(c)} RoTTA
\end{minipage}

\vspace{1em}

\begin{minipage}{\textwidth}
\centering
\includegraphics[width=0.99\linewidth]{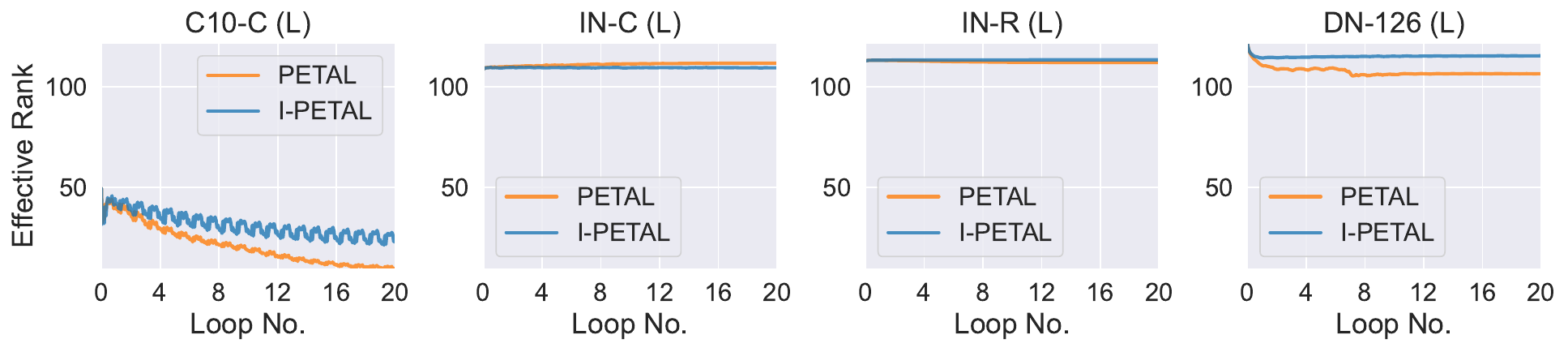}\\
\textbf{(d)} PETAL
\end{minipage}
  \caption{Effective rank values of the features extracted from the \textbf{student} model during the TTA every 50 batches on each benchmark using AdaContrast (\textbf{a}), CoTTA (\textbf{b}), RoTTA (\textbf{c}) and PETAL (\textbf{d}) methods and IT-enhanced version.
  }
  \label{fig:effective_rank}
\end{figure*}

We track the effective rank of the feature representations of the student model every 50 batches throughout the adaptation sequence. The representative results are presented in Fig.~\ref{fig:effective_rank}. 

We observe that a significant drop in effective rank (e.g., a two-fold decrease) correlates with a collapsed solution, where accuracy falls below that of the source model. This pattern is evident in four out of six collapse cases with a batch size of 64, as reported in Tab.~\ref{tab:long}: CoTTA and PETAL on C10-C (L), RoTTA on IN-C (L), and CoTTA on DN-126 (L). The use of IT helps maintain higher feature diversity, thereby preventing collapse. For the remaining collapse cases, the effective rank also exhibits a slight downward trend: AdaContrast on \mbox{IN-R (L)} and RoTTA on DN-126 (L).

\subsection{Semantic segmentation}
\label{sec:semantic_segmentation}
\begin{figure}[!b]
    \centering
    \includegraphics[width=0.5\linewidth]{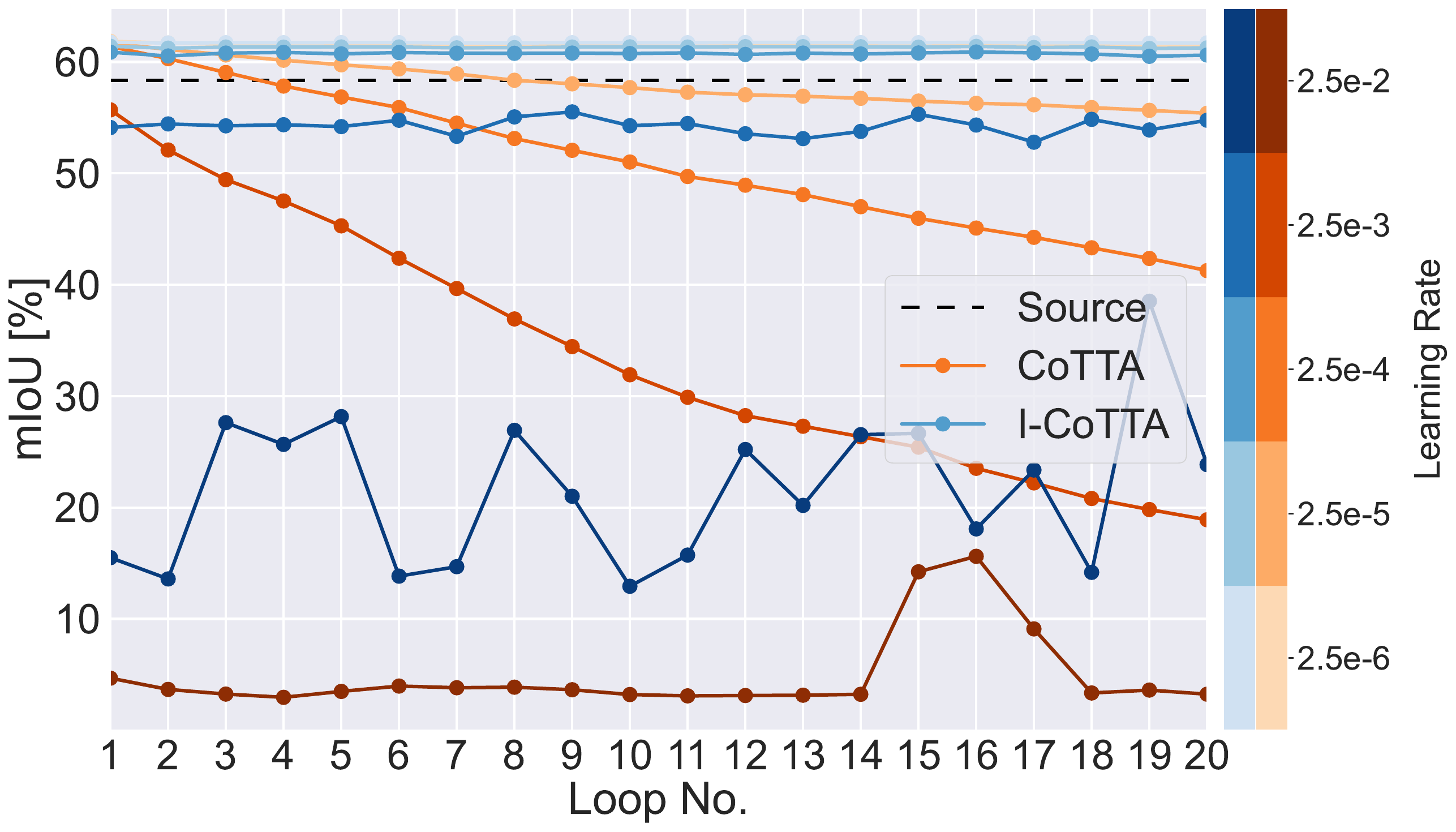}
    \caption{
    Segmentation results (mIoU [\%]) for each repeated \textit{day2night} sequence on CarlaTTA~\citep{carlatta}. Orange and blue colors indicate CoTTA and I-CoTTA, respectively. The darker the color, the higher the learning rate value used to achieve the result. The black dashed line indicates the source model performance as a reference.
    }
    \label{fig:segmentation}
\end{figure}

We evaluate a broader range of tasks and report the mean intersection-over-union (mIoU) results for the semantic segmentation experiment with different learning rate values across the entire long (L) scenario in Tab.~\ref{tab:segmentation}, with results for each loop presented separately in Fig.~\ref{fig:segmentation}.

We focus exclusively on the CoTTA method, as it is the only baseline developed specifically for this task by its authors. The results demonstrate that our initial observations extend to the semantic segmentation task: CoTTA, which utilizes EMA teacher, is highly sensitive to learning rate selection and degrades model performance over time for all chosen values, including the default for the standard-length sequence (2.5e-4), except for the lowest value (where its plot line is occluded by the I-CoTTA plot). In contrast, I-CoTTA exhibits greater robustness to learning rate selection, and IT effectively mitigates the error accumulation issue.

\begin{table}[b!]
\caption{Average segmentation (mIoU [\%]) on the long \textit{day2night} sequence from CarlaTTA~\cite{carlatta} with different learning rate (\textbf{LR}) values. The superscript indicates the improvements over the baseline.}
\centering
\resizebox{0.55\linewidth}{!}{
\begin{tabular}{lccccc}
\toprule
\textbf{LR} & \textbf{2.5e-2} & \textbf{2.5e-3} & \textbf{2.5e-4} & \textbf{2.5e-5} & \textbf{2.5e-6} \\
\midrule
Source & \multicolumn{5}{c}{58.4} \\
CoTTA & 4.9 & 33.9 & 50.9 & 58.0 & \textbf{61.5} \\
I-CoTTA &
21.6\textsuperscript{\textcolor{greenx}{+16.7}} &
54.3\textsuperscript{\textcolor{greenx}{+20.4}} &
60.8\textsuperscript{\textcolor{greenx}{+9.9}} &
61.3\textsuperscript{\textcolor{greenx}{+3.3}} &
\textbf{61.7}\textsuperscript{\textcolor{greenx}{+0.2}} \\
\bottomrule
\end{tabular}
}
\label{tab:segmentation}
\end{table}

\begin{figure*}[b!]
    \centering
    \includegraphics[width=0.8\linewidth]{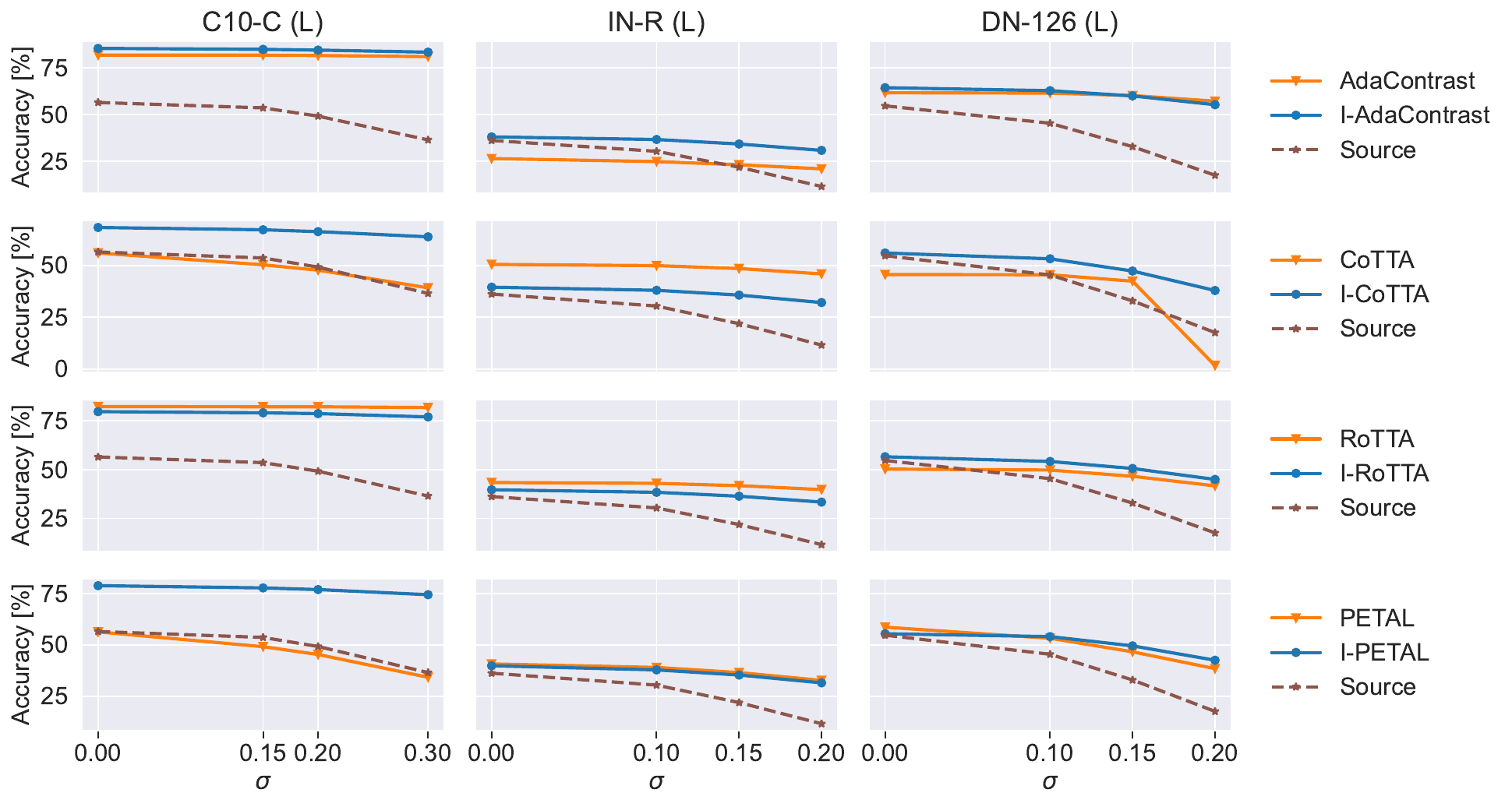}
    \caption{The overall average accuracy on each of the tested benchmarks of the original methods and IT-enhanced versions using source model weights perturbed using Gaussian noise with increasing standard deviation $\sigma$. The $\sigma\!=\!0$ indicates the performance of the model not perturbed by noise. The brown dashed line indicated the source model accuracy.}
    \label{fig:weaker_src_model}
\end{figure*}

\subsection{Dependency on source model performance}
\label{sec:weaker_source}
Since the intransigent teacher strategy fixes the teacher model's weights, its performance may heavily depend on the quality of the source model and could fail if the model is poorly suited to the target domain. To test this hypothesis, we evaluate adaptation performance using source models of varying quality. We generate these models by taking our default model weights and perturbing them with Gaussian noise of varying standard deviation, $\sigma$. For each layer, the noise's standard deviation is scaled by the mean value of that layer's weights. The average accuracies across the entire benchmark for each $\sigma$ are presented in Fig.~\ref{fig:weaker_src_model}. We compare the accuracy trends of the original methods with those enhanced by the IT strategy. The results indicate that, in most cases, weaker source models lead to poorer performance in the original baselines. \mbox{IT-enhanced} methods either follow the same trend as the original baselines (e.g., \mbox{I-AdaContrast} and \mbox{I-RoTTA} on \mbox{C10-C (L)}, \mbox{I-PETAL} and \mbox{I-RoTTA} on \mbox{IN-R (L)}, \mbox{I-AdaContrast} on \mbox{DN-126 (L)}) or exhibit greater robustness to weaker initial training (e.g., \mbox{I-CoTTA} and \mbox{I-PETAL} on \mbox{C10-C (L)}, \mbox{I-CoTTA} and \mbox{I-PETAL} on \mbox{DN-126 (L)}). In other words, IT either maintains the original method's sensitivity to source model quality or improves its robustness. We hypothesize that this behavior occurs because, even when the source model's predictions are less accurate, the fixed teacher model is prevented from further performance degradation and error accumulation due to the student's noisy adaptation, while still allowing the student model to update (via backpropagation) on the target data.

Our observations align with those from~\citet{zhao_2023_pitfallsTTA}, where the authors show that, in general, the performance of TTA methods is highly dependent on the quality of the initial source model.

\begin{table*}[hb!]
\caption{Classification accuracy [\%] for long scenarios with the learning rate \textbf{(LR)} parameter tuned. LR value \textbf{Default} means that the default LR value for the method was used. \textbf{Transfer IN-C} indicates that the LR is tuned utilizing the ImageNet-C benchmark with ground truth labels. In the \textbf{Oracle} rows the learning rate is chosen separately for each of the datasets with Oracle. 
The batch size is equal to 64.}
\centering
\resizebox{0.8\linewidth}{!}{
\begin{tabular}{l@{\hskip 0.5in}l@{\hskip 0.5in}cccc@{\hskip 0.5in}c}
\toprule
\textbf{Method} & \textbf{LR} & \textbf{C10-C (L)} & \textbf{IN-C (L)} & \textbf{IN-R (L)} & \textbf{DN-126 (L)} & \textbf{Avg.} \\ 
\midrule

\multirow{3}{*}{AdaCont.} 
& Default 
& 81.8 & 18.8 & 26.5 & 61.7 & 47.2 \\
& Transfer IN-C
& 81.2 & 36.1 & 40.8 & 59.7 & 54.5 \\
& Oracle 
& 82.6 & 36.1 & 40.8 & 62.7 & 55.6 \\
\multirow{3}{*}{I-AdaCont.} 
& Default 
& 85.4 & 40.4 & 38.2 & 64.4 & 57.1\textsuperscript{\textcolor{greenx}{+9.9}} \\
& Transfer IN-C 
& 85.4 & 40.4 & 38.2 & 64.4 & 57.1\textsuperscript{\textcolor{greenx}{+2.6}} \\
& Oracle 
& 86.2 & 40.4 & 42.3 & 64.4 & 58.3\textsuperscript{\textcolor{greenx}{+2.7}} \\

\midrule
\multirow{3}{*}{CoTTA} 
& Default 
& 56.0 & 52.8 & 50.5 & 45.6 & 51.2 \\
& Transfer IN-C
& 11.2 & 52.8 & 50.5 & 45.6 & 40.0 \\
& Oracle 
& 75.8 & 52.8 & 50.5 & 58.7 & 59.5 \\
\multirow{3}{*}{I-CoTTA} 
& Default 
& 68.4 & 35.4 & 39.6 & 56.8 & 50.1\textsuperscript{\textcolor{redx}{-1.1}} \\
& Transfer IN-C 
& 52.0 & 35.4 & 39.6 & 56.8 & 46.0\textsuperscript{\textcolor{greenx}{+6.0}} \\
& Oracle 
& 79.2 & 35.4 & 40.8 & 56.8 & 53.1\textsuperscript{\textcolor{redx}{-6.4}} \\

\midrule
\multirow{3}{*}{RoTTA} 
& Default 
& 82.3 & 13.2 & 43.4 & 50.3 & 47.3 \\
& Transfer IN-C
& 73.2 & 30.6 & 41.0 & 55.3 & 50.1 \\
& Oracle 
& 82.3 & 30.6 & 43.4 & 55.3 & 52.9 \\
\multirow{3}{*}{I-RoTTA} 
& Default 
& 79.6 & 32.7 & 39.7 & 57.2 & 52.3\textsuperscript{\textcolor{greenx}{+5.0}} \\
& Transfer IN-C 
& 79.3 & 33.3 & 39.9 & 57.2 & 52.4\textsuperscript{\textcolor{greenx}{+2.3}} \\
& Oracle 
& 79.6 & 33.3 & 39.9 & 57.2 & 52.5\textsuperscript{\textcolor{redx}{-0.4}} \\

\bottomrule
\end{tabular}
}
\label{tab:lr_tuning}
\end{table*}

\subsection{Tuning the learning rate value for long scenarios}
\label{sec:add_exp/lr_tune}
We investigate whether tuning the learning rate, arguably the most crucial hyperparameter, could enhance the performance of baseline methods in long adaptation scenarios. Following a realistic approach, we employ an Oracle technique on ImageNet-C (L) as a reference benchmark (inspired by~\citet{rusakSPEGBBB22}, we name it Transfer IN-C) and apply the selected learning rate across all datasets. We also present results when the learning rate is chosen separately for each of the datasets with an Oracle. Tab.~\ref{tab:lr_tuning} reports the complexity of hyperparameter optimization in test-time adaptation.

Our findings show the challenges of hyperparameter tuning. For instance, CoTTA achieves superior accuracy with its default learning rate compared to the version tuned on ImageNet-C. While both AdaContrast and RoTTA show improvements when using Transfer IN-C, our IT approach consistently outperforms these methods, even when specifically tuned for long-sequence adaptation on ImageNet-C. The results of Oracle, an (unrealistic) upper bound technique for learning rate tuning, suggest that optimal learning rates could potentially enable the original methods to outperform the IT version in some cases (e.g. CoTTA). However, this is not always the case. When tuned using Oracle, I-AdaContrast achieves superior accuracy compared to its original version, while Oracle-tuned I-RoTTA performs similarly to its original counterpart. These results underscore both the difficulty of hyperparameter selection and the robust performance of our IT method across varying conditions.

\begin{wraptable}{R}{0.4\textwidth}
\vspace{-1em}
% \begin{table}[b!]
\caption{Wall-clock time (seconds) for processing
10,000 images of CIFAR10-C on a single RTX 4080 GPU. The value in superscript indicates the performance improvements over the baseline.}
\centering
\resizebox{0.5\linewidth}{!}{
\begin{tabular}{lc}
\toprule
\textbf{Method} & \textbf{Time [s]} \\ 
\midrule
Source & 3.4 \\
MEMO & 508.4 \\
\midrule
AdaCont. & 25.3 \\
I-AdaCont. & 25.0\textsuperscript{\textcolor{greenx}{-0.3}} \\
CoTTA & 40.7 \\
I-CoTTA & 40.2\textsuperscript{\textcolor{greenx}{-0.5}} \\
RoTTA & 27.7 \\
I-RoTTA & 27.5\textsuperscript{\textcolor{greenx}{-0.2}} \\
PETAL & 26.5 \\
I-PETAL & 26.2\textsuperscript{\textcolor{greenx}{-0.3}} \\
\bottomrule
\end{tabular}
}
\label{tab:performance}
% \end{table}

\vspace{-2em}
\end{wraptable}

\subsection{Wall-time results}
\label{sec:walltime}
We evaluate computational efficiency by measuring the wall-clock time required to process 10,000 CIFAR10C images on a single RTX 4080 GPU, with results shown in Tab.~\ref{tab:performance}. Our technique maintains a processing speed comparable to baseline methods, adding no computational overhead. MEMO's result highlights a significant issue with single-image adaptation methods, which exhibit substantially longer processing times.

\subsection{On single image adaptation methods.}
\label{sec:single_image_methods}

Single image adaptation approaches, like MEMO~\citep{ZhangLF22Memo}, are not susceptible to the studied problem of error accumulation. However, those methods have their own issues, e.g., they tend to be significantly more inefficient~\citep{evaluation_time_constraints}, as shown in Tab.~\ref{tab:performance}. MEMO is significantly slower than other techniques, with a wall-clock time around $20\times$ higher than AdaContrast. Moreover, it is outperformed by \mbox{IT-modified} methods and the baselines on most datasets, except ImageNet-R (see Tab.~\ref{tab:long} and Tab.~\ref{tab:short}).

\subsection{Results on common-length benchmarks}
\label{sec:short}
The overall accuracy of the tested TTA methods on common-length benchmarks is presented in Tab.~\ref{tab:short}. The results for IT are mixed, sometimes outperforming the baseline methods and sometimes falling behind. The failures can be attributed to its slow adaptability, which stems from the intransigence issue discussed in Sec.~\ref{sec:long}. Moreover, it is worth noting that the original methods were specifically tuned for these benchmarks, whereas no such tuning was performed for IT. Our goal is to highlight the lack of robustness to longer adaptation sequence lengths exhibited by state-of-the-art teacher-student framework-based TTA methods and to provide a simple and robust solution in the form of the intransigent teacher.

\begin{table}[htb!]
\caption{Classification accuracy [\%] for common-length benchmarks. The value in superscript indicates the improvements over the baseline.
}
\centering
\resizebox{0.5\linewidth}{!}{
\begin{tabular}{lcccc}
\toprule
\textbf{Method} & \textbf{C10-C} & \textbf{IN-C} & \textbf{IN-R} & \textbf{DN-126} \\ 

\midrule

Source & 56.5 & 18.0 & 36.2 & 54.7 \\
MEMO & 65.6 & 25.0 & 40.9 & 53.2 \\

\midrule
\multicolumn{5}{c}{\textsc{batch size 10}} \\
\midrule

TestBN & 75.0 & 27.0 & 36.6 & 46.5 \\
TENT & 75.7 & 31.2 & 38.9 & 52.4 \\
EATA & 77.4 & 36.0 & 43.1 & 54.4 \\
SAR & 75.8 & 31.3 & 41.9 & 52.8 \\
RDUMB & 77.2 & 34.8 & 41.3 & 52.0 \\

AdaContrast & 81.3 & 33.3 & 39.5 & 56.5 \\
I-AdaContrast & 82.0\textsuperscript{\textcolor{greenx}{+0.7}} & 33.8\textsuperscript{\textcolor{greenx}{+0.5}} & 39.8\textsuperscript{\textcolor{greenx}{+0.3}} & 59.6\textsuperscript{\textcolor{greenx}{+3.1}} \\

CoTTA & 75.1 & 26.4 & 41.1 & 52.0 \\
I-CoTTA & 69.8\textsuperscript{\textcolor{redx}{-5.3}} & 28.3\textsuperscript{\textcolor{greenx}{+1.9}} & 35.6\textsuperscript{\textcolor{redx}{-5.5}} & 49.5\textsuperscript{\textcolor{redx}{-2.5}} \\

RoTTA & 79.0 & 29.2 & 38.6 & 55.9 \\
I-RoTTA & 73.2\textsuperscript{\textcolor{redx}{-5.8}} & 29.4\textsuperscript{\textcolor{greenx}{+0.2}} & 39.3\textsuperscript{\textcolor{greenx}{+0.7}} & 56.6\textsuperscript{\textcolor{greenx}{+0.7}} \\

PETAL & 68.6 & 20.6 & 37.6 & 51.9 \\
I-PETAL & 74.3\textsuperscript{\textcolor{greenx}{+5.7}} & 29.8\textsuperscript{\textcolor{greenx}{+9.2}} & 37.6\textsuperscript{\textcolor{greenx}{+0.0}} & 54.0\textsuperscript{\textcolor{greenx}{+2.1}} \\

\midrule
\multicolumn{5}{c}{\textsc{batch size 64}} \\
\midrule

TestBN & 79.2 & 31.4 & 39.7 & 54.5 \\
TENT & 77.8 & 37.3 & 42.6 & 58.0 \\
EATA & 79.8 & 42.0 & 45.8 & 59.7 \\
SAR & 79.3 & 37.8 & 42.8 & 57.2 \\
RDUMB & 81.4 & 40.0 & 46.2 & 58.9 \\

AdaContrast & 82.6 & 34.8 & 40.9 & 62.0 \\
I-AdaContrast & 82.4\textsuperscript{\textcolor{redx}{-0.2}} & 35.1\textsuperscript{\textcolor{greenx}{+0.3}} & 41.0\textsuperscript{\textcolor{greenx}{+0.1}} & 61.7\textsuperscript{\textcolor{redx}{-0.3}} \\

CoTTA & 82.2 & 36.8 & 42.8 & 58.9 \\
I-CoTTA & 68.6\textsuperscript{\textcolor{redx}{-13.6}} & 31.7\textsuperscript{\textcolor{redx}{-5.1}} & 35.9\textsuperscript{\textcolor{redx}{-6.9}} & 54.4\textsuperscript{\textcolor{redx}{-4.5}} \\

RoTTA & 80.9 & 32.4 & 39.2 & 56.8 \\
I-RoTTA & 76.7\textsuperscript{\textcolor{redx}{-4.2}} & 30.6\textsuperscript{\textcolor{redx}{-1.8}} & 39.3\textsuperscript{\textcolor{greenx}{+0.1}} & 56.3\textsuperscript{\textcolor{redx}{-0.5}} \\

PETAL & 76.6 & 32.9 & 40.0 & 56.4 \\
I-PETAL & 78.5\textsuperscript{\textcolor{greenx}{+1.9}} & 34.2\textsuperscript{\textcolor{greenx}{+1.3}} & 39.6\textsuperscript{\textcolor{redx}{-0.4}} & 54.4\textsuperscript{\textcolor{redx}{-2.0}} \\

\bottomrule
\end{tabular}
}
\label{tab:short}
\end{table}

\subsection{Potential of adaptive \texorpdfstring{$\beta$}{beta} value}
\label{sec:add_exp/adaptive_beta}
In Tab.~\ref{tab:fr_after_1_loop}, we explore a dynamic approach to adjusting the teacher model's momentum parameter ($\beta$). Our experiment begins with the default value of $\beta\!=\!0.999$, allowing initial teacher model plasticity, then transitions to complete weight preservation of IT ($\beta\!=\!1.0$) after one full cycle through the data. This hybrid approach outperforms our IT technique in several cases, demonstrating the potential of adaptive momentum strategies.

However, the results are not uniformly positive with our standard IT outperforming the hybrid method in some cases (AdaContrast on ImageNet-C (L) and CoTTA on \mbox{DomainNet-126 (L)}). This suggests that the fixed period length is not a universal value and there is a need to adjust it correctly.

\begin{table}[t!]
\caption{
Classification accuracy [\%] for long scenarios with the weights of the teacher fixed only after the 1st loop on the test sequence. The
value in superscript indicates the improvements over the IT technique's performance. The batch size is equal to 64.
}
\centering
\resizebox{0.65\linewidth}{!}{
\begin{tabular}{lccccc}
%\hline
\toprule
\textbf{Method} & 
\textbf{C10-C (L)} &
\textbf{IN-C (L)} &
\textbf{IN-R (L)} &
\textbf{DN-126 (L)} &
\textbf{Avg.}
\\ 

\midrule

AdaCont. &
85.2\textsuperscript{\textcolor{redx}{-0.1}} & 
38.4\textsuperscript{\textcolor{redx}{-2.0}} & 38.2\textsuperscript{\textcolor{greenx}{+0.1}} & 65.3\textsuperscript{\textcolor{greenx}{+0.9}} & 56.8\textsuperscript{\textcolor{redx}{-0.3}} \\

CoTTA &
72.0\textsuperscript{\textcolor{greenx}{+3.7}} & 45.0\textsuperscript{\textcolor{greenx}{+9.6}} & 42.8\textsuperscript{\textcolor{greenx}{+3.3}} & 49.1\textsuperscript{\textcolor{redx}{-6.9}} & 52.2\textsuperscript{\textcolor{greenx}{+2.4}} \\

RoTTA &
80.4\textsuperscript{\textcolor{greenx}{+0.7}} & 36.1\textsuperscript{\textcolor{greenx}{+3.2}} & 41.0\textsuperscript{\textcolor{greenx}{+1.3}} & 57.9\textsuperscript{\textcolor{greenx}{+1.3}} & 53.9\textsuperscript{\textcolor{greenx}{+1.7}} \\

\bottomrule
\end{tabular}
}
\label{tab:fr_after_1_loop}
%}
\end{table}
\begin{table*}[htb!]
\caption{Classification accuracy [\%] for long (L) scenarios while utilizing only the student model for adaptation (\textbf{S. only}) and tuning the learning rate parameter \textbf{(LR)}. The batch size is equal to 64. \textbf{Avg. Oracle} is the average accuracy when the LR value is chosen separately for each of the datasets with Oracle. \textbf{Avg. Transfer IN-C} is the average accuracy with a single LR value chosen on the ImageNet-C dataset using the Oracle method. Average accuracy with the learning rate chosen on standard unrepeated benchmarks using the Oracle is presented in \textbf{Avg. Transfer $1\times$Loop} column. The value in superscript indicates the decrease below the IT technique's performance.
}
\centering
\resizebox{0.9\linewidth}{!}{
\begin{tabular}{lc@{\hskip 0.5in}cccc@{\hskip 0.5in}ccc}
\toprule
\multirow{2}{*}{\textbf{Method}} & 
\multirow{2}{*}{\textbf{LR}} & 
\multirow{2}{*}{\textbf{C10-C (L)}} &
\multirow{2}{*}{\textbf{IN-C (L)}} &
\multirow{2}{*}{\textbf{IN-R (L)}} &
\multirow{2}{*}{\textbf{DN-126 (L)}} &
\multicolumn{3}{c}{\textbf{Avg.}} \\

&&&&&&
\textbf{Oracle} &
\textbf{Transfer IN-C} &
\textbf{Transfer $1\times$Loop}
\\ 
\midrule

\multirow{3}{*}{\makecell{AdaCont.\\S. only}}
& 1e-5 & \textbf{82.4} & 23.0 & 37.4 & 59.1 & 
\multirow{3}{*}{54.7\textsuperscript{\textcolor{redx}{-3.8}}} & 
\multirow{3}{*}{54.5\textsuperscript{\textcolor{redx}{-2.6}}} & 
\multirow{3}{*}{50.5\textsuperscript{\textcolor{redx}{-6.6}}} \\
& 1e-6 & 81.4 & \textbf{36.6} & \textbf{40.7} & \textbf{59.2} & & \\
& 1e-7 & 79.1 & 31.9 & 39.6 & 57.0 & & \\
\midrule

\multirow{3}{*}{\makecell{CoTTA\\S. only}} 
& 1e-4 & 35.3 & 6.3 & 14.1 & 1.1 & 
\multirow{3}{*}{40.6\textsuperscript{\textcolor{redx}{-12.5}}} & 
\multirow{3}{*}{37.1\textsuperscript{\textcolor{red}{-8.9}}} & 
\multirow{3}{*}{26.7\textsuperscript{\textcolor{redx}{-23.1}}} \\
& 1e-5 & 70.8 & \textbf{34.8} & \textbf{41.3} & 1.6 & & \\
& 1e-6 & \textbf{79.2} & 31.9 & 40.0 & \textbf{7.0} & & \\
\midrule

\multirow{3}{*}{\makecell{RoTTA\\S. only}} 
& 1e-4 & 21.9 & 1.4 & 5.8 & 6.9 & 
\multirow{3}{*}{45.9\textsuperscript{\textcolor{redx}{-6.5}}} & 
\multirow{3}{*}{45.9\textsuperscript{\textcolor{redx}{-6.5}}} & 
\multirow{3}{*}{41.4\textsuperscript{\textcolor{redx}{-10.8}}} \\
& 1e-5 & 51.6 & 14.5 & 35.6 & 48.9 & & \\
& 1e-6 & \textbf{86.6} & \textbf{26.0} & \textbf{37.3} & \textbf{53.8} & & \\
\bottomrule
\end{tabular}
}
\label{tab:only_student}
\end{table*}

\subsection{Using only student model}
\label{sec:add_exp/only_student}
Given that the intransigent teacher strategy maintains fixed teacher model weights, one might question whether the teacher model is truly necessary. To address this, we conduct experiments using only the student network (see Tab.~\ref{tab:only_student}). In these experiments, we take the student model's outputs directly as pseudo-labels, without any teacher involvement.

We included various averaged results based on the hyperparameter selection approach, acknowledging the non-trivial nature of the selection in TTA. 
Following a realistic approach, we employ an Oracle technique on \mbox{ImageNet-C (L)} as a reference benchmark (inspired by~\citet{rusakSPEGBBB22}, we name it \mbox{Transfer IN-C}) and apply the selected learning rate across all datasets. Additionally, we choose the learning rate using Oracle on the first loop of the long (L) scenarios (Transfer $1\times$Loop). We also present results when the learning rate is chosen separately for each of the datasets with the Oracle.

While it appears that using only the student model can achieve reasonable performance with carefully tuned hyperparameters, it significantly underperforms compared to the intransigent teacher approach. The intransigent teacher delivers consistent results, demonstrating robustness across various hyperparameter selection methods, whereas the student-only model exhibits higher performance variance using different selection approaches.

\subsection{Effects of intransigence amount extended experiment}
\label{sec:add_exp/it_amount}
\begin{figure}[h!]
    \centering
    \includegraphics[width=0.4\linewidth]{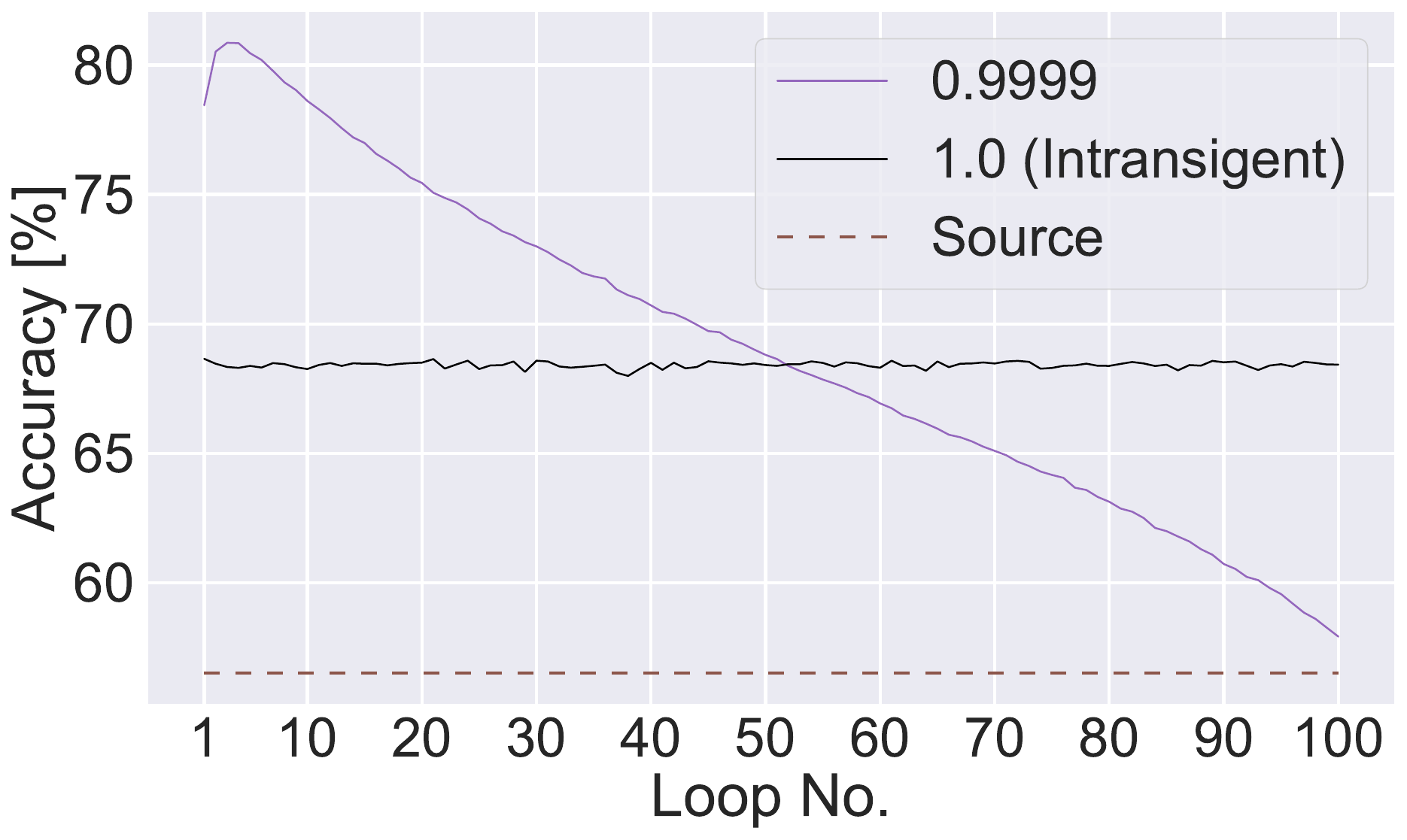}
    \caption{Mean accuracy [\%] of CoTTA with varying $\beta$ for each loop of common CIFAR10-C testing sequence repeated 100 times. The brown dashed line indicates the accuracy of the source model as a reference.}
    \label{fig:100loops}
\end{figure}

To signify the point of Sec.~\ref{sec:long}, Fig.~\ref{fig:100loops} shows results where the test sequence is extended to 100 loops of common CIFAR10-C. It verifies that CoTTA with ET and $\beta\!=\!0.9999$ degrades below the performance of IT, given enough samples in the test sequence. This observation highlights a significant issue of TTA methods, as they can face test sequences of arbitrary lengths after deployment.

\begin{table*}[htb!]
\caption{
Classification accuracy [\%] for long scenarios with restoration probability parameter $p$ of CoTTA method tuned. The batch size is equal to 64. \textbf{Avg. Def.} is the average accuracy with default $p$ value. \textbf{Avg. Transfer IN-C} is the average accuracy with a single $p$ value chosen on the ImageNet-C dataset using the Oracle method. Average accuracy when the $p$ value is chosen separately for each of the datasets with Oracle is presented in \textbf{Avg. Oracle} column.}
\centering
\resizebox{0.9\linewidth}{!}{
\begin{tabular}{l@{\hskip 0.5in}cccc@{\hskip 0.5in}ccc}
\toprule
\multirow{2}{*}{$p$\textbf{-value}} & 
\multirow{2}{*}{\textbf{C10-C (L)}} &
\multirow{2}{*}{\textbf{IN-C (L)}} &
\multirow{2}{*}{\textbf{IN-R (L)}} &
\multirow{2}{*}{\textbf{DN-126 (L)}} &

\multicolumn{3}{c}{\textbf{Avg.}} \\

&&&&& \textbf{Def.} & \textbf{Transfer IN-C} & \textbf{Oracle} \\ 

\midrule
0.1          & \textbf{73.1} & 29.0 & 41.8 & 26.9 & \multirow{6}{*}{\textbf{51.2}} & \multirow{6}{*}{\textbf{51.6}} & \multirow{6}{*}{\textbf{58.1}} \\
0.01         & 53.7 & 24.8 & 35.6 & 13.7 & & & \\
0.001 (Def.) & 56.0 & 52.8 & \textbf{50.5} & 45.6 & & & \\
0.0001       & 54.7 & \textbf{53.7} & 45.0 & 52.9 & & & \\
0.00001      & 54.3 & 53.6 & 49.0 & \textbf{55.0} & & & \\
0.0          & 52.7 & 53.5 & 48.9 & 54.5 & & & \\
\bottomrule
\end{tabular}
}
\label{tab:tune_reset_param}
\end{table*}

\subsection{Discussion on model reset mechanism}
\label{sec:hyperparam/reset}
CoTTA's proposed resetting mechanism aims to preserve source knowledge by stochastically restoring portions of the student model's weights to their original source state during each update iteration. In principle, an effective source knowledge preservation technique should eliminate the need for our IT technique.

However, CoTTA's reset mechanism introduces a restoration probability parameter. To ensure our findings were not biased by suboptimal parameter selection, we conducted parameter tuning experiments, documented in Tab.~\ref{tab:tune_reset_param}. These results reveal that the optimal restoration probability varies across datasets, with model performance dependent on this parameter. When following a realistic scenario of tuning on a single dataset, the performance improvements were marginal (Avg. Transfer IN-C). Only by using an Oracle approach on all benchmarks, we observe performance gains, highlighting the practical limitations of this approach.

\begin{figure*}[ht!]
    \centering
    \includegraphics[width=0.95\linewidth]{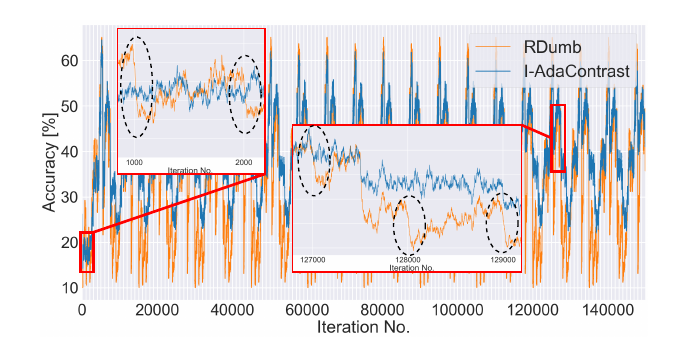}
    \caption{Batch-wise accuracy plots of RDumb and I-AdaContrast methods on ImageNet-C (L) benchmark. The accuracy values were smoothed to make the plot clearer. RDumb resets the model every 1000 iterations, which causes significant drops in accuracy after the reset.}
    \label{fig:rdumb_vs_it}
\end{figure*}

\subsection{Discussion on RDumb}
\label{sec:rdumb}
Similar to us, RDumb focuses on long scenarios and provides a straightforward analysis. However, there are several key differences between our work and RDumb.

First, RDumb experiments exclusively with common corruptions, while we demonstrate the phenomenon across various types of distribution shifts, including ImageNet-R and DomainNet. Additionally, we analyze this issue using several newer methods that incorporate the mean-teacher mechanism, which theoretically should address the problem (whereas only CoTTA among RDumb’s baselines used this mechanism). Furthermore, we highlight a potential, simple solution to mitigate the issue, which is not~explored~in~RDumb.

RDumb method mechanism of periodically resetting the model to its initial state leads to significant accuracy drops immediately following each reset, as illustrated in Fig.~\ref{fig:rdumb_vs_it}. Such instability is particularly concerning since reliable test-time adaptation should maintain consistent performance throughout the adaptation process. Furthermore, the same constant reset interval is likely not optimal for every case, which adds a hyperparameter to select. In contrast, our IT approach achieves comparable performance without requiring parameter tuning.

\subsection{Augmentation count ablation}
\label{sec:aug_abl}
\begin{figure*}[ht!]
    \centering
    \includegraphics[width=0.99\linewidth]{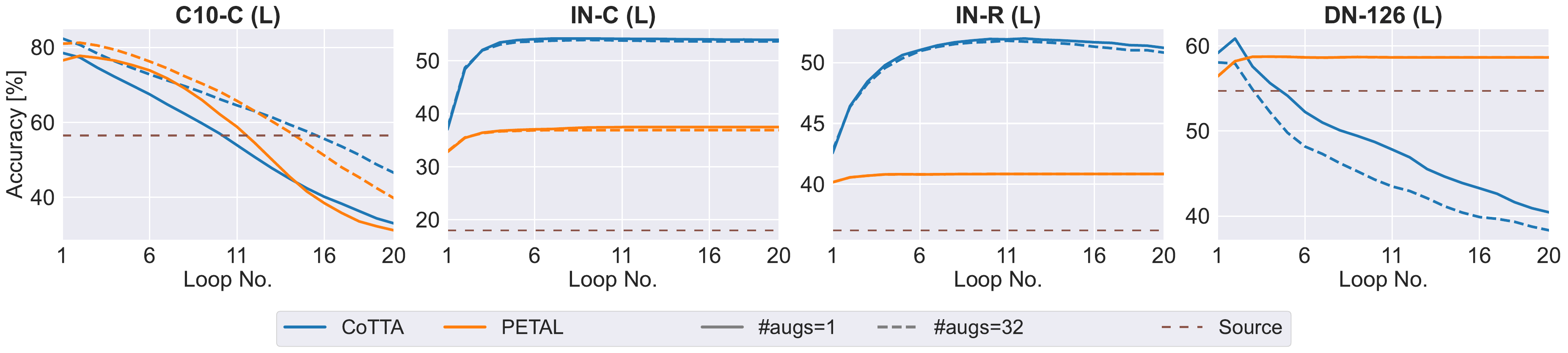}
    \caption{Mean accuracy [\%] for each loop of the common testing sequence on CIFAR10-C (L), ImageNet-C (L), ImageNet-R (L), and DomainNet-126 for CoTTA and PETAL. Methods are tested with pseudo-labels generated by averaging across predictions on either 1 or 32 randomly augmented images. The brown dashed line indicates the source model accuracy as a reference.}
    \label{fig:augs_abl}
\end{figure*}

The CoTTA~\citep{COTTA} and PETAL~\citep{petal} methods update the student network using a consistency loss between the student and teacher models.
When the prediction confidence of the source model falls below a predefined threshold, the teacher generates pseudo-labels by averaging predictions across 32 distinct random augmentations of each image. This ensembling strategy requires 31 additional forward passes per batch, introducing a severe computational bottleneck that significantly reduces throughput compared to alternative state-of-the-art approaches.
To evaluate the necessity of this computational overhead, Fig.~\ref{fig:augs_abl} compares pseudo-label generation using either 1 or 32 random augmentations.

Across evaluations on ImageNet-C (L) and ImageNet-R (L), accuracy profiles and degradation trajectories remain remarkably consistent regardless of the augmentation count. On ImageNet-R (L), CoTTA with a single augmentation even exhibits a marginally slower degradation rate toward the end of the sequence, while on DomainNet-126 (L), it yields a $2\%$ to $4\%$ accuracy improvement over the 32-augmentation baseline. The largest divergence occurs on CIFAR10-C (L), where 32 augmentations yield higher absolute accuracy for both methods. 
However, the underlying degradation trend persists, and both configurations ultimately collapse. Critically, reducing the augmentation count does not accelerate this collapse, demonstrating that the overall evaluation outcomes remain robust.
While a definitive theoretical explanation for this behavior is outside the scope of this work, we hypothesize that averaging predictions across random transformations does not always yield a significant improvement in pseudo-label accuracy.

\subsection{Distillation vs. batch normalization recalibration}
\label{sec:decomp_stats_distill}
To isolate whether the IT's performance gains stem from the distillation process or from simply recalibrating batch normalization (BN) statistics, we evaluate variants of our IT-enhanced methods without backpropagation (w/o BP). We compare these variants against the original baselines and the full IT baselines (Tab.~\ref{tab:norm_dis}).
In the w/o BP configuration, the adaptation process degenerates to adjusting only the student BN statistics, removing all influence from the teacher.

The specific technique for adjusting BN statistics varies across methods: CoTTA and PETAL utilize per-batch BN calculation, matching the TestBN baseline. Hence, their w/o BP results are identical to TestBN. AdaContrast continuously updates the EMA of the BN statistics, while RoTTA leverages its proposed robust batch normalization (RBN) module.

The results show that updating BN statistics alone provides a substantial accuracy boost. However, this update is not the sole driver of optimization. Across all 32 combinations of benchmarks, batch sizes, and baseline methods, the full IT framework outperforms or matches the w/o BP variant in 26 cases. This demonstrates that distillation-based adaptation yields consistent, additive improvements across the majority of evaluations. In the few cases where full IT underperforms relative to w/o BP, the original baseline method itself underperforms compared to both "I-" variants (e.g.,~PETAL on CIFAR10-C~(L) with a batch size of 64, and AdaContrast on ImageNet-R~(L) with a batch size of 10). This pattern indicates that when the underlying adaptation method fails over long horizons, stripping away all plasticity except for the BN statistics update yields the most effective results.
Consequently, the average accuracy of the full IT method exceeds that of the w/o BP variants in all setups except for CoTTA. We attribute this exception to the poor baseline performance of CoTTA on CIFAR10-C~(L), which underscores that the final performance of IT remains fundamentally constrained by its underlying baseline.

In summary, updating batch normalization statistics alone (w/o BP) delivers significant performance gains. However, this approach is strictly limited to architectures containing batch normalization layers. By contrast, distilling knowledge from an intransigent teacher provides further, cross-architecture improvements in most scenarios, validating the broader competitiveness of our approach.

\begin{table}[t!]
\caption{Classification accuracy [\%] for long (L) scenarios. Superscript indicates improvements over the baseline. \textbf{Bold} indicates best performing method. \textcolor{gray}{Gray} color indicates model collapse -- performance worse than the non-adapting model (Source). Results averaged from 3 random seeds.
}
\centering
\resizebox{0.8\linewidth}{!}{
\begin{tabular}{lccccc}
\toprule
\textbf{Method} & \textbf{C10-C (L)} & \textbf{IN-C (L)} & \textbf{IN-R (L)} & \textbf{DN-126 (L)} 
% & \textbf{CCC} 
& \textbf{Avg.}
\\ 
\midrule
Source & 56.5 & 18.0 & 36.2 & 54.7 
% & 16.8 
& 41.4 \\

\midrule
\midrule
\multicolumn{6}{c}{\textsc{batch size 10}} \\
\midrule

TestBN & 75.1 & 26.9 & 36.2 & \textcolor{gray}{49.6} 
% & 22.5 
& 47.0 \\

\midrule

AdaCont. & 72.1 & \textcolor{gray}{2.3} & \textcolor{gray}{8.0} & \textcolor{gray}{47.0} 
% & \textcolor{gray}{0.4} 
& 32.4 \\

I-AdaCont. w/o BP. 
& 77.7\textsuperscript{\textcolor{greenx}{+5.6}}
& 29.6\textsuperscript{\textcolor{greenx}{+6.3}}
& 39.0\textsuperscript{\textcolor{greenx}{+31.0}}
& \textcolor{gray}{54.4}\textsuperscript{\textcolor{greenx}{+7.4}}
% & 
& 50.2\textsuperscript{\textcolor{greenx}{+17.8}} \\

I-AdaCont.
& \textbf{84.1}\textsuperscript{\textcolor{greenx}{+12.0}}
& \textbf{39.5}\textsuperscript{\textcolor{greenx}{+37.2}}
& \textcolor{gray}{35.3}\textsuperscript{\textcolor{greenx}{+27.3}}
& \textbf{63.2}\textsuperscript{\textcolor{greenx}{+16.2}}
% & 21.8\textsuperscript{\textcolor{greenx}{+21.4}}
& \textbf{55.5}\textsuperscript{\textcolor{greenx}{+23.1}} \\

\midrule

CoTTA & \textcolor{gray}{23.8} & \textcolor{gray}{3.8} & \textcolor{gray}{33.7} & \textcolor{gray}{6.0} 
% & 17.3 
& 16.8 \\

I-CoTTA. w/o BP. 
& 75.1\textsuperscript{\textcolor{greenx}{+51.3}}
& 26.9\textsuperscript{\textcolor{greenx}{+23.1}}
& 36.2\textsuperscript{\textcolor{greenx}{+2.5}}
& \textcolor{gray}{49.6}\textsuperscript{\textcolor{greenx}{+43.6}}
% & 22.5 \textsuperscript{\textcolor{greenx}{}}
& 47.0\textsuperscript{\textcolor{greenx}{+30.2}} \\

I-CoTTA 
& 69.7\textsuperscript{\textcolor{greenx}{+45.9}}
& 27.6\textsuperscript{\textcolor{greenx}{+23.8}}
& 37.4\textsuperscript{\textcolor{greenx}{+3.7}}
& \textcolor{gray}{50.3}\textsuperscript{\textcolor{greenx}{+44.3}}
% & 25.7\textsuperscript{\textcolor{greenx}{+8.4}}
& 46.3\textsuperscript{\textcolor{greenx}{+29.5}} \\

\midrule

RoTTA & 82.5 & 24.4 & 43.0 & \textcolor{gray}{45.6} 
% & \textcolor{gray}{1.1} 
& 48.9 \\

I-RoTTA. w/o BP. 
& 70.7\textsuperscript{\textcolor{redx}{-11.8}}
& 24.2\textsuperscript{\textcolor{redx}{-0.2}}
& 36.8\textsuperscript{\textcolor{redx}{-6.2}}
& \textcolor{gray}{53.4}\textsuperscript{\textcolor{greenx}{7.8}}
% & 
& 46.3\textsuperscript{\textcolor{redx}{-2.6}} \\

I-RoTTA 
& 79.0\textsuperscript{\textcolor{redx}{-3.5}}
& 33.6\textsuperscript{\textcolor{greenx}{+9.2}} 
& 39.9\textsuperscript{\textcolor{redx}{-3.1}}
& 57.8\textsuperscript{\textcolor{greenx}{+12.2}}
% & 23.4\textsuperscript{\textcolor{greenx}{+22.3}}
& 52.6\textsuperscript{\textcolor{greenx}{+3.7}} \\

\midrule

PETAL & 68.0 & \textcolor{gray}{3.9} & 37.9 & \textcolor{gray}{47.3} 
% & \textcolor{gray}{0.5} 
& 39.3 \\

I-PETAL. w/o BP. 
& 75.1\textsuperscript{\textcolor{greenx}{+7.1}}
& 26.9\textsuperscript{\textcolor{greenx}{+23.0}}
& 36.2\textsuperscript{\textcolor{redx}{-1.7}}
& \textcolor{gray}{49.6}\textsuperscript{\textcolor{greenx}{+2.3}}
% & 22.5 \textsuperscript{\textcolor{greenx}{}}
& 47.0\textsuperscript{\textcolor{greenx}{+7.7}} \\

I-PETAL 
& 74.2\textsuperscript{\textcolor{greenx}{+6.2}}
& 28.7\textsuperscript{\textcolor{greenx}{+24.8}} 
& 36.5\textsuperscript{\textcolor{redx}{-1.4}}
& \textcolor{gray}{50.7}\textsuperscript{\textcolor{greenx}{+3.4}}
% & \textcolor{gray}{13.9}\textsuperscript{\textcolor{greenx}{+13.4}}
& 47.5\textsuperscript{\textcolor{greenx}{+8.2}} \\

\midrule
\midrule
\multicolumn{6}{c}{\textsc{batch size 64}} \\
\midrule

TestBN & 79.1 & 31.4 & 39.6 & \textcolor{gray}{54.4} 
% & 27.3 
& 51.1 \\

\midrule

AdaCont. & 81.8 & 18.8 & \textcolor{gray}{26.5} & 61.7 
% & \textcolor{gray}{2.4} 
& 47.2 \\

I-AdaCont. w/o BP. 
& 77.9\textsuperscript{\textcolor{redx}{-3.9}}
& 30.4\textsuperscript{\textcolor{greenx}{+11.6}}
& 39.4\textsuperscript{\textcolor{greenx}{+12.9}}
& 55.0\textsuperscript{\textcolor{redx}{-6.7}}
% & 
& 50.7\textsuperscript{\textcolor{greenx}{+3.5}} \\

I-AdaCont. 
& \textbf{85.4}\textsuperscript{\textcolor{greenx}{+3.6}}
& 40.4\textsuperscript{\textcolor{greenx}{+21.6}}
& 38.1\textsuperscript{\textcolor{greenx}{+11.6}} 
& \textbf{64.4}\textsuperscript{\textcolor{greenx}{+2.7}}
% & 23.6\textsuperscript{\textcolor{greenx}{+21.2}} 
& 57.1\textsuperscript{\textcolor{greenx}{+9.9}} \\

\midrule

CoTTA & \textcolor{gray}{56.0} & \textbf{52.8} & \textbf{50.5} & \textcolor{gray}{45.6} 
% & \textcolor{gray}{8.3} 
& 51.2 \\ %1 aug

I-CoTTA. w/o BP. 
& 79.1\textsuperscript{\textcolor{greenx}{+23.1}}
& 31.4\textsuperscript{\textcolor{redx}{-21.4}}
& 39.6\textsuperscript{\textcolor{redx}{-10.9}}
& \textcolor{gray}{54.4}\textsuperscript{\textcolor{greenx}{+8.8}}
% & \textsuperscript{\textcolor{greenx}{}}
& 51.1\textsuperscript{\textcolor{redx}{-0.1}} \\

I-CoTTA 
& 68.3\textsuperscript{\textcolor{greenx}{+12.3}}
& 35.4\textsuperscript{\textcolor{redx}{-17.4}}
& 39.6\textsuperscript{\textcolor{redx}{-10.9}}
& 56.0\textsuperscript{\textcolor{greenx}{+10.4}}
% & 26.3\textsuperscript{\textcolor{greenx}{+18.0}}
& 49.8\textsuperscript{\textcolor{redx}{-1.4}} \\

\midrule

RoTTA & 82.3 & \textcolor{gray}{13.2} & 43.4 & \textcolor{gray}{50.3} 
% & \textcolor{gray}{1.1} 
& 47.3 \\

I-RoTTA. w/o BP. 
& 70.7\textsuperscript{\textcolor{redx}{-11.6}}
& 24.2\textsuperscript{\textcolor{greenx}{+11.0}}
& 36.8\textsuperscript{\textcolor{redx}{-6.6}}
& \textcolor{gray}{53.4}\textsuperscript{\textcolor{greenx}{+3.1}}
% & 
& 46.3\textsuperscript{\textcolor{redx}{-1.0}} \\

I-RoTTA 
& 79.7\textsuperscript{\textcolor{redx}{-2.6}}
& 32.9\textsuperscript{\textcolor{greenx}{+19.7}}
& 39.7\textsuperscript{\textcolor{redx}{-3.7}}
& 56.6\textsuperscript{\textcolor{greenx}{+6.3}}
% & 21.7\textsuperscript{\textcolor{greenx}{+20.6}}
& 52.2\textsuperscript{\textcolor{greenx}{+4.9}} \\

\midrule

PETAL & \textcolor{gray}{56.3} & 36.8 & 40.7 & 58.6 
% & \textcolor{gray}{13.1} 
& 48.1\\

I-PETAL. w/o BP. 
& 79.1\textsuperscript{\textcolor{greenx}{+22.8}}
& 31.4\textsuperscript{\textcolor{redx}{-5.4}}
& 39.6\textsuperscript{\textcolor{redx}{-1.1}}
& \textcolor{gray}{54.4}\textsuperscript{\textcolor{redx}{-4.2}}
% & 
& 51.1\textsuperscript{\textcolor{greenx}{+3.0}}\\

I-PETAL
& 78.8\textsuperscript{\textcolor{greenx}{+22.5}}
& 32.3\textsuperscript{\textcolor{redx}{-4.5}}
& 39.8\textsuperscript{\textcolor{redx}{-0.9}}
& 55.4\textsuperscript{\textcolor{redx}{-3.2}}
% & \textcolor{gray}{16.3}\textsuperscript{\textcolor{greenx}{+3.2}}
& 51.6\textsuperscript{\textcolor{greenx}{+3.5}} \\

\bottomrule
\end{tabular}
}
\label{tab:norm_dis}
\end{table}

\subsection{IT teacher with fixed source BN statistics}
\label{sec:add_exp/fixed_bn}
Using batch normalization (BN) statistics computed at test time is a common strategy in TTA. In our case, we adopt the same strategy for computing BN statistics as the baseline methods: CoTTA utilizes the TestBN technique, whereas AdaContrast and RoTTA update BN statistics using exponential moving average-based approaches.

To provide additional insight, we also experimented with fixed teacher statistics at test time, which are calculated during training on the source data (see Table~\ref{tab:fixed_stats}).
The results suggest that adjusting the BN statistics in the teacher model improves overall performance, aligning with the findings of the TTA community~\citep{rotta,bn_adapt,tta_survey}. This indicates that adapting these statistics can significantly mitigate performance drops on out-of-distribution data.

\begin{table}[h]
\centering
\caption{
Classification accuracy [\%] for long scenarios while utilizing fixed teacher statistics calculated on source data in batch normalization layers. The batch size is equal to 64. Superscript indicates improvements over the full IT approach.
}
\resizebox{0.7\linewidth}{!}{
\begin{tabular}{lccccc}
\toprule
\textbf{Method} & \textbf{C10-C (L)} & \textbf{IN-C (L)} & \textbf{IN-R (L)} & \textbf{DN-126 (L)} & \textbf{Avg.} \\
\midrule
I-AdaCont. Src. BN
& \textbf{83.2} & \textbf{24.8} & 35.3 & \textbf{63.3} & \textbf{51.7}\textsuperscript{\textcolor{redx}{-5.4}} \\

I-CoTTA Src. BN
& 49.4 & 17.6 & 35.5 & 52.3 & 38.7\textsuperscript{\textcolor{redx}{-11.1}} \\

I-RoTTA Src. BN
& 61.1 & 16.2 & \textbf{37.4} & 51.8 & 41.6\textsuperscript{\textcolor{redx}{-10.6}} \\
\bottomrule
\end{tabular}
}
\label{tab:fixed_stats}
\end{table}

\section{Remaining Results}
\label{sec:add_plots}
\begin{figure*}[p]
\vspace*{\fill}

% --- FIRST FIGURE SET ---
\centering
\begin{minipage}{0.45\textwidth}
  \centering
  \includegraphics[width=1.0\linewidth]{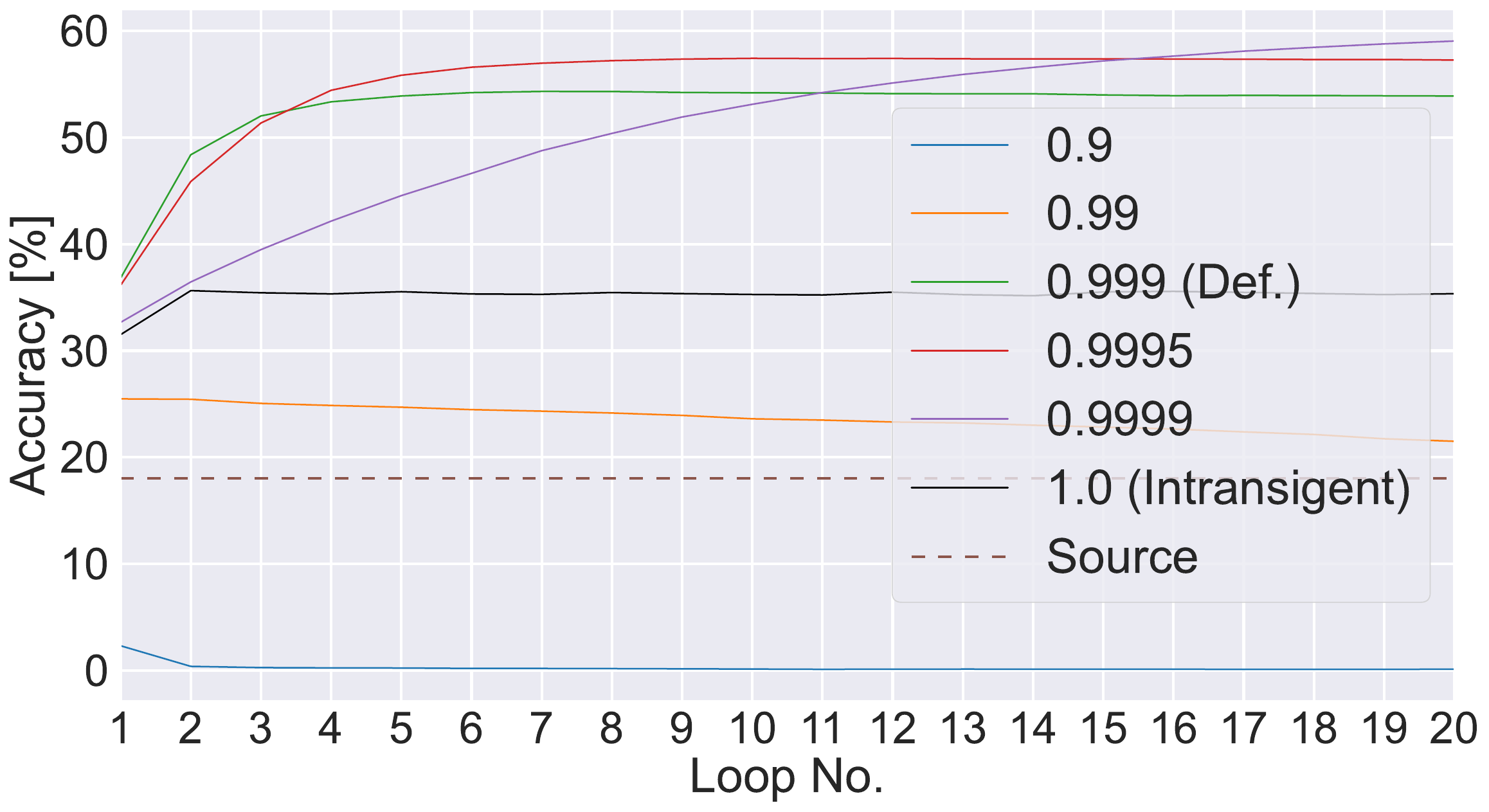}
  \label{fig:ema_cotta_inc}
\end{minipage}\hfill 
\begin{minipage}{0.45\textwidth}
  \centering
  \includegraphics[width=1.0\linewidth]{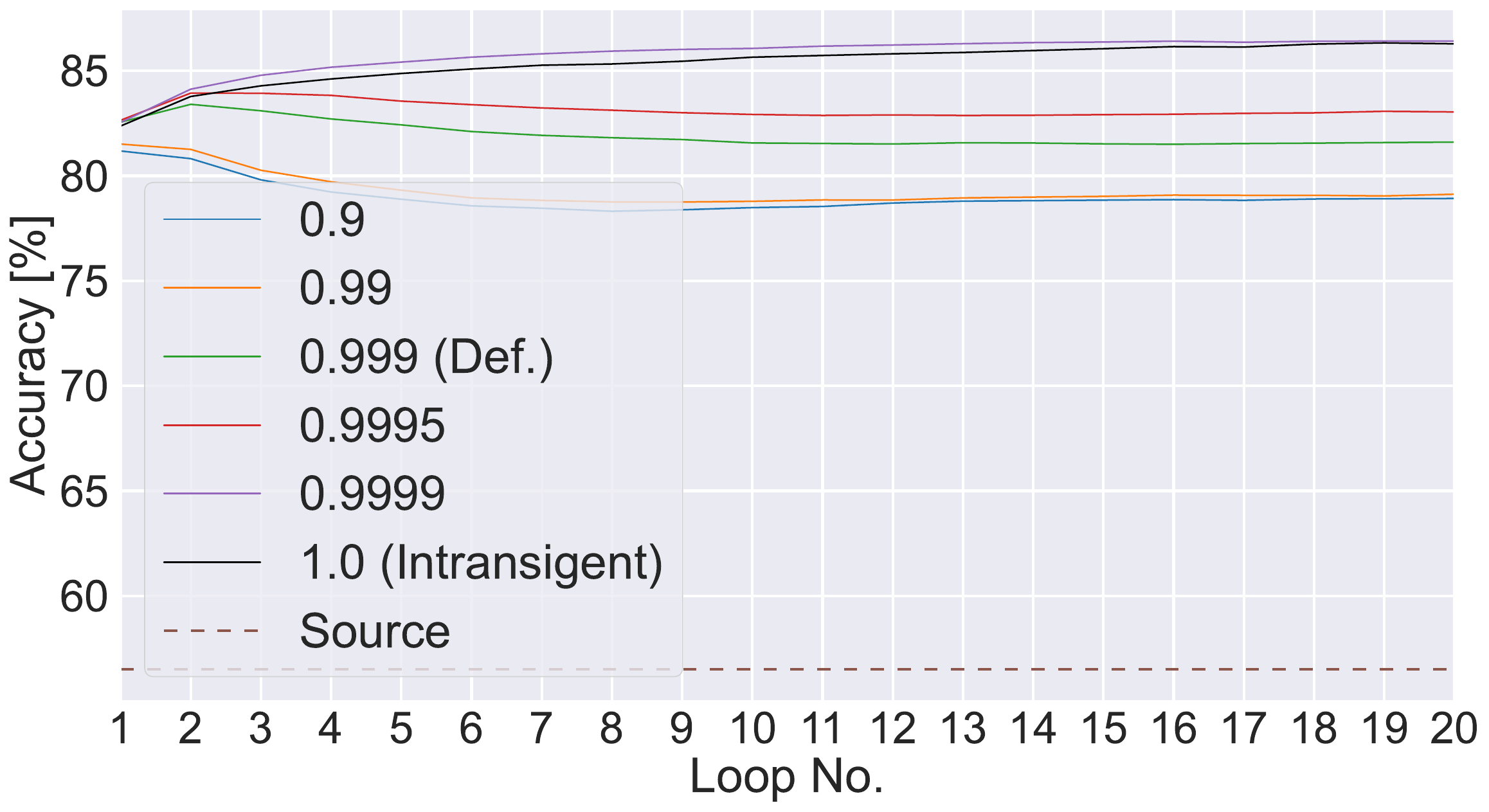}
  \label{fig:ema_ada_cifar}
\end{minipage}
\caption{Mean accuracy [\%] for each loop of common testing sequence on ImageNet-C (L) using CoTTA \textbf{(left)} and on CIFAR10-C (L) using AdaContrast \textbf{(right)}. The brown dashed line indicates the source model accuracy.}
\label{fig:ema_cotta_ada}

\vspace{3em}

% --- SECOND FIGURE SET ---
\centering
\begin{minipage}{0.45\textwidth}
  \centering
  \includegraphics[width=1.0\linewidth]{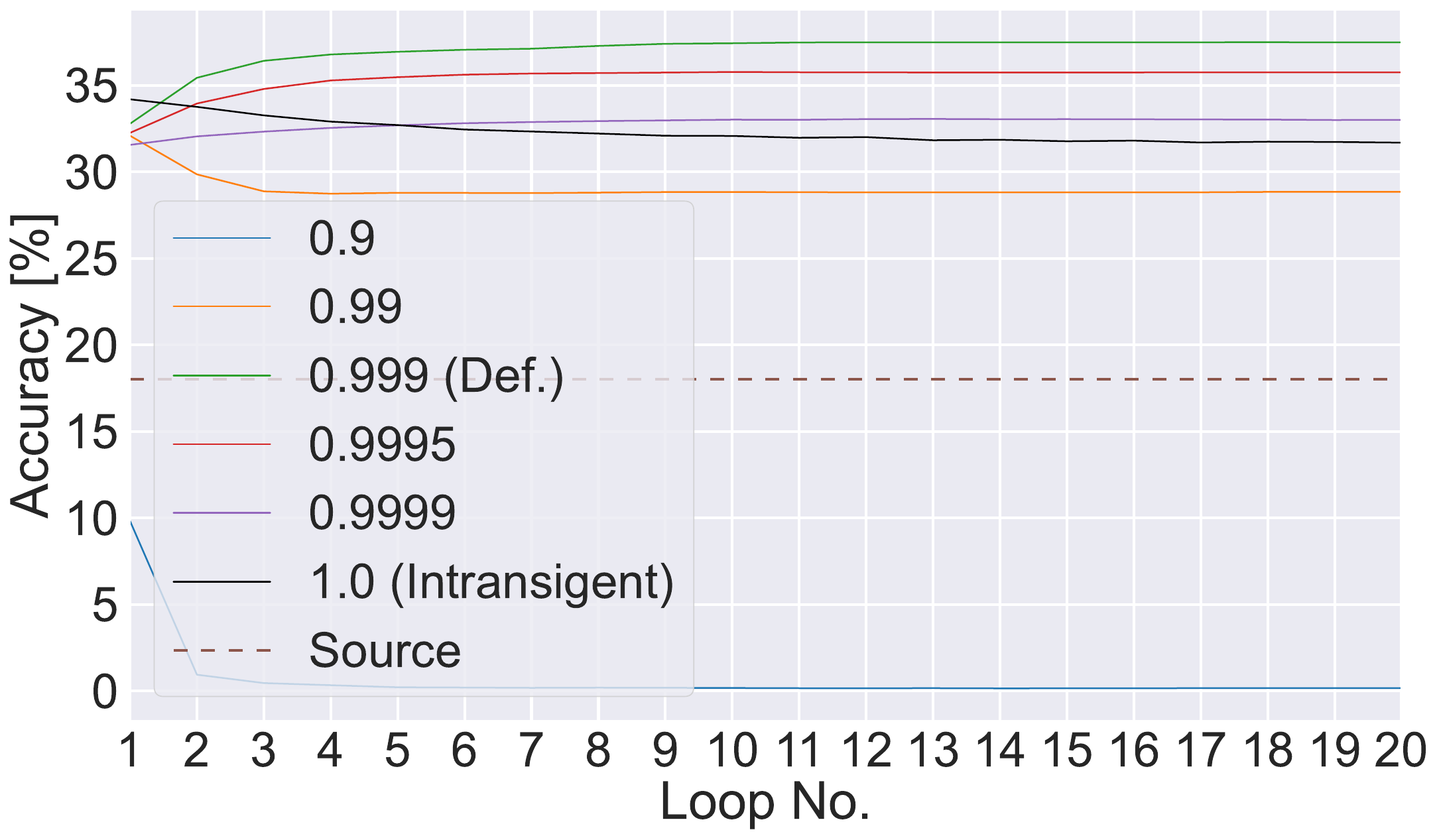}
  \label{fig:ema_petal_inc}
\end{minipage}\hfill
\begin{minipage}{0.45\textwidth}
  \centering
  \includegraphics[width=1.0\linewidth]{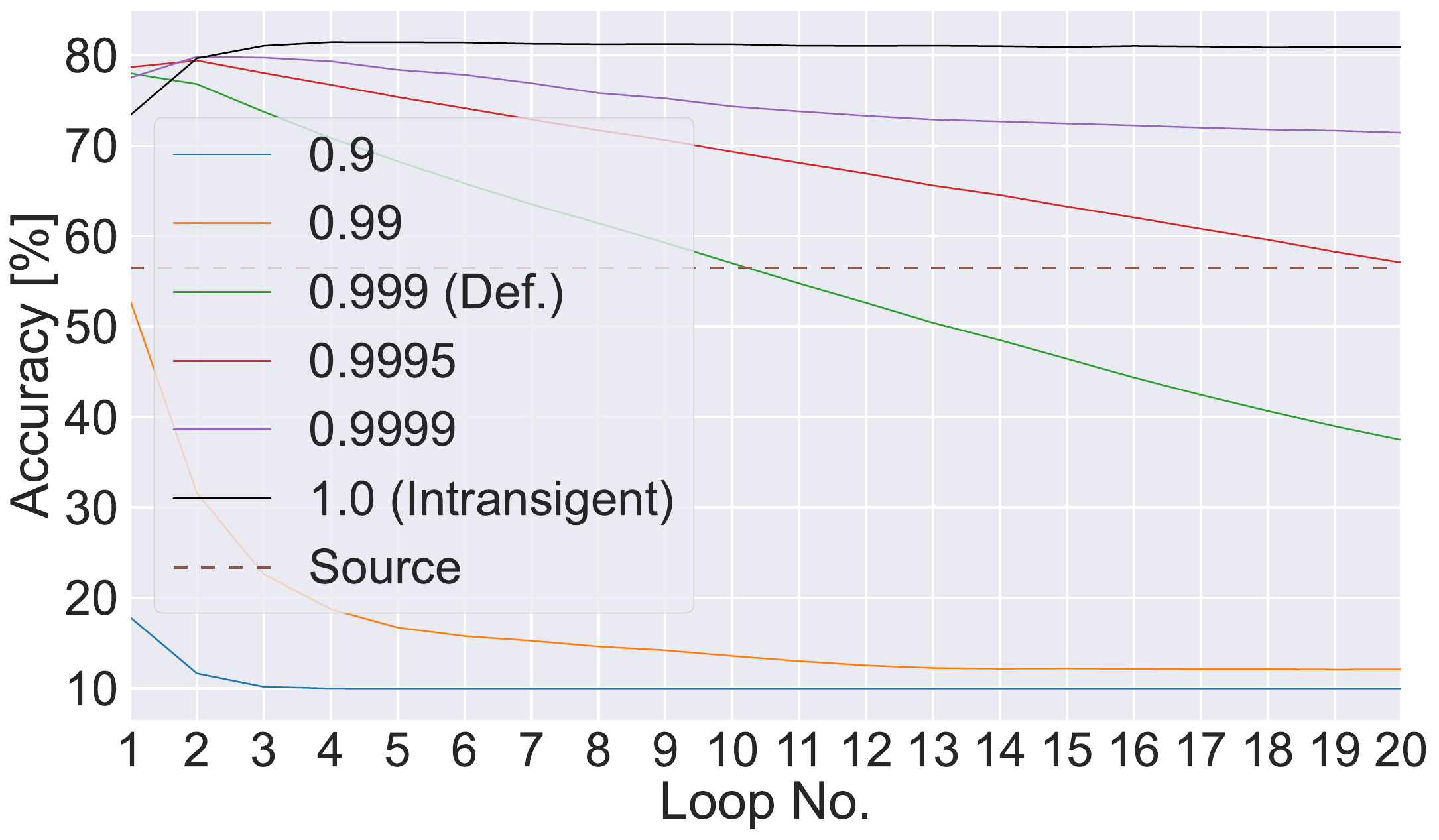}
  \label{fig:ema_petal_cifar}
\end{minipage}
\caption{Mean accuracy [\%] for each loop of common testing sequence on ImageNet-C (L) \textbf{(left)} and CIFAR10-C \textbf{(right)} using PETAL. The brown dashed line indicates the source model accuracy.}
\label{fig:ema_petal}

\vspace*{\fill}
\end{figure*}

\begin{figure}[tbp]
    \centering
    % Row 1
    \begin{subfigure}{0.48\textwidth}
        \includegraphics[width=\linewidth]{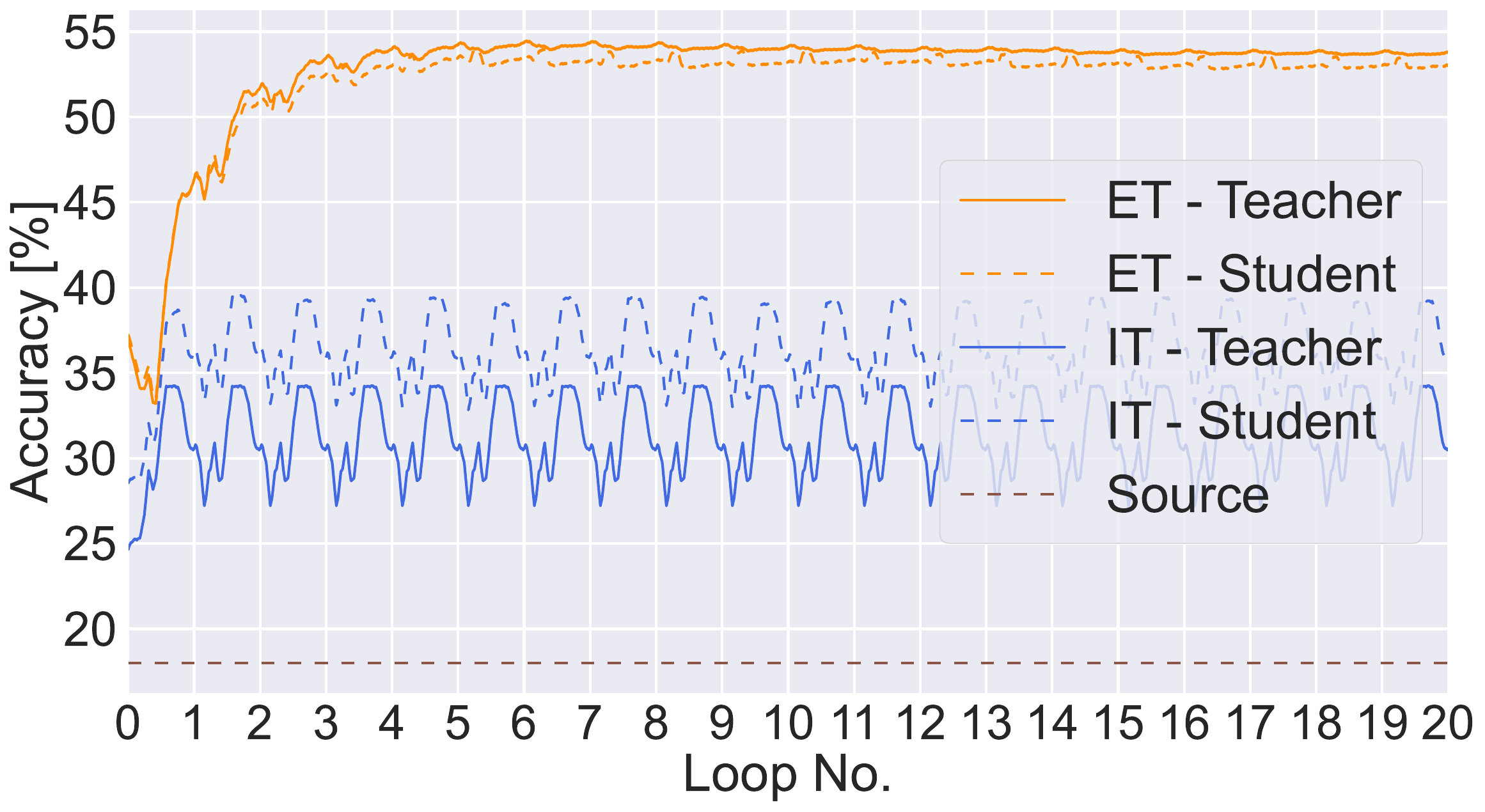}
        \caption{CoTTA, ResNet50}
        \label{fig:ts}
    \end{subfigure}
    \hfill
    \begin{subfigure}{0.48\textwidth}
        \includegraphics[width=\linewidth]{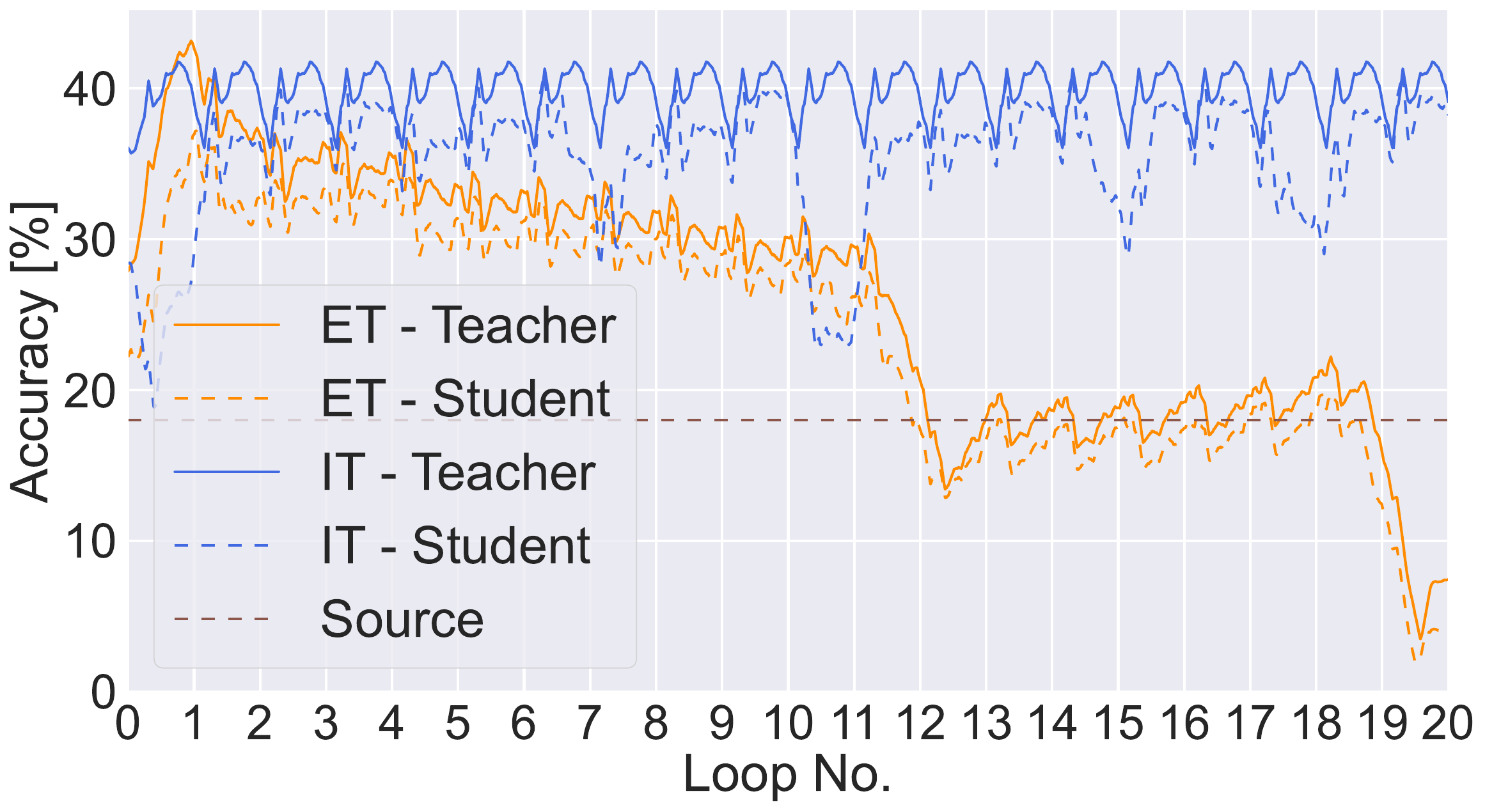}
        \caption{CoTTA, ViT-B16}
        \label{fig:ts_cotta_inc_vit}
    \end{subfigure}

    \vspace{1em} % Add vertical spacing between rows

    % Row 2
    \begin{subfigure}{0.48\textwidth}
        \includegraphics[width=\linewidth]{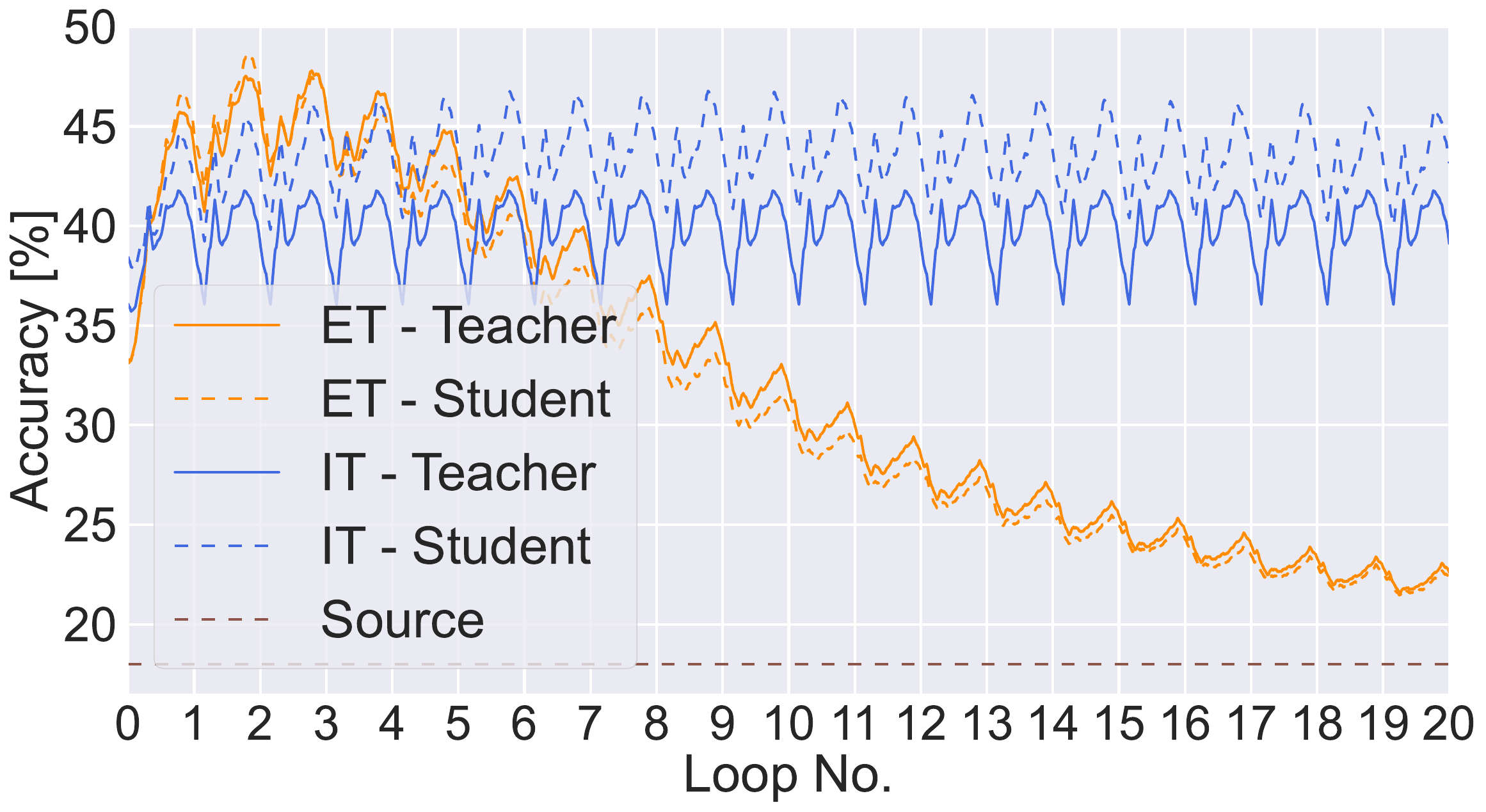}
        \caption{AdaContrast, ViT-B16}
        \label{fig:ts_ada_inc_vit}
    \end{subfigure}
    \hfill
    \begin{subfigure}{0.48\textwidth}
        \includegraphics[width=\linewidth]{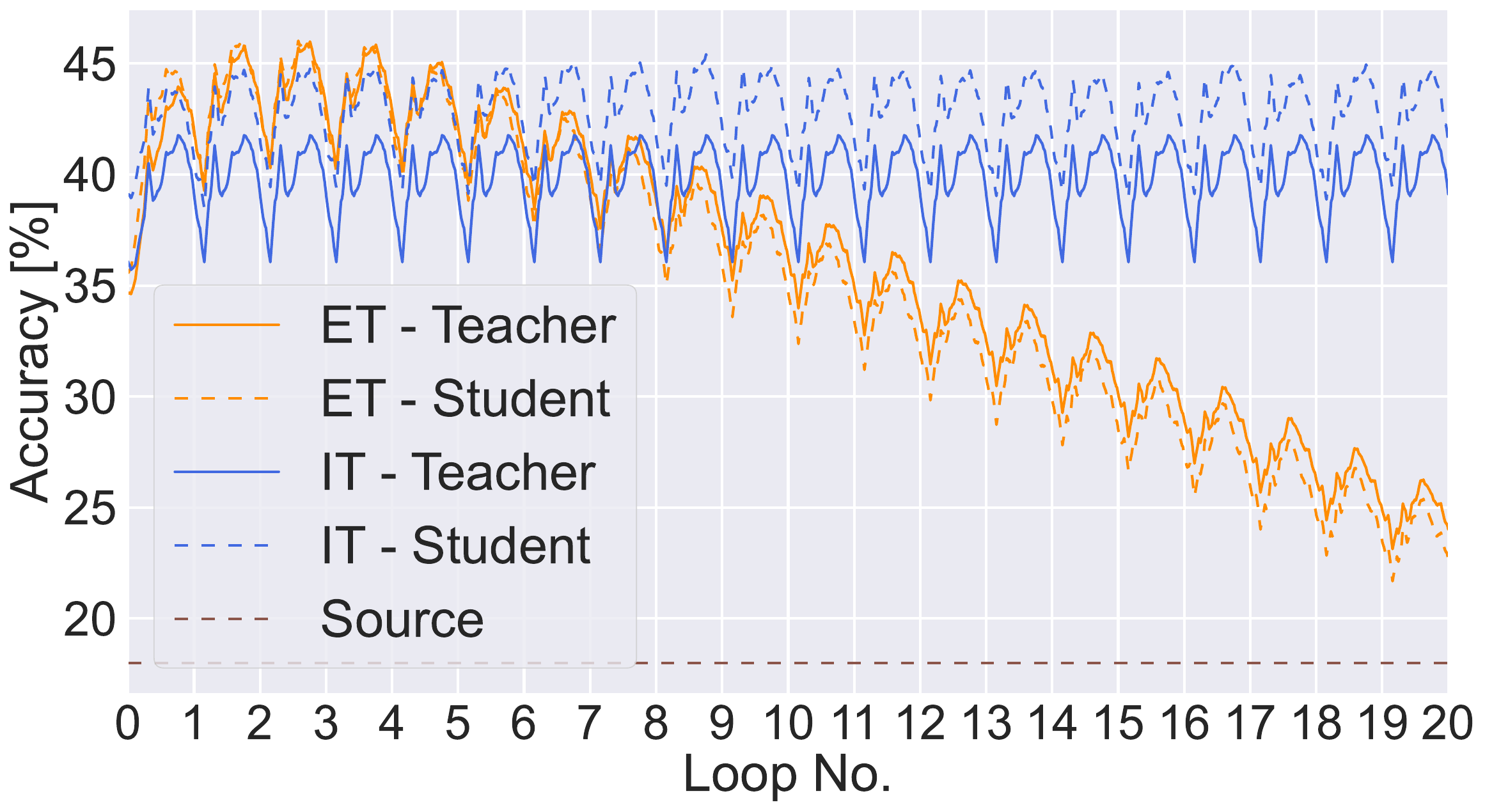}
        \caption{RoTTA, ViT-B16}
        \label{fig:ts_rotta_inc_vit}
    \end{subfigure}

    \vspace{1em}

    % Row 3
    \begin{subfigure}{0.48\textwidth}
        \includegraphics[width=\linewidth]{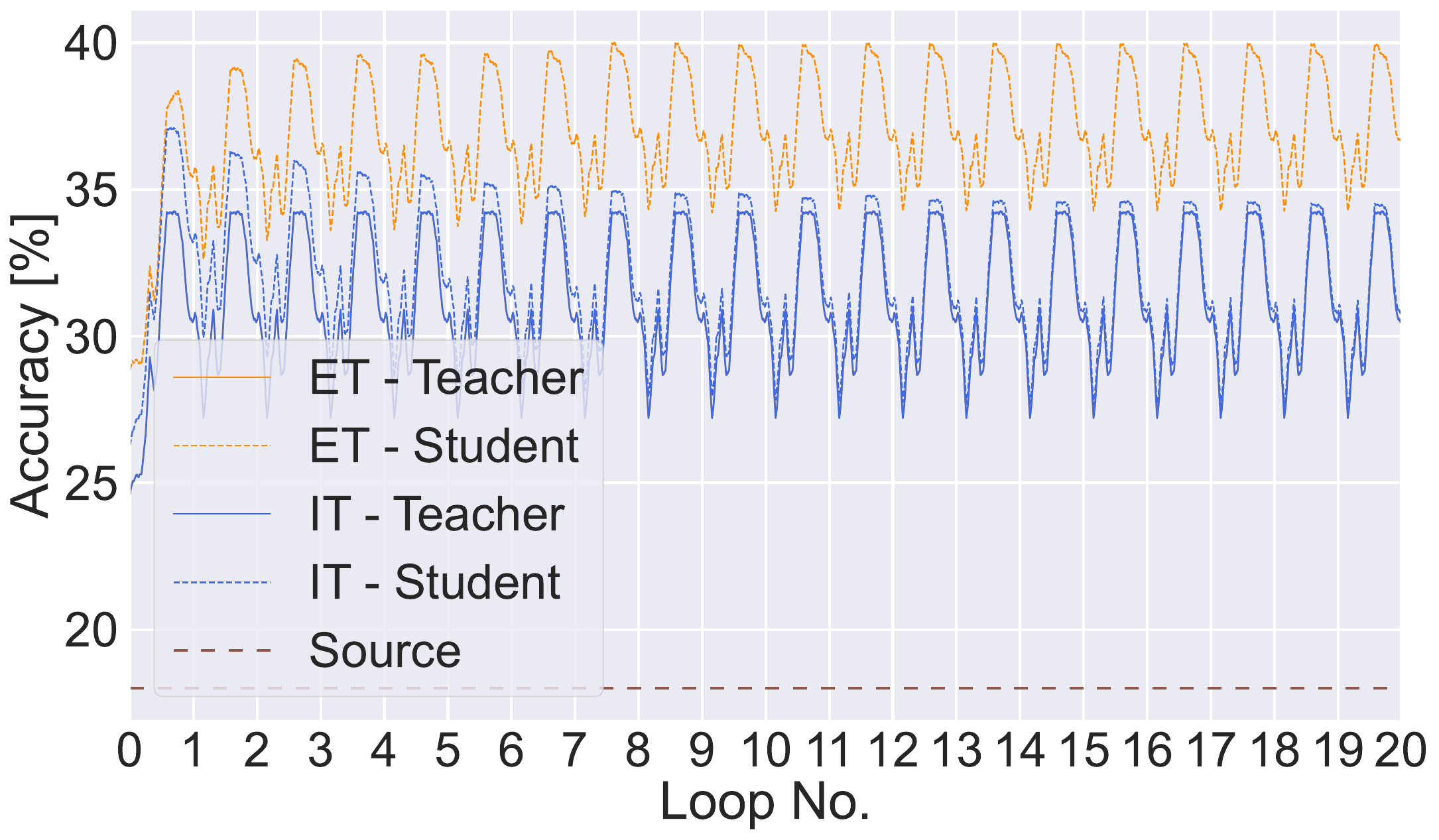}
        \caption{PETAL, ResNet50}
        \label{fig:inc_petal}
    \end{subfigure}
    \hfill
    \begin{subfigure}{0.48\textwidth}
        \includegraphics[width=\linewidth]{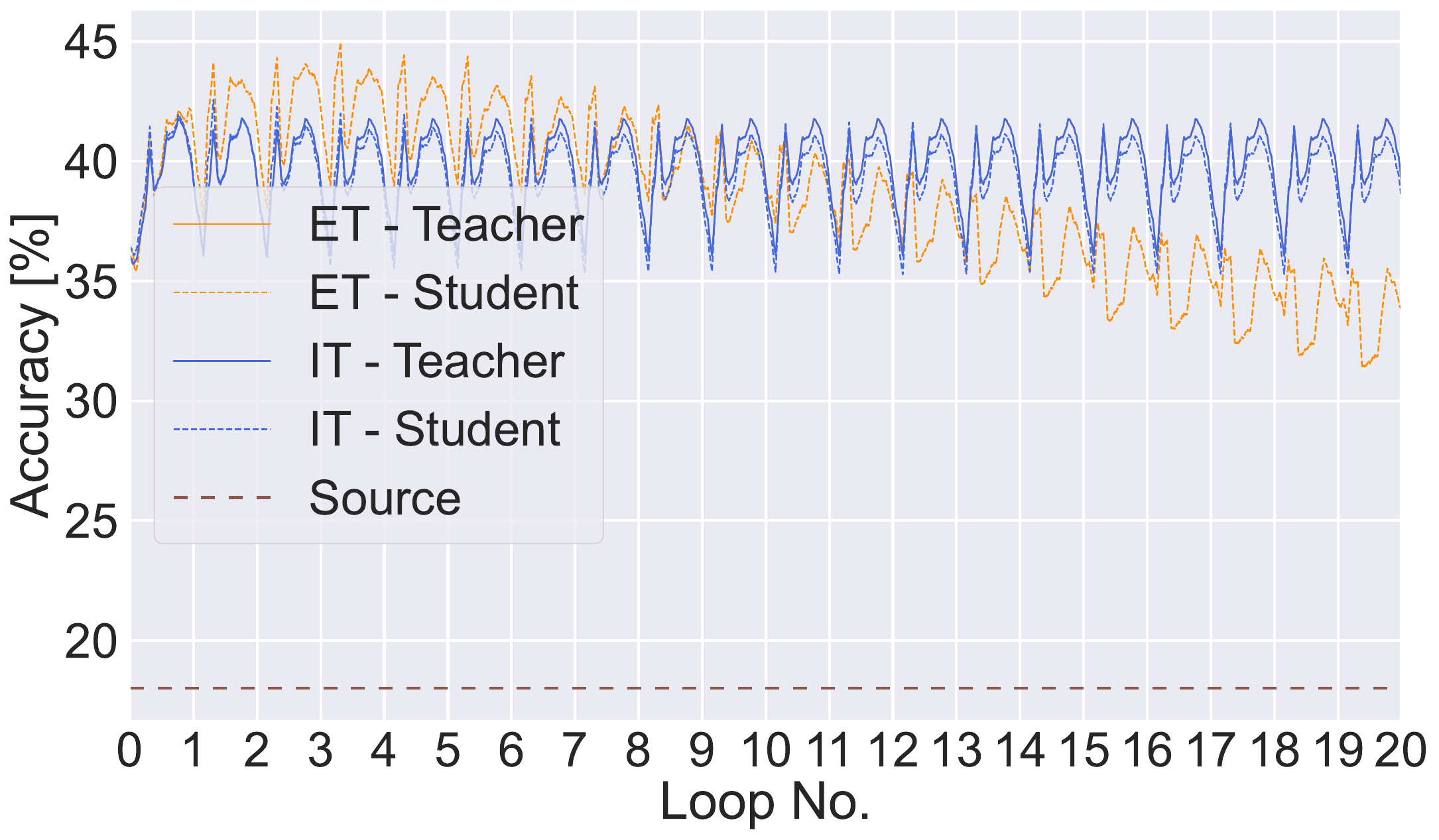}
        \caption{PETAL, ViT-B16}
        \label{fig:inc_petal_vit}
    \end{subfigure}

    \caption{Per-batch accuracy [\%] on ImageNet-C (L). Comparison across different methods and architectures with EMA teacher (ET, orange) and intransigent teacher (IT, blue), for both teacher (solid) and student (dashed).}
    \label{fig:imagenet_results}
\end{figure}

\begin{figure}[tbp]
    \centering
    % Row 1
    \begin{subfigure}{0.48\textwidth}
        \includegraphics[width=\linewidth]{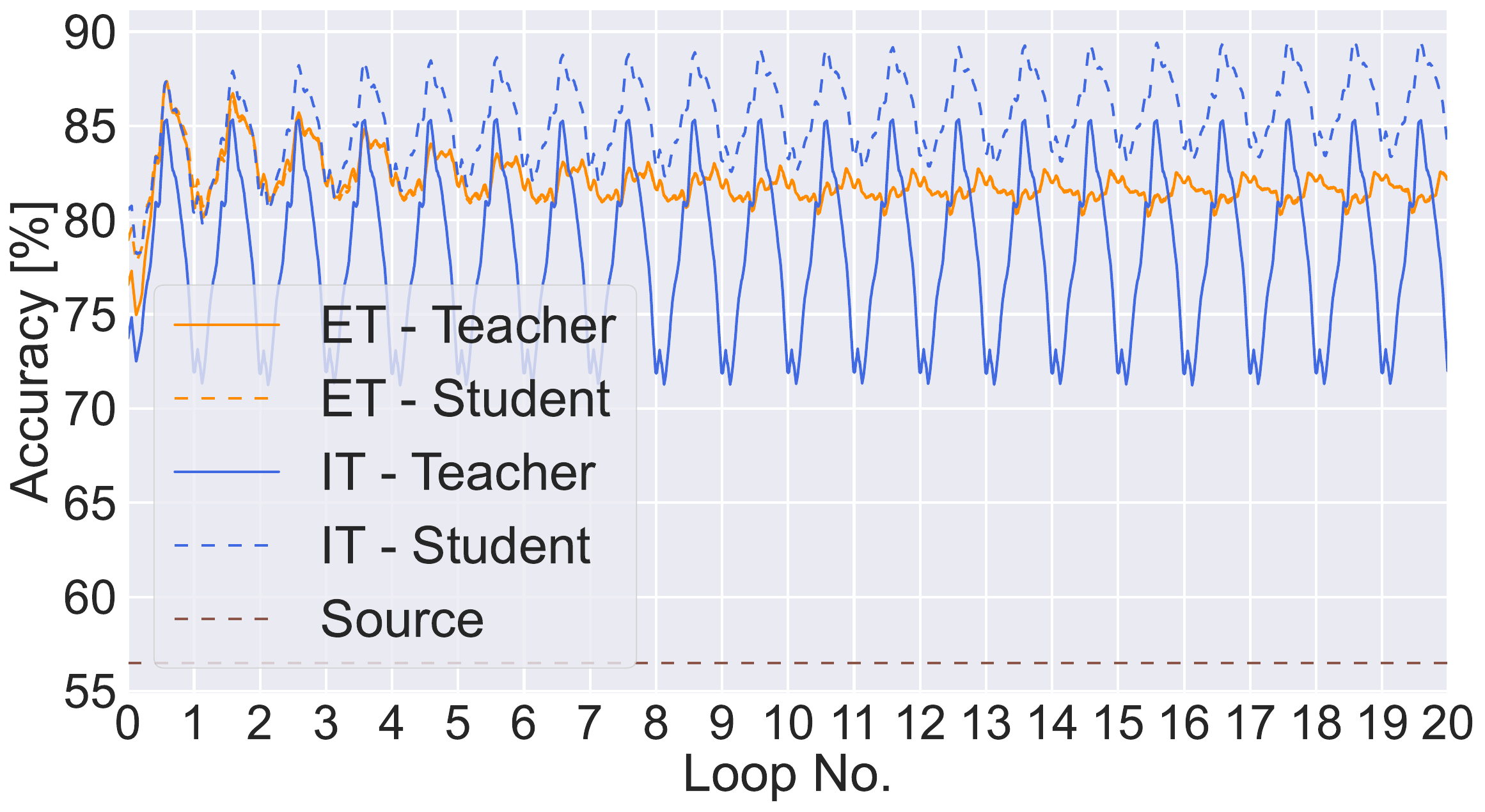}
        \caption{AdaContrast}
        \label{fig:cifar_ada}
    \end{subfigure}
    \hfill
    \begin{subfigure}{0.48\textwidth}
        \includegraphics[width=\linewidth]{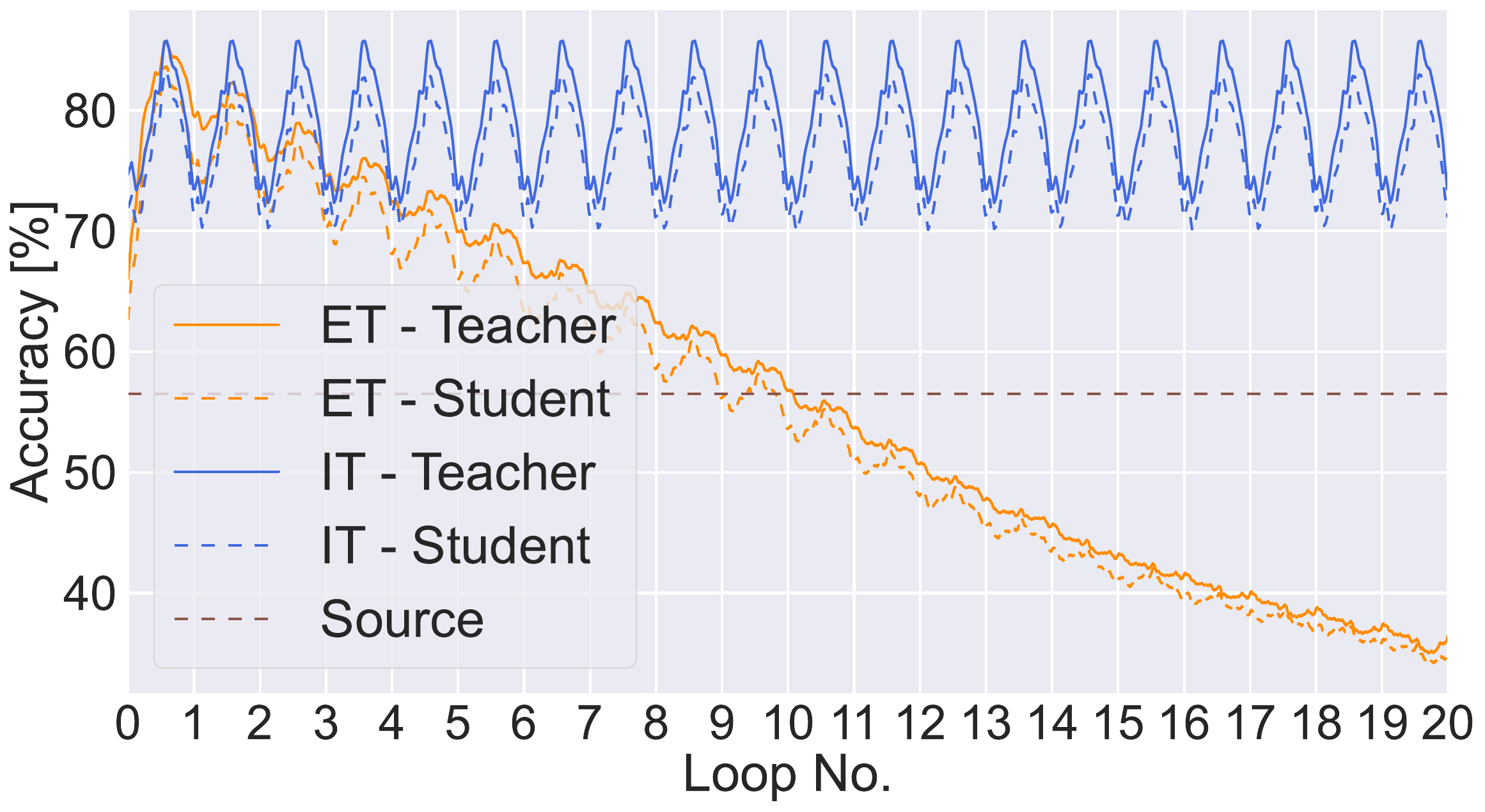}
        \caption{CoTTA}
        \label{fig:cifar_cotta}
    \end{subfigure}

    \vspace{1em}

    % Row 2
    \begin{subfigure}{0.48\textwidth}
        \includegraphics[width=\linewidth]{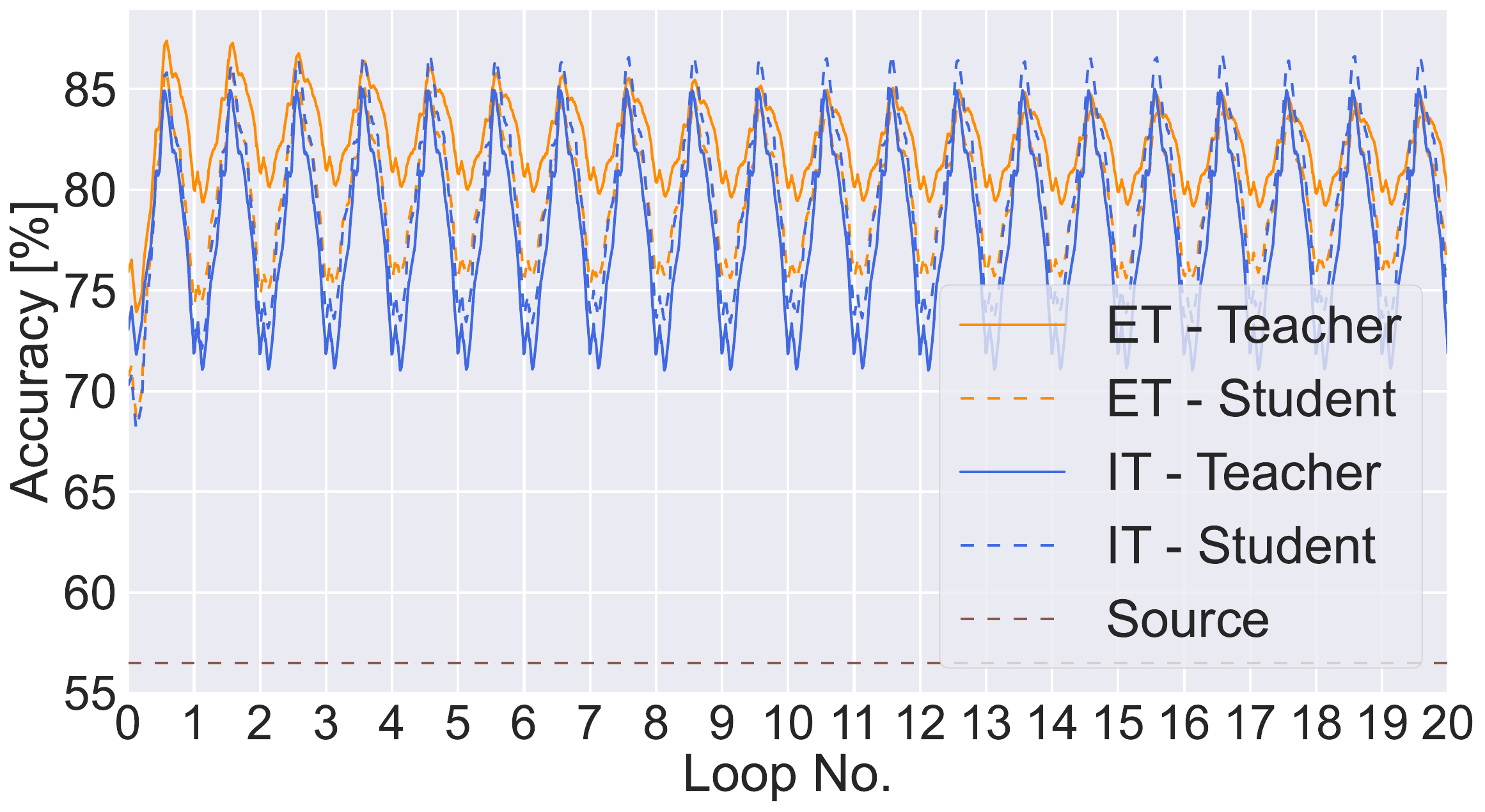}
        \caption{RoTTA}
        \label{fig:cifar_rotta}
    \end{subfigure}
    \hfill
    \begin{subfigure}{0.48\textwidth}
        \includegraphics[width=\linewidth]{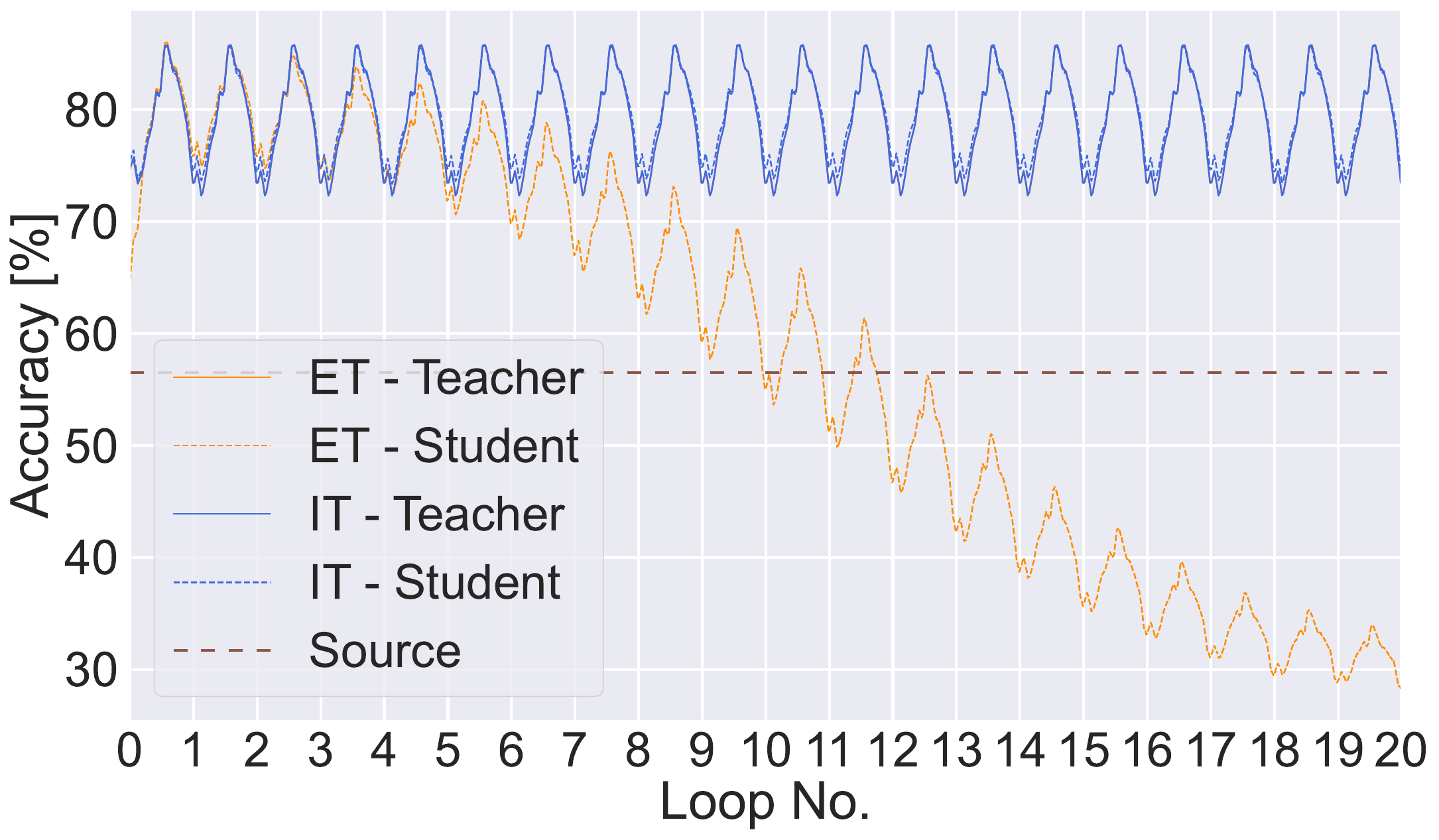}
        \caption{PETAL}
        \label{fig:cifar_petal}
    \end{subfigure}

    \caption{Per-batch accuracy [\%] on CIFAR10-C (L) using WideResNet-28. Comparison with EMA teacher (ET, orange) and intransigent teacher (IT, blue), for both teacher (solid) and student (dashed).}
    \label{fig:cifar_results}
\end{figure}

\begin{figure*}[ht!]
\centering
\begin{minipage}{\textwidth}
\centering
\includegraphics[width=0.89\linewidth]{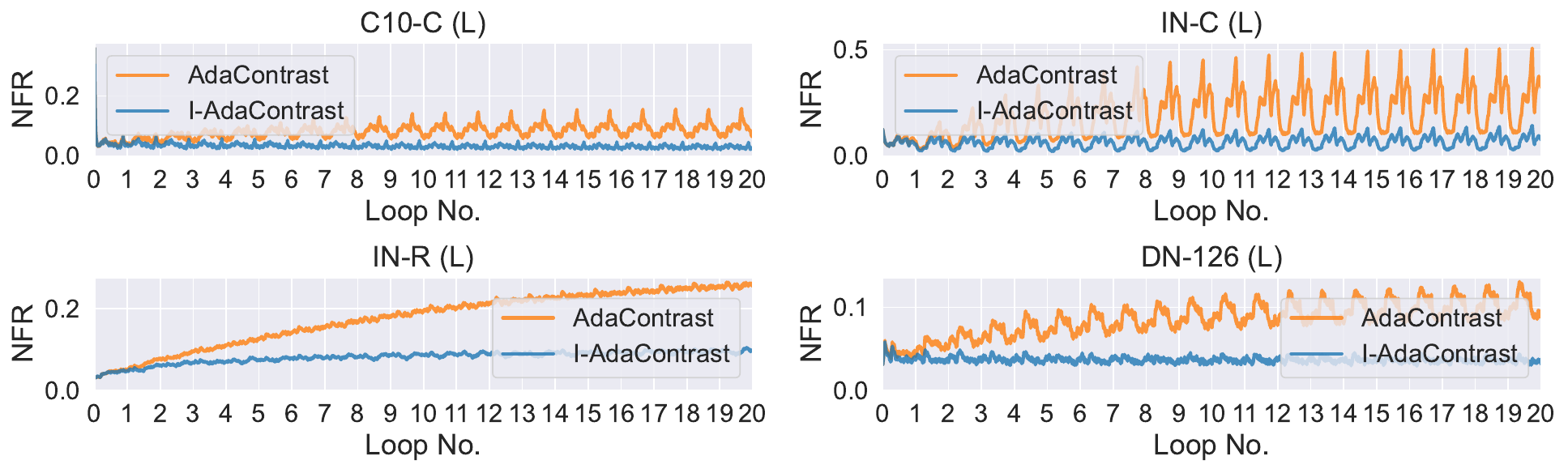}\\
\textbf{(a)} AdaContrast
\end{minipage}

\vspace{1em}  % space between images

\begin{minipage}{\textwidth}
\centering
\includegraphics[width=0.89\linewidth]{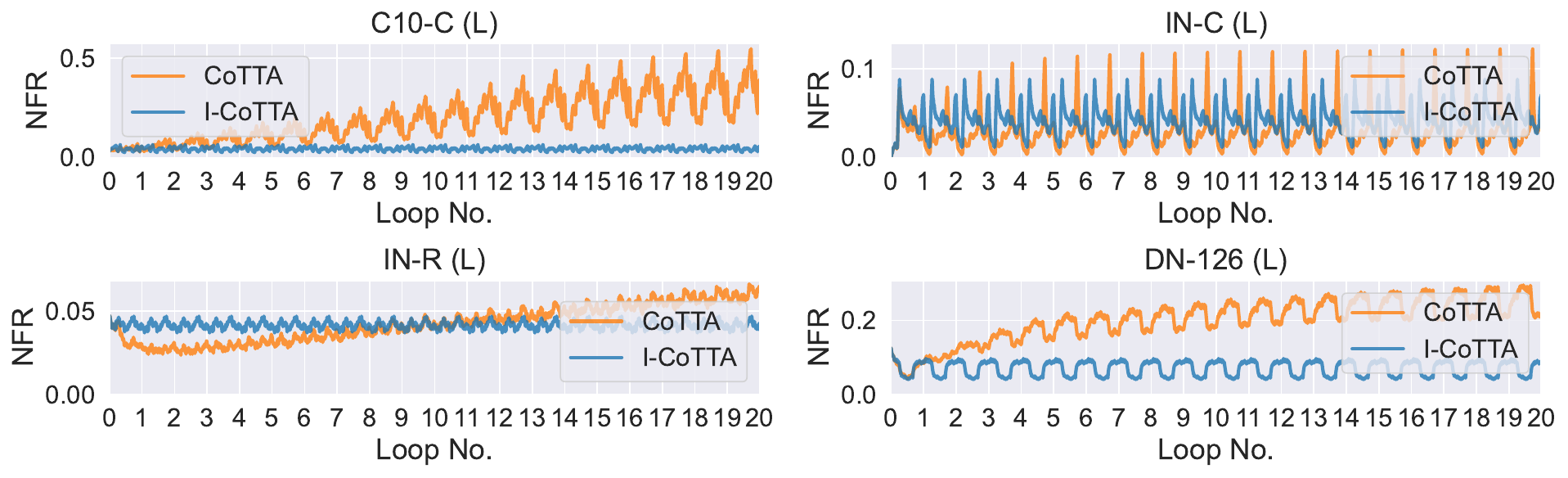}\\
\textbf{(b)} CoTTA
\end{minipage}

\vspace{1em}

\begin{minipage}{\textwidth}
\centering
\includegraphics[width=0.89\linewidth]{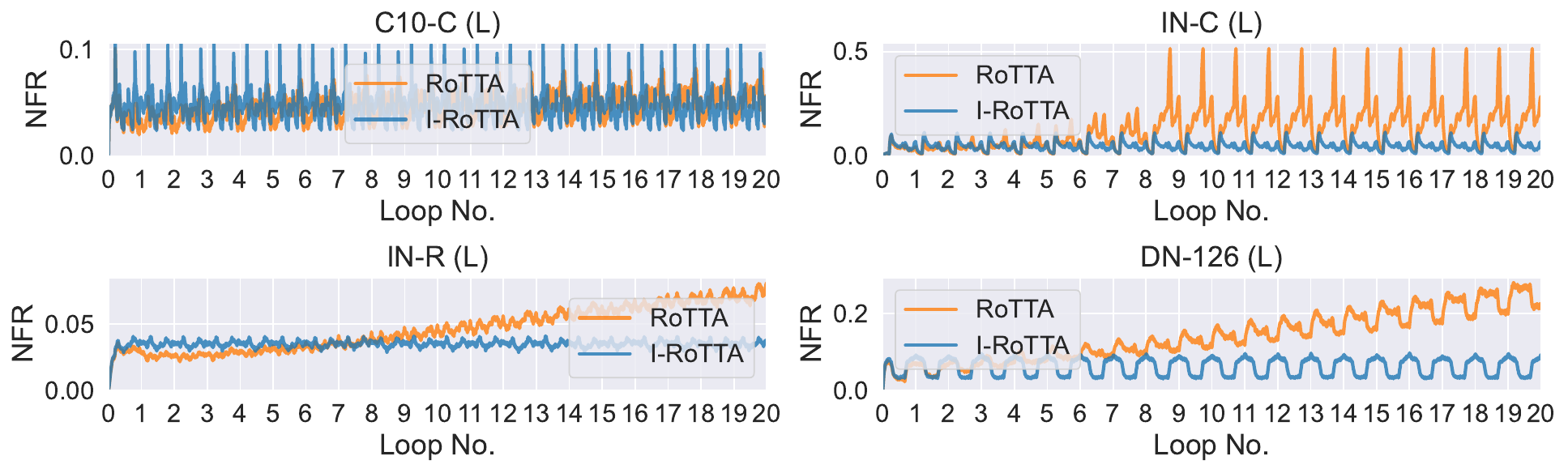}\\
\textbf{(c)} RoTTA
\end{minipage}

\vspace{1em}

\begin{minipage}{\textwidth}
\centering
\includegraphics[width=0.89\linewidth]{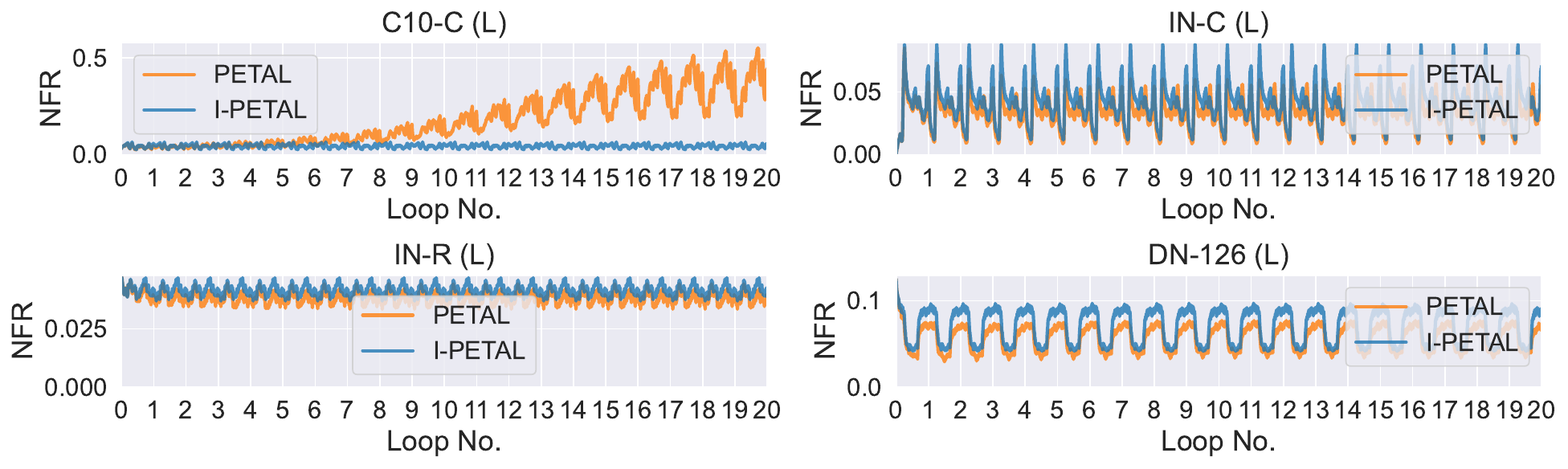}\\
\textbf{(d)} PETAL
\end{minipage}

\caption{
  Per-batch Negative Flips Rate (NFR) of the \textbf{student} model on each benchmark using AdaContrast (\textbf{a}), CoTTA (\textbf{b}), RoTTA (\textbf{c}) and PETAL (\textbf{d}) methods and IT-enhanced version.
}
\label{fig:negative_flips}
\end{figure*}

This section provides additional results not included in the main text. These include results on various architectures when the learning rate is tuned exclusively for the original methods on standard-length benchmarks and directly reused for IT-augmented versions (Tab.~\ref{tab:arch_defseq_tuned}), 
baseline evaluations for different $\beta$ parameters (Tab.~\ref{tab:ema_param}, Figs.~\ref{fig:ema_cotta_ada} and~\ref{fig:ema_petal}) and accuracy plots incorporating the IT technique (Figs.~\ref{fig:imagenet_results} and~\ref{fig:cifar_results}) for various benchmarks and architectures. Additionally, we include per-batch negative flip rate plots (Fig.~\ref{fig:negative_flips}).

\begin{table}[bh!]
\caption{Classification accuracy [\%] for long scenarios on CIFAR10-C and ImageNet-C with different neural network architectures. The value in superscript indicates the improvements over the baseline. The learning rate parameter is adjusted using the Oracle method on the original benchmarks exclusively for original methods and reused for IT-augmented ones.}
% \vspace{-1em}
\centering
\resizebox{0.6\linewidth}{!}{
\begin{tabular}{l@{\hskip 0.3in}c@{\hskip 0.3in}cccc}
\toprule
& \textbf{C10-C (L)} &  \multicolumn{4}{c}{\textbf{IN-C (L)}} \\
& RN26GN & RNXt-50 & ViT-B16 & SViT-T & CNXt tiny \\ 
\midrule
Source & 67.3 & 21.1 & 39.8 & 28.3 & 29.1 \\

AdaCont. & 75.5 & 21.3 & 33.9 & 17.0 & 19.2 \\
I-AdaCont. 
& 79.6\textsuperscript{\textcolor{greenx}{+4.1}} 
& 42.9\textsuperscript{\textcolor{greenx}{+21.6}} 
& 43.5\textsuperscript{\textcolor{greenx}{+9.6}} 
& 30.8\textsuperscript{\textcolor{greenx}{+13.8}} 
& 32.5\textsuperscript{\textcolor{greenx}{13.3}} 
\\

CoTTA & 16.5 & 57.1 & 35.3 & 25.6 & 22.1 \\
I-CoTTA 
& 61.6\textsuperscript{\textcolor{greenx}{+45.1}} 
& 38.4\textsuperscript{\textcolor{redx}{-18.7}} 
& 39.9\textsuperscript{\textcolor{greenx}{+4.6}} 
& 28.9\textsuperscript{\textcolor{greenx}{+3.3}} 
& 29.5\textsuperscript{\textcolor{greenx}{+7.4}} 
\\

RoTTA & 62.8 & 16.2 & 37.1 & 7.6 & 18.7 \\
I-RoTTA 
& 70.5\textsuperscript{\textcolor{greenx}{+7.7}} 
& 35.2\textsuperscript{\textcolor{greenx}{+19.0}} 
& 40.7\textsuperscript{\textcolor{greenx}{+3.6}} 
& 26.5\textsuperscript{\textcolor{greenx}{+18.9}} 
& 29.7\textsuperscript{\textcolor{greenx}{+11.0}} 
\\

PETAL & 15.0 & 56.1 & 38.4 & 22.1 & 24.8 \\
I-PETAL
& 52.8\textsuperscript{\textcolor{greenx}{+37.8}} 
& 36.2\textsuperscript{\textcolor{redx}{-19.9}} 
& 39.4\textsuperscript{\textcolor{greenx}{1.0}} 
& 26.6\textsuperscript{\textcolor{greenx}{4.5}} 
& 27.9\textsuperscript{\textcolor{greenx}{3.1}} 
\\

\bottomrule
\end{tabular}
}
\label{tab:arch_defseq_tuned}
\end{table}

\begin{table}[!ht]
\caption{Mean accuracy [\%] for different exponential moving average $\beta$ parameter. AdaContrast and CoTTA default originally to 0.999. (L) stands for the adaptation sequence being repeated 20 times. \textcolor{gray}{Gray} color indicates model collapse -- performance worse than no adaptation (Source).
}
\centering
\resizebox{0.6\linewidth}{!}{

\begin{tabular}{c@{\hskip 0.1in}c@{\hskip 0.1in}cccccc}
    \toprule
    \textbf{Dataset} & \textbf{Method} & \textbf{0.9} & \textbf{0.99} & \textbf{0.999} & \textbf{0.9995} & \textbf{0.9999} & \textbf{IT}  \\
    \midrule
    \multirow{2}{*}{C10-C (L)} 
    % & Source & - & - & - & - & - & \textit{56.5} \\
    & AdaCont. & 79.0 & 79.3 & 81.9 & 83.2 & \textbf{85.8} & 85.4 \\
    % intransigent: source preds: 68.4, teacher preds: 76.8
    & CoTTA & \textcolor{gray}{10.5} & \textcolor{gray}{14.9} & \textcolor{gray}{55.9} & 68.6 & \textbf{78.3} & 68.4 \\
    \midrule
    \multirow{2}{*}{IN-C (L)} 
    % & Source & - & - & - & - & - & \textit{18.0} \\
    & AdaCont.  & \textcolor{gray}{1.2} & \textcolor{gray}{5.4} & 18.8 & 25.6 & 38.6 & \textbf{40.4} \\
    % intransigent: source preds: 35.4, teacher preds: 31.4
    & CoTTA & \textcolor{gray}{0.3} & 23.6 & 52.8 & \textbf{55.1} & 50.9 & 35.4 \\
    \bottomrule
\end{tabular}
}
\label{tab:ema_param}
\end{table}

\clearpage

% --- Figure 2: CIFAR10-C Results (4 plots) ---

\end{document}